\documentclass{article}

    \PassOptionsToPackage{numbers, compress}{natbib}
\usepackage[preprint]{neurips_2025}

\usepackage[utf8]{inputenc} % allow utf-8 input
\usepackage[T1]{fontenc}    % use 8-bit T1 fonts
\usepackage[title]{appendix}
\usepackage{amsmath}
\usepackage{amsfonts}       % blackboard math symbols
\usepackage{amsthm}
\usepackage{bm}
\usepackage{mathrsfs}
\usepackage{etoc}
\usepackage{microtype}      % microtypography
\usepackage{graphicx}
\usepackage[table]{xcolor}  % colors (+ table support)
\usepackage{hyperref}       % hyperlinks
\usepackage{url}            % simple URL typesetting
\usepackage{nicefrac}       % compact symbols for 1/2, etc.

\usepackage{booktabs}       % professional-quality tables
\usepackage{array}
\usepackage{tabularx}
\usepackage{longtable}
\usepackage{multirow}
\usepackage{makecell}
\usepackage{pifont}
\usepackage{enumitem}

\usepackage{algorithm}
\usepackage{algorithmicx}
\usepackage[noend]{algpseudocode}

\usepackage{subcaption}
\usepackage{placeins}

\usepackage{siunitx}

\usepackage{colortbl}
\usepackage{xcolor}
\usepackage{tcolorbox}
\usepackage{amssymb}
\usepackage{graphicx}
\usepackage[utf8]{inputenc}

\usepackage[table]{xcolor}
\usepackage{booktabs}
\usepackage{longtable}
\usepackage{multirow}

\tcbuselibrary{skins, breakable} 

\definecolor{promptblue}{HTML}{005CC5}   % JSON Key
\definecolor{promptpurple}{HTML}{6F42C1} % Enums / Strings
\definecolor{promptgreen}{HTML}{22863A}  % Types / Examples
\definecolor{promptred}{HTML}{D73A49}    % Critical / Emphasis
\definecolor{promptteal}{HTML}{008080}   % Persona / Audience
\definecolor{codebg}{HTML}{F6F8FA}

\newcommand{\grayarrow}{\textcolor{gray}{\normalfont$\hookrightarrow$}\; }
\newcommand{\jsonkey}[1]{\textcolor{promptblue}{"#1"}}
\newcommand{\jsonval}[1]{\textcolor{promptpurple}{"#1"}}
\newcommand{\jsontype}[1]{\textcolor{promptgreen}{#1}}

\newcommand{\dsmodel}{\raisebox{-0.2em}{\includegraphics[height=1em]{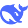}}\,\,deepseek-chat}
\newcommand{\qwenmodel}{\raisebox{-0.2em}{\includegraphics[height=1em]{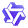}}\,\,qwen3-vl-flash}
\newcommand{\gptmodel}{\raisebox{-0.2em}{\includegraphics[height=1em]{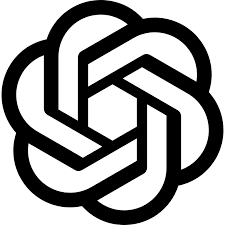}}\,\,gpt-5-mini}
\newcommand{\geminimodel}{\raisebox{-0.2em}{\includegraphics[height=1em]{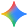}}\,\,gemini-3-pro-preview}

\newcommand{\imagefocus}{\raisebox{-0.2em}{\includegraphics[height=1em]{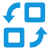}}\,\,\textit{Image Focus}}
\newcommand{\texttodiagram}{\raisebox{-0.2em}{\includegraphics[height=1em]{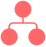}}\,\,\textit{Text to Diagram}}
\newcommand{\keynote}{\raisebox{-0.2em}{\includegraphics[height=1em]{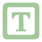}}\,\,\textit{Keynote}}
\newcommand{\dataviz}{\raisebox{-0.2em}{\includegraphics[height=1em]{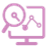}}\,\,\textit{Data Visualization}}
\newcommand{\motion}{\raisebox{-0.2em}{\includegraphics[height=1em]{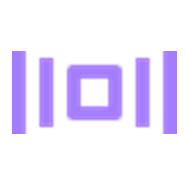}}\,\,\textit{Motion}}
\newcommand{\bggg}{\raisebox{-0.2em}{\includegraphics[height=1em]{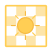}}\,\,\textit{Background}}
\newcommand{\autolayout}{\raisebox{-0.2em}{\includegraphics[height=1em]{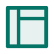}}\,\,\textit{Auto Layout}}

\newtcolorbox{innerpromptbox}[1][]{
    enhanced,           
    breakable,          
    colback=white,      
    colframe=black!80,     
    boxrule=0.5pt,      
    arc=0pt,            
    outer arc=0pt,
    left=6pt, right=6pt, top=6pt, bottom=6pt,
    attach boxed title to top center={yshift=3mm}, 
    coltitle=black!60,         
    boxed title style={empty, size=minimal},
    before skip=2em,       
    after skip=1em,
    underlay={
        \draw[black!80, line width=0.5pt]
            ([xshift=3pt, yshift=0pt]frame.north east) --
            ([xshift=3pt, yshift=-3pt]frame.south east) --
            ([xshift=0pt, yshift=-3pt]frame.south west); 
    },
    fontupper=\scriptsize\ttfamily\raggedright,
    parbox=false, 
    #1 
}

\newenvironment{promptbox}[3][]{%
    \begin{innerpromptbox}[title={\small \textbf{Listing #2}: System Prompt of \textbf{#3}},#1]%
}{%
    \end{innerpromptbox}%
}

\definecolor{DeepSlideColor}{HTML}{0D9488}
\definecolor{ChildToParent}{HTML}{CC6A1A} % children -> parent
\definecolor{ParentToChild}{HTML}{258D8D} % parent -> children
\newcommand{\cmark}{\textcolor{DeepSlideColor}{\ding{51}}}
\newcommand{\xmark}{\textcolor{red}{\ding{55}}}
\newcommand{\umark}{\textcolor{gray}{\(\circ\)}}

\newcommand{\DPW}{0.23\linewidth}
\newcolumntype{L}{>{\raggedright\arraybackslash}p{\DPW}}
\newcolumntype{Y}{>{\centering\arraybackslash}X}

\newcommand{\GroupRow}[2]{%
  \multicolumn{8}{>{\columncolor{black!7}}l}{%
    \textbf{#1}\hspace{0.1em}\textcolor{gray}{(#2)}%
  }\\[-0.15ex]
}
\definecolor{deltaGray}{HTML}{6C6C6C}
\definecolor{barBg}{HTML}{F0F7FF}
\definecolor{barFill}{HTML}{8BB8FF}
\definecolor{barFillHi}{HTML}{3A6DCC}

\newcommand{\BarW}{4.8}
\newcommand{\BarT}{0.34ex}
\newcommand{\CellGap}{0.18ex}
\newcommand{\CellShift}{0.80ex}

\newcommand{\baronly}[2]{%
  \begin{tikzpicture}[x=1em,y=1ex,baseline=-0.10ex]
    \draw[line cap=round,line width=\BarT,draw=barBg] (0,0) -- (\BarW,0);
    \draw[line cap=round,line width=\BarT,draw=#2]   (0,0) -- ({\BarW*#1},0);
  \end{tikzpicture}%
}
\newcommand{\cellbar}[3]{%
  \raisebox{-\CellShift}{%
    \vbox{\offinterlineskip
      \hbox{\strut #2}%
      \kern\CellGap
      \hbox{\baronly{#1}{#3}}%
    }%
  }%
}
\newcommand{\barcell}[2]{\cellbar{#1}{#2}{barFill}}
\newcommand{\barcellhi}[2]{\cellbar{#1}{\textbf{#2}}{barFillHi}}
\definecolor{DomainGray}{HTML}{F2F2F2}
\definecolor{DeepSlideRow}{HTML}{E9F6EF}
\definecolor{DeepSlideStrong}{HTML}{BFE8D0}
\newcommand{\ds}[1]{\textbf{#1}}
\newcommand{\dsstrong}[1]{\cellcolor{DeepSlideStrong}\ds{#1}}
\newcommand{\metrichead}[1]{\textsc{#1}}

\definecolor{CatBlue}{HTML}{F2F8FF}
\newcommand{\CatRow}[1]{%
  \rowcolor{CatBlue}%
  \multicolumn{3}{@{}l@{}}{\bfseries\scshape\strut\hspace{0.0em}#1}%
  \\[-0.15ex]
}

\definecolor{Stage1Bg}{HTML}{FE9100}  % 更深更暗的琥珀色
\definecolor{Stage2Bg}{HTML}{F56868}  % 更深更暗的鲑鱼色
\definecolor{Stage3Bg}{HTML}{2ABE78}  % 更深更暗的青绿色
\definecolor{Stage4Bg}{HTML}{26A6FF}  % 更深更暗的天蓝色

\newcommand{\stagerow}[1]{%
  \rowcolor{gray!15}[0pt][0pt]%
  \multicolumn{2}{@{}l@{}}{\textit{#1}}\\
}

\newtheorem{example}{Example}

\title{DeepSlide: From Artifacts to Presentation Delivery}

\author{
  Ming Yang\thanks{Project Leader},
  Zhiwei Zhang,
  Jiahang Li,
  Haoseng Liu,
  Yuzheng Cai,
  Weiguo Zheng\thanks{Corresponding Author} \\
  School of Data Science, Fudan University, Shanghai, China \\
  \texttt{\{yangm24, zwzhang25, jiahangli25, hsliu23, yuzhengcai21\}@m.fudan.edu.cn} \\
  \texttt{zhengweiguo@fudan.edu.cn}
}

\begin{document}
\etocdepthtag.toc{mtmain} 

\maketitle

\begin{abstract}
Presentations are a primary medium for scholarly communication, yet most AI slide generators optimize the \textit{artifact} (a visually plausible deck) while under-optimizing the \textit{delivery process} (pacing, narrative, and presentation preparation).
We present \textcolor{DeepSlideColor}{\textit{DeepSlide}}, a human-in-the-loop multi-agent system that supports preparing the \textit{full presentation process}, from requirement elicitation and time-budgeted narrative planning, to evidence-grounded slide--script generation, attention augmentation, and rehearsal support.
DeepSlide integrates (i) a controllable logical-chain planner with per-node time budgets, (ii) a lightweight content-tree retriever for grounding, (iii) Markov-style sequential rendering with style inheritance, and (iv) sandboxed execution with minimal repair to ensure renderability.
We further introduce a dual-scoreboard benchmark that cleanly separates static artifact quality from dynamic delivery excellence.
Across 20 domains and diverse audience profiles, \textit{DeepSlide} matches strong baselines on artifact quality while consistently achieving larger gains on delivery metrics, improving narrative flow, pacing precision, and slide--script synergy with clearer attention guidance.
\end{abstract}

\begin{figure}[htbp]
% \small
  \centering
  % \vspace{-2mm}
  \includegraphics[width=0.85\linewidth]{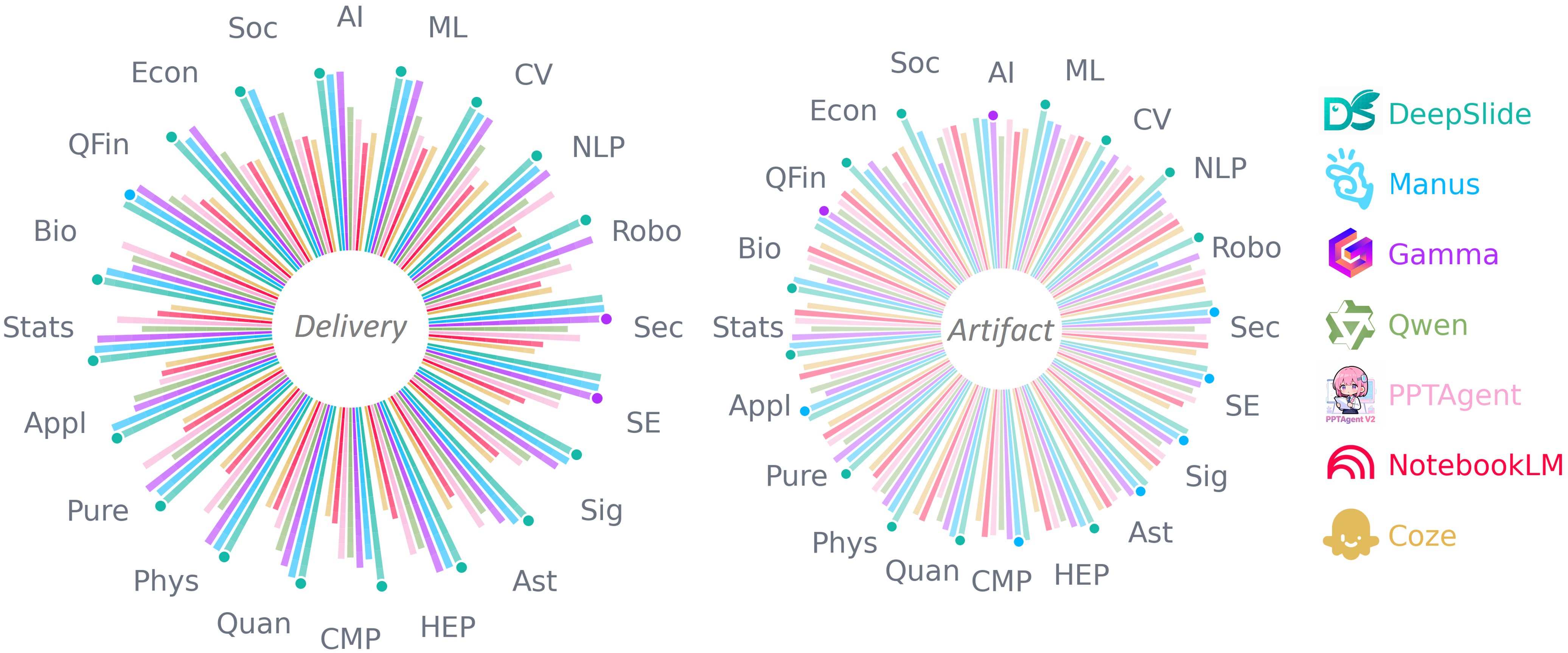}
  % \vspace{-4mm}
  \caption{Main experiment on 20 domains (left is delivery scoreboard, while right is artifact scoreboard).}
  \label{fig:main_exp_perf}
  % \vspace{-2mm}
\end{figure}

\section{Introduction}
Presentations are a primary medium for communicating information and ideas, playing a central role in scholarly exchange~\cite{how_scientists_communicate}. Delivering a high-quality talk typically requires not only a well-designed slide deck, but also a coherent, well-paced narrative aligned with the slides and substantial preparation (e.g., rehearsal and planning for audience interaction)~\cite{the_craft_of_sc_pre}. As a result, talk preparation remains time-consuming, spanning both content authoring and delivery readiness.

Recent advances in large language models (LLMs) and multimodal models have begun to reduce this burden~\cite{ge2025autopresent,maheshwari2024presentations,rw_mondal-etal-2024-presentations} by improving document understanding and information extraction, and enabling increasingly reliable multimodal synthesis. Leveraging these model capabilities, agentic systems can ingest source materials, synthesize textual and visual content, and orchestrate multi-step workflows with structured decisions (e.g., outlining, emphasis, pacing). Meanwhile, engineering frameworks such as LangChain~\cite{langchain}, AutoGPT~\cite{autogpt}, AutoGen~\cite{autogen}, CAMEL~\cite{camel}, and MetaGPT~\cite{hong2024metagpt} further lower the implementation cost of tool use and multi-agent coordination. Together, these advances have made presentation agents practical and productizable, exemplified by Manus~\cite{manus-web}, Gamma~\cite{Gamma-web}, NotebookLM~\cite{notebooklm-web}, Qwen~\cite{Qwen-web}, Coze~\cite{Coze-web}, and the open-source PPTAgent~\cite{rw_zheng-etal-2025-pptagent}.These systems can already produce visually polished slide decks quickly, often with editing and refinement support.

However, prior research (e.g.,~\cite{what_makes_a_good_presentation}) together with practical observations suggests that a good presentation should not be defined primarily by how visually appealing the materials are. Instead, it hinges on whether information is structured and delivered in a way that aligns with the audience's cognitive and attentional constraints---clear organization, restrained distractions, and sustained guidance of audience attention. Consequently, even if these systems can produce polished and editable decks, artifact quality alone is insufficient to be equated with a high-quality presentation~\cite{how_scientists_communicate}. Effective presentation delivery is a human-centered communication problem, not a slide-design problem alone.

We summarize three key gaps in existing slide agents:
(1) \textbf{Missing narrative strategy choices.} Most systems either skip explicit narrative planning or output a single, generic outline that is often non-editable and weakly personalized. More critically, they rarely treat \emph{storytelling strategy} as a controllable design space: users are not offered multiple narrative styles to choose from (e.g., skeptic-to-believer persuasion, myth-busting reframing, trade-off navigation, or detective-style ablation reveal), nor are they guided to allocate time-budgeted emphasis accordingly (what to elaborate vs.\ compress).
(2) \textbf{Limited delivery-time attention strategy.} Existing agents predominantly deliver static slide decks; for long-form logical arguments, complex experimental results, or detail-dense figures, they often lack content-aware mechanisms to guide attention during delivery (e.g., progressive revelation, focus/zoom cues, and tailored visual encodings). Although some agents (e.g., Qwen~\cite{Qwen-web}, Coze~\cite{Coze-web}) generate mind maps or data visualizations, these outputs are often static and template-driven, providing limited \emph{delivery-time} attention guidance tailored to the actual slide content; predefined animations or template-based highlights, when available, are rarely content-aware enough to adapt attention strategy to dense figures and complex results.
(3) \textbf{Insufficient rehearsal support.} Most approaches stop at producing slides, offering limited support for rehearsal and delivery preparation (e.g., generating a well-aligned, non-redundant script, anticipating live contingencies, and providing practice-time feedback), leaving users to design narration, plan interaction, and rehearse on their own.
Overall, current slide agents mainly reduce the cost of deck authoring, but do not fully alleviate the end-to-end workload of presentation preparation.

\begin{figure}
    \centering
    \includegraphics[width=\linewidth]{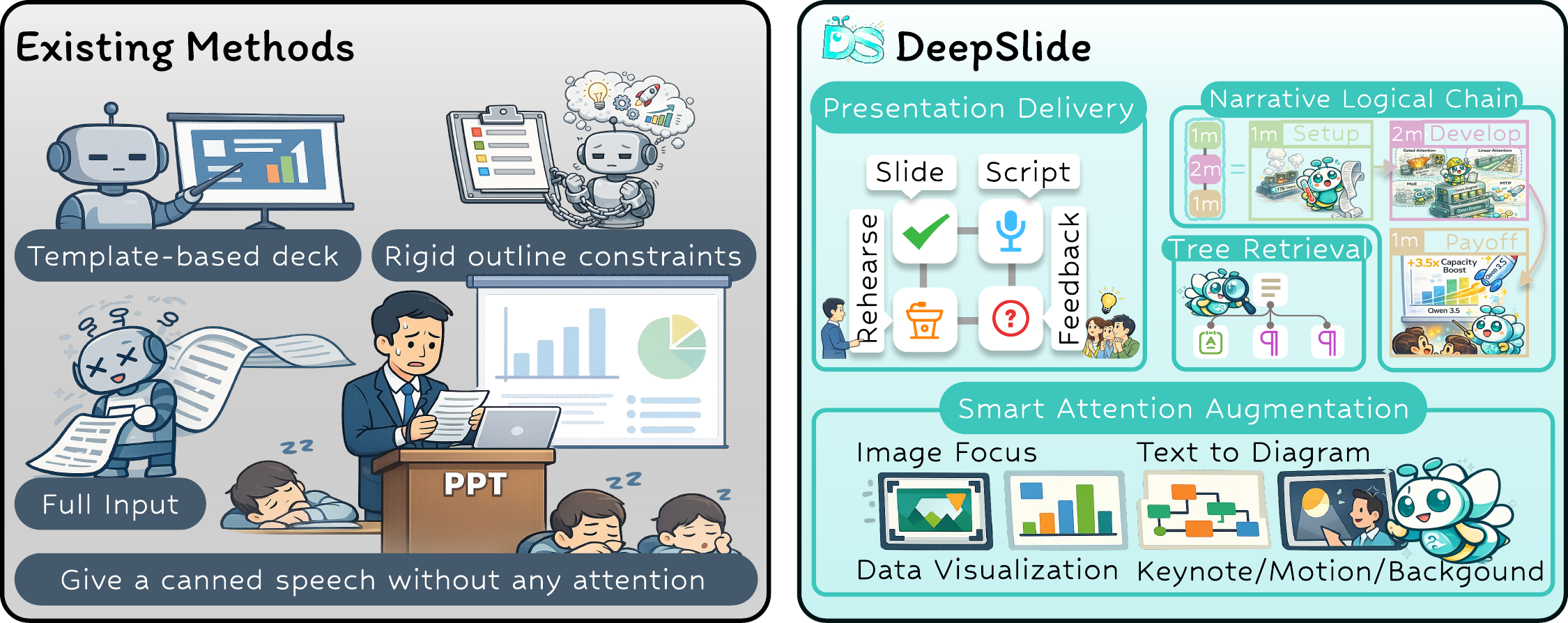}
    \caption{Limitations of existing approaches and the \textit{DeepSlide}.}
    \label{fig:comp}
\end{figure}

\begin{figure}[htbp]
% \small
  \centering
  % \vspace{-2mm}
  \includegraphics[width=\linewidth]{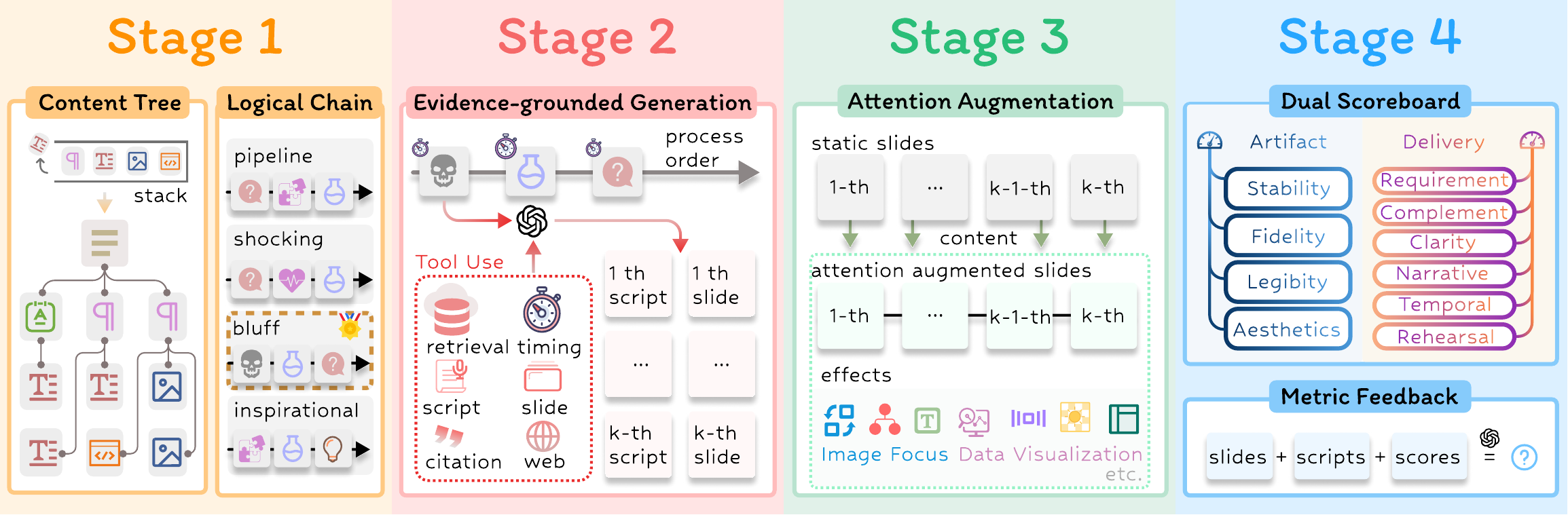}
  % \vspace{-4mm}
  \caption{
    Overview. 
      \emph{\textcolor{Stage1Bg}{Stage~1}:}
       requirement elicitation and narrative proposal;
      \emph{\textcolor{Stage2Bg}{Stage~2}:} logical chain editing and evidence-grounded generation;
      \emph{\textcolor{Stage3Bg}{Stage~3}:} interactive slide refinement and attention-oriented augmentation;
      \emph{\textcolor{Stage4Bg}{Stage~4}:} rehearsal and dual-scoreboard evaluation.
      Effects in Stage~3: 
      \imagefocus, \texttodiagram, \keynote, \dataviz, \motion, \bggg, \autolayout \,\,
    (Qwen3.5~\cite{qwen3.5} as example).}
  \label{fig:overview}
  % \vspace{-2mm}
\end{figure}

To address these gaps, we propose \textcolor{DeepSlideColor}{\textit{DeepSlide}}, a four-stage (Fig.~\ref{fig:overview}) end-to-end presentation agent that optimizes for full presentation delivery rather than static materials alone.
First, instead of producing a generic, fixed outline, it elicits audience-, time-, and goal-specific requirements through free-form dialogue and generates multiple time-budgeted narrative logical-chain candidates,  enabling users to select and refine an appropriate storytelling strategy tailored to the target audience before slide generation.
Second, it enables \emph{narrative-node-level} logical-chain editing---users can reorder/insert/remove/modify nodes, adjust per-node time budgets, and add non-linear cross-references---so the speaker explicitly controls the macro storyline and emphasis plan; each node can be expanded into one or more slides during generation.
Once the chain is finalized, \textit{DeepSlide} retrieves supporting evidence from the indexed source materials to ground subsequent slide generation for each narrative node.
Third, to better sustain audience attention during delivery, \textit{DeepSlide} augments static decks with content-aware, selectable attention-control tools (e.g., narrative-intent-driven figure zooming, table-to-visualization transformation, and text-to-diagram concretization), together with interactive refinement and one-click layout optimization.
Fourth, it goes beyond deck delivery by co-generating a synchronized, non-redundant speaker script aligned with the narrative nodes, supporting audience-perspective rehearsal (with optional audio), and providing evaluation with actionable revision suggestions and slide-level question simulation.

Together, these designs reduce end-to-end preparation workload by letting the speaker lock in high-level decisions (narrative skeleton, pacing/emphasis, and audience-conditioned style intent) while \textit{DeepSlide} handles evidence-grounded realization, delivery-time attention augmentation, and rehearsal-time feedback.

Current benchmarks~\cite{rw_3_aggarwal2025pass,rw_zheng-etal-2025-pptagent,rw_xu2025pregenieagent,rw_cachola2024knowledge,rw_slidesgen2025,rw_ofengenden2025pptarena,rw_huang2025pptbench} mainly evaluate the \emph{slide artifact} alone, leaving delivery-oriented qualities under-measured. Accordingly, we extend conventional static deck evaluation with a \emph{dual-scoreboard} benchmark consisting of an \textit{Artifact Scoreboard} and a \textit{Delivery Scoreboard}. The former aligns with prior practice and assesses the quality of generated slide materials, while the latter evaluates delivery-oriented qualities, including narrative coherence, pacing control, and audience-attention guidance.

\vspace{-2mm}
\paragraph{Contributions.}
In summary, we make the following contributions:
\vspace{-2mm}
\begin{itemize}[leftmargin=*]
  \item We present \textcolor{DeepSlideColor}{\textit{DeepSlide}}, a four-stage, human-in-the-loop multi-agent system for deliverable presentation preparation.
  \item We introduce a dual-scoreboard benchmark that disentangles artifact quality from delivery quality, enabling evaluation beyond static slides.
  \item We develop lightweight yet principled mechanisms for controllable and reliable generation, including tree-aware retrieval for grounding, time-budgeted logical-chain editing with pacing feedback, Markov-style sequential rendering with style inheritance, and browser-sandbox validation for robust rendering.
\item Extensive evaluations across 20 research domains and 5 audience profiles show that \textit{DeepSlide} matches strong baselines on the Artifact Scoreboard while substantially improving the Delivery Scoreboard, reducing both preparation overhead and live-speaker effort.
\end{itemize}

\section{\textit{DeepSlide}}
\label{sec:deepslide}
\subsection{Overview}
\textit{DeepSlide} supports multi-source inputs in diverse formats and optimizes for the full delivery process. 
After materials are uploaded, it follows a four-stage pipeline.
\paragraph{Stage~1: Requirement elicitation and narrative proposal.}
\textit{DeepSlide} conducts free dialogue (text or speech) to elicit presentation requirements, including the target audience, total duration, key emphasis, and preferred style,
while building a lightweight \emph{content-tree index} over the source materials for later evidence retrieval.
It then produces four time-budgeted \emph{narrative logic chain} candidates, each distributing the total duration in a well-balanced manner.

\paragraph{Stage~2: Logical chain editing and evidence-grounded generation.}
A logical chain is an ordered sequence of narrative nodes that abstracts the talk progression.
DeepSlide supports logical chain editing (reorder/insert/remove/modify nodes, and adjust duration) and \emph{non-linear narrative enhancement} by creating cross-references to 
serve as a connecting link between the preceding and the following.
Once finalized, the system follows the chain, retrieves supporting evidence blocks from the content-tree index, and generates an audience-friendly, time-consistent slide along with a synchronized script aligned, where the two are designed to stay on-topic while minimizing redundancy.

\paragraph{Stage~3: Interactive slide refinement and attention-oriented augmentation.}
Users can refine slides and scripts via dialogue or speech with instant preview.
\textit{DeepSlide} provides one-click layout/typography optimization and mutiple freely selectable AI tools for delivery-time \emph{attention control}: 
It identifies corresponding regions in complex figures based on the narrative intent and automatically zooms in on the corresponding areas.
converts static tables into interactive visualizations; and turns verbose text into diagrams to improve readability and reduce cognitive load.
It also supports text keynote, animations, and audio preview by extracting a speaker voice from conversational speech and synthesizing narration via TTS model~\cite{indextts}.

\paragraph{Stage~4: Rehearsal and dual-scoreboard evaluation.}
\textit{DeepSlide} supports audience-perspective rehearsal (with optional rehearsal audio). 
It further provides a dual-scoreboard evaluation that assesses artifact quality and delivery quality, and computes scores and revision suggestions grounded in the evaluation results, source materials, and user requirements. 
To further reduce preparation burden, it can simulate likely audience questions. 
Finally, DeepSlide enables one-click packaging/export.

\subsection{Stage~1: Requirement Elicitation and Narrative Proposal}
\begin{figure}[htbp]
% \small
  \centering
  % \vspace{-2mm}
  \includegraphics[width=\linewidth]{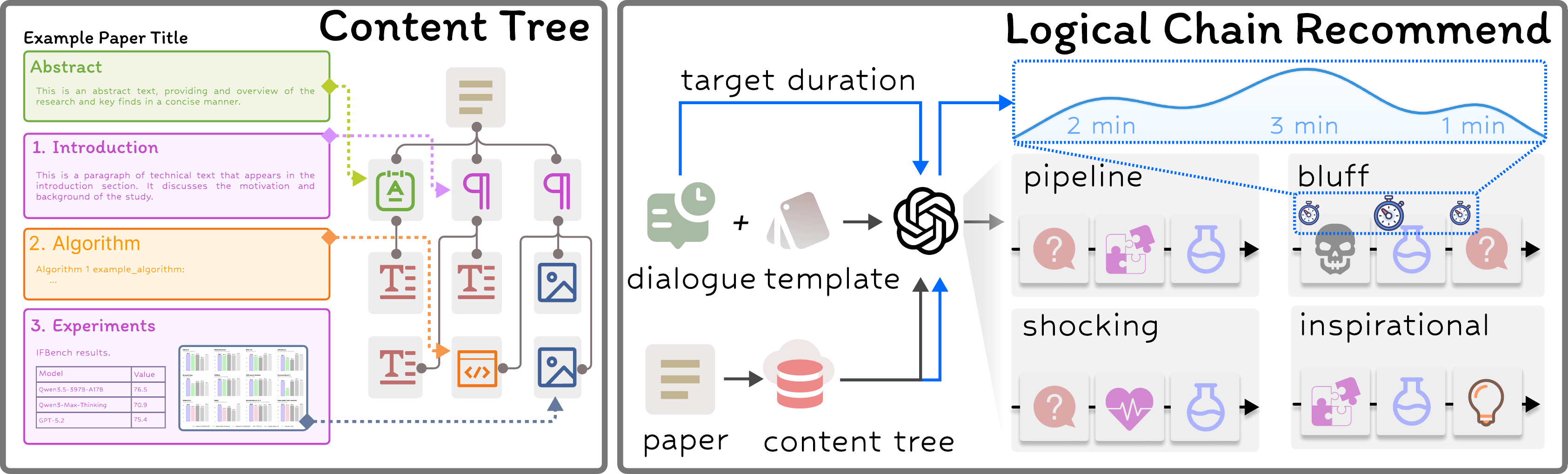}
  % \vspace{-4mm}
  \caption{Requirement elicitation and narrative proposal (Stage~1).}
  \label{fig:stage1}
  % \vspace{-2mm}
\end{figure}
This stage aligns the user's presentation intent with the paper content and, under a time budget, proposes candidate narrative modes as a stable scaffold for slide generation. Guided by system-prompt constraints, the agent elicits key requirements (audience, duration, focus, and style) and uses paper-aware follow-up questions to converge to a complete presentation blueprint through multi-turn dialogue. The dialogue yields two artifacts: a content tree and a set of candidate logical chains,
which is shown in Fig~\ref{fig:stage1}.

\paragraph{Content tree.}
Naively passing the full paper to an LLM on every generation step is both token-wasteful and imprecise: large undifferentiated text blocks introduce irrelevant context that degrades generation quality.
A vector-database pipeline avoids this but is unnecessarily heavy for single-document parsing.
The content tree occupies the middle ground: a typed, hierarchical index built directly from the source, enabling targeted retrieval without embeddings or approximate-search infrastructure.
Construction proceeds in three steps.
\textbf{\emph{(i) Source normalization.}}
Mixed uploads (\LaTeX, Markdown, Word, PDF, or nested archives) are unified into \LaTeX{} via \texttt{pandoc}; the root file is located and all \verb|\input|/\verb|\include| directives are recursively flattened into a single stream.
\textbf{\emph{(ii) Lossless segmentation.}}
Structural headings partition the stream into typed slices; artifact slices (\texttt{type}$\in$\{\texttt{equation},\texttt{figure},\texttt{table},\texttt{theorem},\texttt{algorithm}\}) are never merged with adjacent text, preserving them as atomic retrieval units.
Each slice is assigned an LLM-generated \texttt{abstract} summarizing its own content.
\textbf{\emph{(iii) Hierarchical index construction.}}
Each slice becomes a \texttt{ContentTreeNode} $\langle\textit{id},\textit{title},\textit{level},\textit{type},\textit{abstract},\textit{content}\rangle$, where \textit{level} follows the document hierarchy ($\ell{=}1$ for \verb|\section|, $\ell{=}2$ for \verb|\subsection|, etc.).
The slices are retained in document order, forming the flat sequence $\mathcal{N}=\langle n_1,\dots,n_m\rangle$; a monotonic stack $S$ (Alg.~\ref{alg:stack_tree_min}) then links them into a tree in a single $O(m)$ pass: when processing $n_i$, all nodes with $\ell(S\text{.top})\ge\ell(n_i)$ are popped; the remaining top is $n_i$'s parent (or $n_i$ becomes a root if $S$ is empty), then $n_i$ is pushed.
$S$ thus always holds the unique ancestor chain from the current root to the deepest open section, correctly nesting each node under its nearest enclosing heading.
The resulting tree mirrors the paper's logical outline and supports context-aware BM25 retrieval: each node's relevance score is jointly influenced by its own content, its children's scores, and its parent's score, enabling multi-granularity access from section-level overviews down to fine-grained leaf nodes.

\begin{algorithm}[t]
\small
\caption{Hierarchical content tree construction.}
\label{alg:stack_tree_min}
\begin{algorithmic}[1]
\Require Ordered nodes $\mathcal{N}=\langle n_1,\dots,n_m\rangle$ with integer level $\ell(n)$
\Ensure Root list $\mathcal{R}$ and parents/children for all nodes

\State $\mathcal{R}\gets\emptyset$, $S\gets\emptyset$ 
\Comment{monotonic stack increasing levels}
\For{$i\gets 1$ to $m$}
  \State $n \gets n_i$
  \While{$S\neq\emptyset \land \ell(\mathrm{top}(S)) \ge \ell(n)$}
    $\mathrm{pop}(S)$
  \EndWhile
  \If{$S=\emptyset$}
    \ append $n$ to $\mathcal{R}$ 
    % \Comment{new root}
  \Else
    \ $p \gets \mathrm{top}(S)$, $n.parent \gets p$, $p.children \gets p.children \cup \{n\}$
    % \Comment{set $n.parent \gets p$ and add $n$ to $p.children$}
  \EndIf
  \State $\mathrm{push}(S,n)$
\EndFor
\State \Return $\mathcal{R}$
\end{algorithmic}
\end{algorithm}

\paragraph{Logical chain recommendation.}
After the content tree is built, DeepSlide generates four candidate logical chains, each embodying a distinct narrative style, via two complementary pipelines.
\emph{(i) Template-guided think--retrieve generation.}
The agent first selects four narrative templates from a template library conditioned on the multi-turn dialogue history and the paper abstract.
For each template, it executes a think--retrieve loop: the agent retrieves the paper outline and BM25-ranked evidence from the content tree, then synthesizes the retrieved content with the dialogue context and template constraints to produce a personalized chain.
\emph{(ii) Duration-aware time allocation.}
The agent parses the target talk duration from the dialogue history and retrieves section-level evidence to estimate relative node importance.
It then distributes the total time across nodes proportionally to their \texttt{importance ratios}.
% enforcing that every node receives at least one minute and that all allocations sum exactly to the target duration.

\subsection{Logical Chain Editing and Evidence-grounded Generation (Stage~2).}
\begin{figure}[htbp]
% \small
  \centering
  % \vspace{-2mm}
  \includegraphics[width=\linewidth]{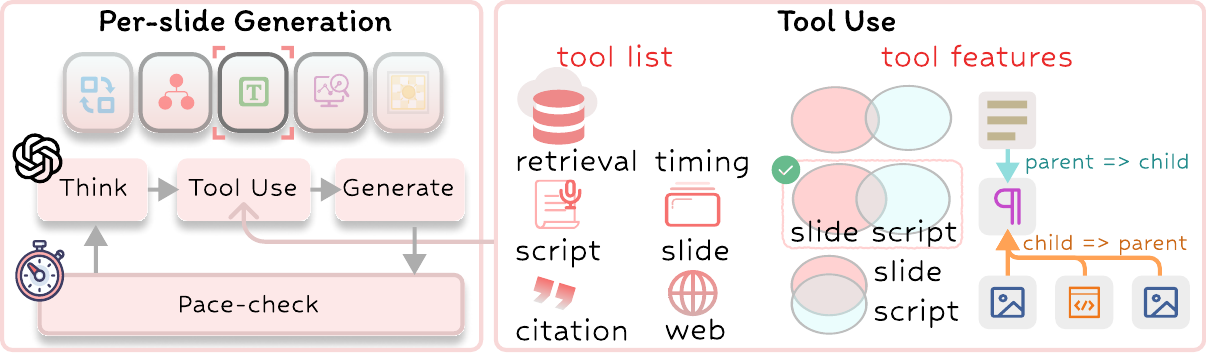}
  % \vspace{-4mm}
  \caption{Logical chain editing and evidence-grounded generation (Stage~2).}
  \label{fig:stage2}
  % \vspace{-2mm}
\end{figure}

Stage~2 supports editing of the selected logical chain,
and evidence-grounded slide generation (Fig~\ref{fig:stage2}).

\subsubsection{Logical Chain Editing}
Stage~2 enables users to refine the narrative plan before slide generation by editing the logical chain.
Users can reorder, insert, remove, or modify logical nodes (e.g., ``add an ablation slide'' or ``move limitations earlier''), and adjust per-node time budgets to match the target duration.
% 当用户完成编辑后可点击进行Slide生成，系统将进行逻辑链检索与Slide&Script生成。

\subsubsection{Content Tree Retrieval}
\paragraph{Okapi-BM25 index.}
To support fast/precise retrieval, we overlay a lightweight Okapi-BM25 index on Stage~1 content tree.
We strip \LaTeX{} commands, truncate overly long spans by hyperparameter $l_{\mathrm{max}}$, and compute corpus statistics once, including document length \(\mathrm{dl}(n)\), average length \(\mathrm{avgdl}\), and term document frequency \(\mathrm{df}(t)\).
% Specifically, we truncate nodes longer than \(l_{\mathrm{max}}\) tokens to curb retrieval cost and prevent long spans from skewing \(\mathrm{avgdl}\) and term weights (ablation in Evaluation).

\paragraph{Tree-aware scoring.}
Given a query \(q\), we first tokenize it into a multiset of terms \(Q=\mathrm{tokenize}(q)\) and compute a base Okapi-BM25 score \(s_0(n)\) for every node-document \(n\in\mathcal{N}\).
In our implementation, BM25 is computed with standard length normalization (Alg.~\ref{alg:tree_bm25}): for each query term \(t\in Q\), we form \(\mathrm{idf}(t)=\log\!\big(1+\frac{|\mathcal{N}|-\mathrm{df}(t)+0.5}{\mathrm{df}(t)+0.5}\big)\) and accumulate a saturated term-frequency contribution
\(
s_0(n)=\sum_{t\in Q}\mathrm{idf}(t)\cdot \frac{\mathrm{tf}(t,n)\cdot (k_1+1)}{\mathrm{tf}(t,n)+k_1\cdot(1-b+b\cdot \mathrm{dl}(n)/\mathrm{avgdl})},
\)
where \(\mathrm{tf}(t,n)\) is the within-node term frequency and \(\mathrm{dl}(n)\) is the node-document length.
We then inject structural priors from the content tree in two complementary directions.
First, \emph{child-to-parent promotion} makes a section-level node competitive when its descendants contain the most relevant evidence; second, \emph{parent-to-child promotion} propagates topic-level relevance downward so that fine-grained nodes are not overly penalized when the query matches a broader heading.
Concretely, we define the tree-aware score
\begin{equation}
s(n)=s_0(n)
+{\color{ChildToParent}\underbrace{\alpha_{\mathrm{tree}}\sum_{c\in \mathrm{children}(n)} s_0(c)}_{\mathrm{children\to parent}}}
+{\color{ParentToChild}\underbrace{\beta_{\mathrm{tree}} s_0(\mathrm{parent}(n))\vphantom{\sum_{c\in \mathrm{children}(n)} s_0(c)}}_{\mathrm{parent\to children}}}.
\end{equation}
with \(s_0(\mathrm{parent}(n))=0\) for roots. 
% In practice, we set \((\alpha_{\mathrm{tree}},\beta_{\mathrm{tree}})=(0.4,0.2)\).
Finally, we apply a query-granularity depth bias to align retrieval granularity with user intent: if \(|\mathcal{Q}|\le m_0\) (overview queries), we downweight deep nodes by \(\frac{1}{1+\gamma_{\mathrm{tree}}\cdot \mathrm{depth}(n)}\); otherwise (detail queries), we upweight deep nodes by \((1+\delta_{\mathrm{tree}}\cdot \mathrm{depth}(n))\).
%  We use \(m_0=4\), \(\gamma_{\mathrm{tree}}=0.25\), and \(\delta_{\mathrm{tree}}=0.1\) in the system.
The total query-time cost is dominated by computing \(s_0(n)\) over all nodes and aggregating child scores over edges. With precomputed term statistics, the scoring is \(O(|\mathcal{N}|\cdot|\mathcal{Q}|+|\mathcal{E}|)\).

\begin{algorithm}[t]
\small
\caption{Tree-aware BM25 retrieval over a content tree.}
\label{alg:tree_bm25}
\begin{algorithmic}[1]
\Require 
Nodes $\mathcal{N}$,
corpus stats $\mathrm{df}/\mathrm{dl}/\mathrm{avgdl}$,
fusion params $(\alpha_{\mathrm{tree}},\beta_{\mathrm{tree}})$,
depth-bias params $(\gamma_{\mathrm{tree}},\delta_{\mathrm{tree}})$, 
threshold $m_0$, 
query $q$, 
top-$K$
\Ensure Ranked nodes $R$

\State $\mathcal{Q} \gets \mathrm{tokenize}(q)$,\quad $m \gets |Q|$
\ForAll{$t \in \mathcal{Q}$}
  \ $\mathrm{idf}(t) \gets \log\!\Big(1+\frac{|\mathcal{N}|-\mathrm{df}(t)+0.5}{\mathrm{df}(t)+0.5}\Big)$
\EndFor

\ForAll{$n\in\mathcal{N}$}
  \ $s_0(n) \gets \textsc{BM25}(n,\mathcal{Q},\mathrm{idf},
  \mathrm{dl},\mathrm{avgdl})$
\EndFor

\ForAll{$n\in\mathcal{N}$}
\Comment{$s_0(\mathrm{parent}(n))=0$ if root}
  \State $s(n) \gets s_0(n)\;+\;\alpha_{\mathrm{tree}}\!\cdot\!\sum_{c\in \mathrm{children}(n)} s_0(c)\;+\;\beta_{\mathrm{tree}}\!\cdot\! s_0(\mathrm{parent}(n))$ 
  \If{$m \le m_0$}
    \ $s(n) \gets s(n)\cdot \frac{1}{1+\gamma_{\mathrm{tree}}\cdot \mathrm{depth}(n)}$
  \Else
    \ $s(n) \gets s(n)\cdot (1+\delta_{\mathrm{tree}}\cdot \mathrm{depth}(n))$
  \EndIf
\EndFor
\State \Return $\mathrm{topK}(\{(n,s(n)):\, s(n)>0\},K)$
\end{algorithmic}
\end{algorithm}

\subsubsection{Evidence-grounded Generation}
\textit{DeepSlide} instantiates a tool-augmented generation agent to convert each logical-chain node into a short segment of slides and an aligned speaker script under a strict per-node time budget.
Rather than a single-pass ``retrieve once, then write'' workflow (e.g., Gamma~\cite{Gamma-web}, NoteBookLM~\cite{notebooklm-web}), 
the agent follows a multi-turn loop: it iteratively retrieves evidence, drafts a candidate slide, checks pacing, and then decides whether to refine, expand, or stop.
This closed-loop design is critical for producing delivery-ready artifacts that remain grounded while being sensitive to time and narrative constraints.

\paragraph{Tool-call driven multi-round reasoning.}
For a target logical node with a topic name and description, the agent is equipped with a small set of retrieval and creation tools, and can call them multiple times in a single generation of a logical node:
\begin{itemize}[leftmargin=*]
  \item \textit{Tree retrieval and reading.} The agent first retrieves top-$K$ relevant content-tree nodes w.r.t its thinking using the tree-aware scoring in Alg.~\ref{alg:tree_bm25}. It then reads the full content spans of the selected nodes to collect grounded evidence before writing.
  \item \textit{Complementarity between slide and script.} After interpreting the evidence, the agent can append (i) a new slide frame, (ii) its aligned speaker script, and (iii) optional bibliography entries. Slide content and spoken script are designed to be \emph{highly related but not redundant}: slides keep key points and visual anchors, while details is delegated to the script.
  \item \textit{Pacing control.} A countdown timer w.r.t. a logical node is continuously updated and, after each slide is generated, produces immediate feedback based on the estimated script length to regulate the remaining generation granularity under the time budget. When ample time remains, the timer stays silent; once the remaining time drops below half, it issues a cautionary prompt; when the budget is exhausted, it instructs the agent to aggressively compress and finish; and if a preset overrun threshold is exceeded, it forcibly terminates further slide generation.
  \item \textit{Web search.} When a logical node requires background that is not present in the uploaded document, the system can optionally query the web to fetch supporting evidence.
\end{itemize}
% Within one logic node, the agent typically alternates between retrieval (to resolve what to say) and structured writing (to decide what to show on slides).

\subsubsection{Non-linear Narrative Enhancement and Compilation}
Beyond linear ordering, DeepSlide also supports \emph{non-linear narrative enhancement}:
the agent may automatically create cross-references between nodes to bridge the preceding and following context, improving transitions and global coherence (e.g., foreshadowing an upcoming result or recall a prior definition).
A compilation agent renders the slides under system prompts, automatically detects build failures, and performs sandboxed debugging in a Docker container via iterative read/write and code-execution tools.

\subsection{Stage~3: Interactive slide refinement and attention-oriented augmentation}
Stage~3 turns the static outputs from Stage~2 into a delivery-ready, interactive deck.
Let the deck be a slide sequence \(\{f^{\mathrm{src}}_k\}_{k=1}^{m}\).
Stage~3 produces an augmented dynamic sequence \(\{f^{\mathrm{dst}}_k\}_{k=1}^{m}\), 
while supporting refinement via text/voice dialogue (see Fig~\ref{fig:stage3}).

\subsubsection{Dialogue Refinement}
\textit{DeepSlide} supports user-in-the-loop refinement at both the slide and script levels.
A user refinement instruction can be given via text or speech (transcribed by ASR).
We use a planner--executor manner to translate complex instruction into a few edit actions and apply minimal, 
localized updates to the affected parts.
% To ensure reliability, we keep an explicit page-to-slide alignment so edits are routed deterministically, and reuse this alignment for both dialogue editing and deck augmentation.

\subsubsection{Attention Augmentation}
% The core goal is to convert \(f_{\mathrm{src}}\) into \(f_{\mathrm{dst}}\) under two simultaneous optimization targets: 
% \textcolor{blue}{
The core goal is to convert the source sequence $\{f_k^{\mathrm{src}}\}_{k=1}^m$ into an augmented sequence $\{f_k^{\mathrm{dst}}\}_{k=1}^m$ under two simultaneous optimization targets:
% }
\emph{(i) stability/fidelity}, preserving the original content and structure, and 
\emph{(ii) attention gain}, introducing controlled dynamic effects that improve audience attention and delivery clarity.
Unlike template-based PPT agents that hard-code stability by forcing slides into fixed layouts, 
DeepSlide adopts a two-step augmentation framework.
% First, we generate slides in order via a Markov-style sequential control, where agent autonomously executes function calls to enhance focused attention based on pre-defined strategies;
% \textcolor{blue}{
First, we augment slides in delivery order via Markov-style sequential control, where the agent triggers deterministic effect functions to improve focused attention under predefined strategies;
% }
Second, since template-free rendering may be unstable, we validate the generated deck in a sandboxed browser 
and feed back runtime errors to revise the plans, ensuring successful rendering finally.
This encourages a consistent style while maximizing the dynamic/modern rendering of an agent-based template-free approach.
\begin{figure}[htbp]
% \small
  \centering
  % \vspace{-2mm}
  \includegraphics[width=\linewidth]{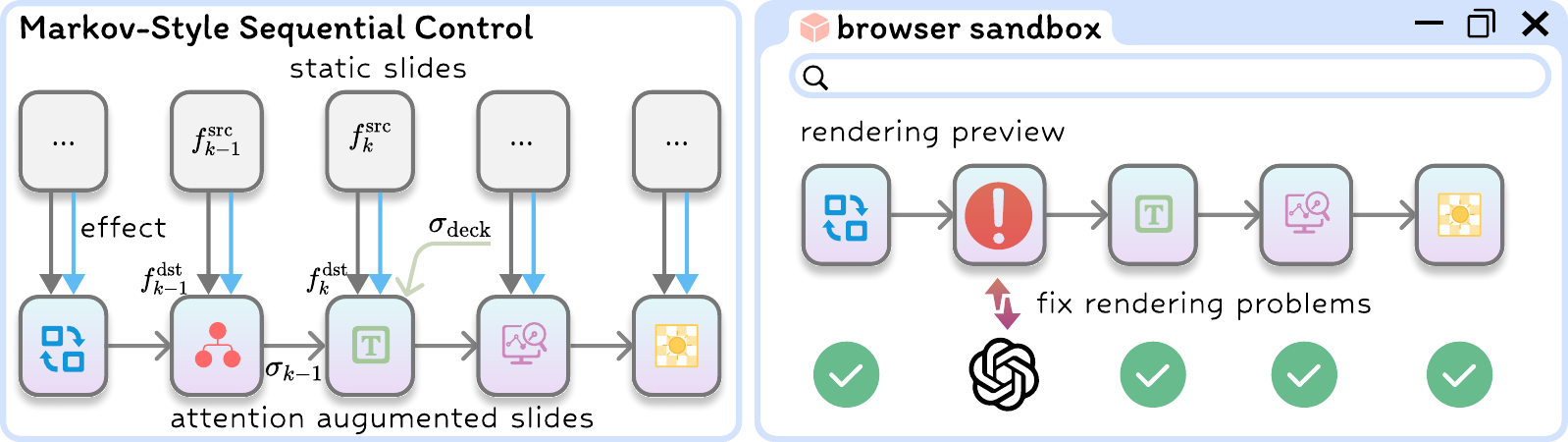}
  % \vspace{-4mm}
  \caption{
  Interactive slide refinement and attention-oriented augmentation (Stage~3), where \(f_k^{\mathrm{src}}\) and \(f_k^{\mathrm{dst}}\) are the original and augmented slides, \(\sigma_{\mathrm{deck}}\) is the Stage~1 deck style, and \(\sigma_{k-1}\) is the previous-slide style.
  }
  \label{fig:stage3}
  % \vspace{-2mm}
\end{figure}

\paragraph{Markov sequential control.}
% We generate \(f^{\mathrm{dst}}\) slide by slide.
As shown in Fig~\ref{fig:stage3}, for slide \(f_k\), an LLM-based planner conditions on its source content \(f_k^{\mathrm{src}}\), 
a user requirement profile \(\mathcal{U}\), 
a deck-level style summary \(\sigma_{\mathrm{deck}}\), 
and a compact summary of the previously generated slide \(\sigma_{k-1}\), 
yielding a  Markov-style(first-order) dependency:
\(
s_k=(f_k^{\mathrm{src}},\mathcal{U},\sigma_{\mathrm{deck}},\sigma_{k-1}),\quad
f^{\mathrm{dst}}_k = T\!\big(f_k^{\mathrm{src}},\,\pi(s_k)\big),
\)
where \(\pi\) denotes the LLM that maps the state \(s_k\) to a structured augmentation decision
for slide \(k\) (e.g., which effects to apply and how to stage them), 
and \(T\) is a deterministic renderer that executes the decision to produce \(f^{\mathrm{dst}}_k\).
The overall process is shown in Alg~\ref{alg:markov_seq}, where the term \(\sigma_{k-1}\) enforces cross-slide style continuity (instead of any template), while \(f_k\) and \(\mathcal{U}\) enable per-slide, content-conditioned customization under validity constraints.

\begin{algorithm}[t]
\caption{Markov-style sequential generation with style inheritance.}
\label{alg:markov_seq}
\begin{algorithmic}[1]
\Require Source slides $\{f_k^{\mathrm{src}}\}_{k=1}^m$, requirement profile $\mathcal{U}$, deck style summary $\sigma_{\mathrm{deck}}$, planner $\pi$, renderer $T$, style summarizer $S$
\Ensure Augmented slides $\{f_k^{\mathrm{dst}}\}_{k=1}^m$
\State $\sigma_0 \gets \sigma_{\mathrm{deck}}$ 
% \Comment{initialize previous-slide style summary}
\For{$k \gets 1,\dots,m$} 
% \Comment{generate in delivery order}
  \State $s_k \gets \big(f_k^{\mathrm{src}},\,\mathcal{U},\,\sigma_{\mathrm{deck}},\,\sigma_{k-1}\big)$
  , $a_k \gets \pi(s_k)$ 
  , $f_k^{\mathrm{dst}} \gets T\!\left(f_k^{\mathrm{src}}, a_k\right)$ 
  % \Comment{deterministic rendering (layout/style/effects)}
  \State $\sigma_k \gets $
  summarize style description for next Markov state
\EndFor
\State \Return $\{f_k^{\mathrm{dst}}\}_{k=1}^m$ 
\end{algorithmic}
\end{algorithm}
\vspace{-2mm}

\paragraph{Attention strategies.}
DeepSlide represents attention enhancement as freely selectable effect set \(\mathcal{E}\) encoded in the Stage~3 and rendered deterministically.
We separate effects into four classes:

\begin{itemize}[leftmargin=*]
  \item \textit{Image focus.} This feature enables localized focus on complex figures during delivery.
  Rather than allowing arbitrary boxes (current VLMs struggle to identify focused regions), 
  the system selects a focus template from a small predefined set (e.g., left/right split, \(2\times2\) grid), which deterministically maps to normalized regions of interest (ROIs).
  The renderer then generates clickable ROI tiles with a lightbox interaction to guide attention without modifying the original figure.

  \item \textit{Text to diagram.} For slides with convoluted and verbose text, we switch to a diagram-style layout to improve delivery clarity.
  We use a LLM to draft a diagram design specification from \(f^{\mathrm{src}}_k\)'s text and then invoke an open-source diagram generator (e.g., next-ai-draw-io~\cite{nextaidrawio}) to render it.

  \item \textit{Data Visualization.} When \(f^{\mathrm{src}}_k\) contains tabular data, the system can detect it and generate a table-centric design, which is then rendered as interactive visualizations using open-source frontend libraries such as ECharts~\cite{echarts} with matching form.
  
  \item \textit{Other Effects.} We further provide lightweight attention cues that add no new semantic content.
  For example, \emph{Text Keynote} highlights a few critical phrases 
  (e.g., key numbers ``3$\times$'').
  \emph{Auto Layout} restructures dense paragraphs into clearer visual organizations (e.g., step-wise process blocks or card-style summaries).
  \emph{Motion} introduces subtle entry animations to guide reading order, 
  and \emph{Background} non-intrusively adds a content-aware, \textit{bento}-style backdrop that improves visual coherence without occluding the main text.
\end{itemize}

\paragraph{Effect conflict and gating.}
To prevent conflicts between effects, we apply a deterministic gating rule before planning:
each slide can have \emph{at most one primary visual effect} from \(\{\)Image Focus, Table Viz, Text to Diagram\(\}\),
with priority \(\text{Image Focus} \succ \text{Table Visualization} \succ \text{Text to Diagram}\).
The effect is selected if and only if the effect set $\mathcal{E}$ includes ``structural recognition'' (e.g., figure/tabular) and the priority conditions are satisfied.

\paragraph{Sandbox-guided rendering stabilization.}
Since template-free augmentation can occasionally render poorly, 
\textit{DeepSlide} executes each generated slide in a lightweight browser sandbox 
to detect layout and runtime failures.
Detected issues are fed back for a minimal repair.

\subsubsection{Audio Rehearsal with User Voice.}
\textit{DeepSlide} provides a per-slide speech audio preview by synthesizing $p_k$ into a waveform via TTS~\cite{indextts}. The user’s voice profile is derived from both a audio library and real-time voiceprint captured during previous dialogues, which enables users to review their presentation from an audience’s perspective.

\subsection{Stage~4: Rehearsal and dual-scoreboard evaluation}
\label{sec:stage4}
Stage~4 turns a generated deck into a deliverable talk by providing rehearsal guidance (including audio preview generated in Stage~3) and post-hoc evaluation.
The dual-scoreboard protocol is formalized in Section~\ref{sec:dual-scoreboard}; here we focus on how \textit{DeepSlide} uses \emph{slides + scripts + evaluation signals} to give concrete revision suggestions and to simulate likely audience questions.

For each slide, we assemble a compact context 
\(\langle\) \texttt{slide}, \texttt{script}, \texttt{per-slide metrics}, \texttt{estimated time comsume} \(\rangle\) 
% \(\langle\)slide text, script text, per-slide metrics, deck-level timing summary\(\rangle\) 
and prompt an LLM to produce:
(i) 3--6 short rehearsal tips that are specific to the slide (e.g., where to slow down, what to cut, what to emphasize), and
(ii) the top-3 likely audience questions for that slide, biased toward metrics-indicated risks (e.g., unclear assumptions, missing baselines, or abrupt topic jumps).
These support an iterative rehearsal: speakers rehearse with the audio preview, inspect advice/questions, and revise slides/scripts before re-running the diagnostics.

\begin{example}
  Consider a slide titled \emph{``Ablation Study''} without any \textit{DeepSlide}'s augmentation whose body contains a dense table and long bullet list, while its script repeats most on-slide text.
DeepSlide estimates a \(75\)s narration for a \(45\)s time budget and detects high slide--script similarity, leading to the following coaching outputs:
\textbf{(i) Suggestions:} ``Remove two minor ablations''; ``State one takeaway before details''; ``Move table reading to appendix''; ``Explain the strongest baseline first''.
\textbf{(ii) Likely audience questions:} ``Which component contributes most to the gain?''; ``Are improvements statistically significant?''; ``Why is method X not included as a baseline?''.
% The speaker can then shorten \(p_k\), simplify \(f_k\) (e.g., keep only key rows and highlight the best numbers), and add a one-sentence transition to connect with adjacent slides; re-running Stage~4 typically improves pacing and complementarity without changing the core content.
\end{example}

\subsection{Multi-Agents}
\label{sec:multi-agent}

\textit{DeepSlide} employs a role-based multi-agent pipeline. Rather than unconstrained generation, each agent performs a bounded task, e.g., requirement normalization, logical planning, or rehearsal coaching, and updates shared structured artifacts. This decomposition enhances controllability, enabling stage-wise verification and localized repair. We treat each system-prompted role as a distinct agent, regardless of the underlying model. All agents in used are in Table~\ref{tab:agent_inventory}.

% preamble:
% \usepackage[table]{xcolor}

% \input{tab_multi-agents}

\section{Dual-Scoreboard Evaluation}
\label{sec:dual-scoreboard}
% In this work, we evaluate presentation agents with a \emph{dual-scoreboard} protocol that disentangles the quality of static deliverables from the quality of end-to-end talk delivery. The key intuition is that visually strong slides do not necessarily yield a presentation that is friendly to both presenters and audiences. A delivery-ready system should satisfy audience and duration requirements, maintain a coherent (often nonlinear) narrative, and coordinate slides with an accompanying script so that the speaker can explain rather than duplicate on-slide text.

% Each benchmark instance consists of a source document $\mathcal{T}$ and a requirement profile $\mathcal{U}$, which specifies the target audience, the intended talk duration, and any additional constraints. 
% Given $(\mathcal{T}, \mathcal{U})$, a system produces a presentation package comprising slides $\mathcal{F}=\{f_k\}_{k=1}^m$ and an aligned script $\mathcal{P}=\{p_k\}_{k=1}^m$. Optionally, the system may also output (i) a dynamic delivery artifact, which we convert to a static counterpart to ensure fair comparison across systems, and (ii) auxiliary materials such as narrative candidates and per-slide time budgets. For each instance, we run the system $\tau=5$ times to improve the stability of our evaluation.

We evaluate agents using a dual-scoreboard protocol that decouples static slide quality from end-to-end delivery performance. For each instance consisting of a source document $\mathcal{T}$ and requirement profile $\mathcal{U}$ (e.g., audience and duration), systems generate a package of slides $\mathcal{F}=\{f_k\}_{k=1}^m$ and aligned scripts $\mathcal{P}=\{p_k\}_{k=1}^m$, with $\tau=5$ runs to ensure stability. The Artifact Scoreboard assesses the visual and content quality of the slides, while the Delivery Scoreboard measures narrative coherence and slide-script coordination, ensuring the system explains rather than merely duplicates on-slide text to meet audience-specific constraints.

\subsection{Artifact Scoreboard}
% Artifact scoreboard measures traditional artifact quality 
% (slides+script) and is designed to remain fair to strong 
% static slide systems.
The Artifact Scoreboard evaluates systems across four dimensions: Stability, Fidelity, Legibility, and Aesthetics.

\paragraph{Stability.}
We measure stability by the success rate $P=\frac{\#\{r\in\{1,\dots,\tau\}: \,\text{run }r\text{ succeeds}\}}{\tau}$,
where a run is counted as successful if it produces a usable deck with complete
slide--script alignment.

\paragraph{Fidelity.}
We quantify how faithfully the generated presentation is grounded in the source. For each slide $k$, we align the slide--script pair $(f_k,p_k)$ to the most relevant source chunk(s) in $\mathcal{T}$ via retrieval, and denote the resulting reference text by $t_k$. Textual fidelity $F_t$ is a weighted combination of a lexical-overlap metric (ROUGE) and a semantic-similarity metric (BERTScore), averaged over both slide text and the aligned script:
{
% \small
\begin{equation}
\label{eq:fidelity_text}
F_t
=
\frac{\omega_{\mathrm{rouge}}}{2m}\sum_{k=1}^m
\Big(\mathrm{ROUGE}(f_k,t_k)+\mathrm{ROUGE}(p_k,t_k)\Big)
+
\frac{\omega_{\mathrm{bert}}}{2m}\sum_{k=1}^m
\Big(\mathrm{BERT}(f_k,t_k)+\mathrm{BERT}(p_k,t_k)\Big),
\end{equation}
}
where $\omega_{\mathrm{rouge}}$ and $\omega_{\mathrm{bert}}$ control the contribution of the two metrics ($\omega_{\mathrm{rouge}}+\omega_{\mathrm{bert}}=1$). Visual fidelity $F_v$ is defined as the fraction of slides whose visual items are supported by the aligned source:
\begin{equation}
\label{eq:fidelity_vis}
F_v=\frac{1}{m}\sum_{k=1}^m \mathbb{I}\left\{\text{slide }f_k\text{ contains a visual element matching }t_k\right\},
\end{equation}
where $\mathbb{I}\{\cdot\}$ is the indicator function that equals 1 if the condition is satisfied and 0 otherwise.

\paragraph{Legibility and aesthetics.}
We compute an automatic legibility score $L\in[0,1]$ 
from slide-level readability checks. 
For each slide $f_k$, we extract (i) the minimum fontsize, (ii) the number of words in on-slide text. We define a per-slide
\(
\mathrm{penalty}_k
=
\mathbb{I}\!\left[\mathrm{minfont}(f_k) < 12\right]
+
\mathbb{I}\!\left[\mathrm{wc}(f_k) > 140\right],
\)
and aggregate the legibility score
\begin{equation}
\label{eq:legibility}
  L = 1-\frac{1}{m}\sum_{k=1}^m \min(1, \mathrm{penalty}_k),
\end{equation}
Since purely subjective aesthetics is hard to standardize, we use an automatic Vision-Language Model (VLM) generate $Ae\in[0,1]$ that favors readable decks with balanced content. 
We define the aesthetic score as
\begin{equation}
  Ae=\mathrm{clip}\!\left(0,1,\;
0.6L
+0.2\Big(1-\big|\mathrm{frac}_{\mathrm{img}}-0.6\big|\Big)
+0.2\,\mathrm{frac}_{\mathrm{script}}
\right),
\end{equation}
where $\mathrm{frac}_{\mathrm{img}},\mathrm{frac}_{\mathrm{script}}\in[0,1]$ are the fraction of slides that contain at least one image and
the fraction of slides with speaker scripts. 
The weights of 0.6 and 0.2 are empirical hyperparameters.

\paragraph{Aggregation.}
We report the Artifact score as:
\begin{equation}
  S_A=
\alpha_{\mathrm{stab}}\cdot P
\;+\;\alpha_{\mathrm{fid}}\cdot\big(\beta F_t+(1-\beta)F_v\big)
\;+\;\alpha_{\mathrm{read}}\cdot\big(\gamma L+(1-\gamma)Ae\big),
\end{equation}
where $\alpha_{\mathrm{stab}}+\alpha_{\mathrm{fid}}+\alpha_{\mathrm{read}}=1$ and $\beta,\gamma\in[0,1]$.

\subsection{Delivery Scoreboard}
Delivery scoreboard measures end-to-end delivery quality beyond static artifacts,
mainly focusing on: (i) satisfying requirements and audience constraints, (ii) narrative control, (iii) slide--script complementarity, (iv) temporal pacing, (v) delivery-time guidance, and (vi) rehearsal support.

\paragraph{Requirement satisfaction.}
$R$ evaluates whether a system's output satisfies the requirement profile $\mathcal{U}$.
We decompose $R$ into (i) a fully automatic \emph{timing-compliance} score and (ii) \emph{content-level} requirement checks.

\begin{itemize}[leftmargin=*]
  \item \textit{Timing compliance.} We compute the timing score from the implied seconds-per-slide $s$ and use a piecewise-linear schedule with a \emph{sweet} range ($12\!\le\! s\!\le\!60$ sec/slide) and a broader \emph{hard} tolerance range ($6\!<\! s\!<\!120$ sec/slide); values outside the hard range receive zero score:
  \begin{equation}
    \label{eq:r_time}
    % \small
    R_{\mathrm{time}}
    =
    \min\Big\{
    \underbrace{\vphantom{\min\big(\frac{s-6}{6},\frac{120-s}{60}\big)}1}_{\textcolor{teal}{\mathrm{sweet}:\; 12\le s\le 60}},
    \;
    \max\Big\{
      \underbrace{\min\Big(\frac{s-6}{6},\;\frac{120-s}{60}\Big)}_{\textcolor{orange}{\mathrm{hard}:\; 6<s<120}},
      \;
    \underbrace{\vphantom{\min\big(\frac{s-6}{6},\frac{120-s}{60}\big)}0}_{\textcolor{gray}{\mathrm{outside\ hard}:\; s\le 6\ \mathrm{or}\ s\ge 120}}
    \Big\}
    \Big\}.
  \end{equation}
\item \textit{Richer requirements.} For requirements not captured by simple proxies (e.g., topic coverage, audience fit, and user-specified priorities), we use an LLM-based judging rubric: the judge is prompted with $\mathcal{U}$ and the generated slides/script and returns a scalar compliance score.
\end{itemize}
% The final requirement satisfaction score $R$ is computed as $R=\omega_{\mathrm{time}}\cdot R_{\mathrm{time}} + \omega_{\mathrm{rich}}\cdot R_{\mathrm{rich}}$.

\paragraph{Narrative diversity and controllability.}
We score narrative quality as $N=\eta\,s_{\mathrm{diver}}+(1-\eta)\,s_{\mathrm{ctrl}}$ with $\eta=0.4$.
When multiple narrative-plan candidates are available, $s_{\mathrm{diver}}$ measures plan diversity and
$s_{\mathrm{ctrl}}$ measures narrative controllability.
In the single-deck automatic setting (no plan candidates), we set $s_{\mathrm{diver}}=0$ and estimate
$s_{\mathrm{ctrl}}$ from two deck-internal cues: (i) the fraction of slides with a non-empty title line, and
(ii) the smoothness of topic transitions between consecutive slides, approximated via embedding similarity over
slide text and speaker scripts. We aggregate the two cues by averaging, and thus report $N=(1-\eta)\,s_{\mathrm{ctrl}}$.

\paragraph{Complementarity between slide and script.}
We compute the complementarity score $C$ as the mean of (i) a redundancy ``sweet-spot'' term (encouraging the script to complement rather than duplicate the slide) and (ii) a lightweight coverage term (encouraging the script to mention key points):
\begingroup
% \fontsize{8.5}{8}\selectfont % 9pt 字号，11pt 行距（可调）
\begin{equation}
C
=
\frac{1}{2m}\Bigg(
\underbrace{\sum_{k=1}^m
\Big(1-\frac{\min(|\mathrm{sim}(k)-l|,\;|\mathrm{sim}(k)-u|)}{\max(l,1-u)+\epsilon}\Big)_+}_{\textit{(i) redundancy sweet-spot}}
\;+\;
\underbrace{\sum_{k=1}^m
\frac{\#\{\text{words in $f_k$ also appearing in $p_k$}\}}
{\#\{\text{words in $f_k$}\}}}_{\textit{(ii) key-point coverage}}
\Bigg),
\end{equation}
\endgroup
where $\mathrm{sim}(k)\in[0,1]$ is the cosine similarity between slide text of $f_k$ and script $p_k$.
We use a target similarity interval $[l,u]=[0.25,0.55]$, and $(\cdot)_+=\max(\cdot,0)$.
Term (i) rewards similarity near the target interval, while (ii) measures the fraction of key content words from $f_k$ that also appear in $p_k$.

\paragraph{Temporal delivery quality.}
We estimate per-slide speaking time from script length (150 words/min) and compute $Q$ as the mean of two lightweight proxies:
\begingroup
\fontsize{8.5}{9.5}\selectfont
\newcommand{\Tall}{\vphantom{\Big(1-\frac{\mathrm{std}(\{\widehat{d}_k\})}{\max(10,D/6)+\epsilon}\Big)_+}}

\begin{equation}
\label{eq:Q_tdq}
T
=
\frac{1}{2}\Bigg(
\underbrace{\Tall\Big(1-\frac{\mathrm{std}(\{\widehat{d}_k\})}{\mathrm{std}_{\max}+\epsilon}\Big)_+}_{\textit{(i) pacing smoothness}}
\;+\;
\underbrace{\Tall\min\!\Big(1,\frac{\#\mathrm{transition\ markers}}{4}\Big)}_{\textit{(ii) transitions}}
\Bigg).
\end{equation}
\endgroup

\noindent
Here $\widehat{d}_k$ is derived from the word count of the script $p_k$ and $\mathrm{std}_{\max}$ is hyperparameter about dataset.
Intuitively, (i) rewards balanced per-slide timing, and (ii) rewards explicit transition cues in the script.

\paragraph{Attention choreography quality and rehearsal readiness.}
$T$ measures whether delivery-time guidance (e.g., focus/highlight) is synchronized with the speaker's script, while $R'$ measures how close the package is to being ``walk-on-stage'' ready.
We score both metrics in $[0,1]$ using an LLM-as-judge rubric over slide text and speaker scripts.
\paragraph{Aggregation.}
To prevent gaming the delivery scoreboard with ``flashy but wrong'' presentations, we include a portion of Artifact fundamentals:
\begin{equation}
  S_D =
\sum_{x\in\{R,N,C,T,R'\}} \omega_x\cdot x
 + \omega_{\mathrm{stab}}\cdot P
 + \omega_{\mathrm{fid}}\cdot\big(\beta F_t+(1-\beta)F_v\big),
\end{equation}
\subsection{Main Experiment: Domain-wise Evaluation}
Fig.~\ref{fig:main_exp_perf} visualizes per-domain score profiles over $20$ domains.
Across domains, DeepSlide consistently achieves strong Delivery scores while remaining competitive on Artifact quality, indicating that optimizing for end-to-end talk delivery does not require sacrificing static slide fundamentals.
Detailed metric breakdowns are reported in Table~\ref{tab:main_results} and Table~\ref{tab:supp_metrics}.

% \usepackage[table]{xcolor}
% \usepackage{booktabs}
% \usepackage{longtable}
% \usepackage{multirow}
%\definecolor{DomainGray}{HTML}{F2F2F2}
%\definecolor{DeepSlideRow}{HTML}{E9F6EF}
%\definecolor{DeepSlideStrong}{HTML}{BFE8D0}
%\newcommand{\ds}[1]{\textbf{#1}}
%\newcommand{\dsstrong}[1]{\cellcolor{DeepSlideStrong}\ds{#1}}
%\newcommand{\metrichead}[1]{\textsc{#1}}

\begin{table}[t]
  \centering
  \caption{Main experiment results. Detail metrics are shown in Table~\ref{tab:supp_metrics}.}
\label{tab:main_results}
\renewcommand{\arraystretch}{1.00}
\setlength{\tabcolsep}{4.5pt}
  \begin{tabular}{lcc|cc|cc|cc|cc|cc|cc}
  
\toprule
\multirow{2}{*}{\textbf{Method}} & \multicolumn{2}{c}{\textbf{01. AI}} & \multicolumn{2}{c}{\textbf{02. ML}} & \multicolumn{2}{c}{\textbf{03. CV}} & \multicolumn{2}{c}{\textbf{04. NLP}} & \multicolumn{2}{c}{\textbf{05. Robo}} & \multicolumn{2}{c}{\textbf{06. Sec}} & \multicolumn{2}{c}{\textbf{07. SE}} \\
\cmidrule(lr){2-3}\cmidrule(lr){4-5}\cmidrule(lr){6-7}\cmidrule(lr){8-9}\cmidrule(lr){10-11}\cmidrule(lr){12-13}\cmidrule(lr){14-15}
 & $S_A$ & $S_D$ & $S_A$ & $S_D$ & $S_A$ & $S_D$ & $S_A$ & $S_D$ & $S_A$ & $S_D$ & $S_A$ & $S_D$ & $S_A$ & $S_D$ \\
\midrule
PPTAgent & 0.82 & 0.56 & 0.80 & 0.54 & 0.80 & 0.54 & 0.80 & 0.70 & 0.83 & 0.64 & 0.79 & 0.64 & 0.80 & 0.61 \\
Qwen & 0.73 & 0.61 & 0.76 & 0.61 & 0.75 & 0.62 & 0.76 & 0.59 & 0.76 & 0.60 & 0.73 & 0.59 & 0.77 & 0.60 \\
Coze & 0.78 & 0.50 & 0.80 & 0.52 & 0.81 & 0.52 & 0.81 & 0.49 & 0.81 & 0.54 & 0.74 & 0.45 & 0.68 & 0.41 \\
Gamma & 0.84 & 0.76 & 0.83 & 0.76 & 0.81 & 0.74 & 0.81 & 0.74 & 0.82 & 0.76 & 0.82 & 0.75 & 0.80 & 0.75 \\
Manus & 0.82 & 0.75 & 0.84 & 0.75 & 0.75 & 0.73 & 0.82 & 0.71 & 0.60 & 0.58 & 0.84 & 0.75 & 0.84 & 0.75 \\
NotebookLM & 0.75 & 0.46 & 0.81 & 0.49 & 0.77 & 0.46 & 0.80 & 0.49 & 0.81 & 0.50 & 0.81 & 0.49 & 0.75 & 0.47 \\
\rowcolor{DeepSlideRow}
\ds{DeepSlide} & \ds{0.83} & \dsstrong{0.76} & \dsstrong{0.93} & \dsstrong{0.78} & \dsstrong{0.85} & \dsstrong{0.77} & \dsstrong{0.93} & \dsstrong{0.74} & \dsstrong{0.86} & \dsstrong{0.77} & \ds{0.83} & \ds{0.74} & \ds{0.82} & \ds{0.75} \\
\midrule
\addlinespace[1em]
\multirow{2}{*}{\textbf{Method}} & \multicolumn{2}{c}{\textbf{08. Sig}} & \multicolumn{2}{c}{\textbf{09. Ast}} & \multicolumn{2}{c}{\textbf{10. HEP}} & \multicolumn{2}{c}{\textbf{11. CMP}} & \multicolumn{2}{c}{\textbf{12. Quan}} & \multicolumn{2}{c}{\textbf{13. Phys}} & \multicolumn{2}{c}{\textbf{14. Pure}} \\
\cmidrule(lr){2-3}\cmidrule(lr){4-5}\cmidrule(lr){6-7}\cmidrule(lr){8-9}\cmidrule(lr){10-11}\cmidrule(lr){12-13}\cmidrule(lr){14-15}
 & $S_A$ & $S_D$ & $S_A$ & $S_D$ & $S_A$ & $S_D$ & $S_A$ & $S_D$ & $S_A$ & $S_D$ & $S_A$ & $S_D$ & $S_A$ & $S_D$ \\
\midrule

PPTAgent & 0.81 & 0.59 & 0.80 & 0.67 & 0.80 & 0.66 & 0.80 & 0.64 & 0.65 & 0.57 & 0.80 & 0.65 & 0.81 & 0.72 \\
Qwen & 0.75 & 0.60 & 0.77 & 0.63 & 0.78 & 0.64 & 0.77 & 0.64 & 0.78 & 0.65 & 0.78 & 0.61 & 0.73 & 0.60 \\
Coze & 0.80 & 0.49 & 0.80 & 0.54 & 0.69 & 0.45 & 0.73 & 0.47 & 0.79 & 0.50 & 0.72 & 0.47 & 0.75 & 0.47 \\
Gamma & 0.79 & 0.72 & 0.83 & 0.71 & 0.82 & 0.76 & 0.82 & 0.68 & 0.81 & 0.71 & 0.83 & 0.76 & 0.83 & 0.76 \\
Manus & 0.84 & 0.75 & 0.84 & 0.75 & 0.84 & 0.76 & 0.84 & 0.65 & 0.84 & 0.75 & 0.84 & 0.76 & 0.78 & 0.74 \\
NotebookLM & 0.81 & 0.49 & 0.81 & 0.50 & 0.80 & 0.49 & 0.81 & 0.49 & 0.81 & 0.49 & 0.74 & 0.47 & 0.81 & 0.49 \\
\rowcolor{DeepSlideRow}
\ds{DeepSlide} & \ds{0.83} & \dsstrong{0.76} & \ds{0.83} & \dsstrong{0.77} & \dsstrong{0.87} & \dsstrong{0.78} & \ds{0.83} & \dsstrong{0.76} & \dsstrong{0.85} & \dsstrong{0.76} & \dsstrong{0.91} & \dsstrong{0.77} & \dsstrong{0.83} & \dsstrong{0.77} \\
\midrule
\addlinespace[1em]
\multirow{2}{*}{\textbf{Method}} & \multicolumn{2}{c}{\textbf{15. Appl}} & \multicolumn{2}{c}{\textbf{16. Stats}} & \multicolumn{2}{c}{\textbf{17. Bio}} & \multicolumn{2}{c}{\textbf{18. QFin}} & \multicolumn{2}{c}{\textbf{19. Econ}} & \multicolumn{2}{c}{\textbf{20. Soc}} & \multicolumn{2}{c}{\textbf{Avg}} \\
\cmidrule(lr){2-3}\cmidrule(lr){4-5}\cmidrule(lr){6-7}\cmidrule(lr){8-9}\cmidrule(lr){10-11}\cmidrule(lr){12-13}\cmidrule(lr){14-15}
 & $S_A$ & $S_D$ & $S_A$ & $S_D$ & $S_A$ & $S_D$ & $S_A$ & $S_D$ & $S_A$ & $S_D$ & $S_A$ & $S_D$ & $S_A$ & $S_D$ \\
\midrule

PPTAgent & 0.64 & 0.51 & 0.80 & 0.66 & 0.80 & 0.70 & 0.80 & 0.57 & 0.63 & 0.50 & 0.81 & 0.50 & 0.78 & 0.61 \\
Qwen & 0.77 & 0.63 & 0.71 & 0.55 & 0.78 & 0.64 & 0.77 & 0.61 & 0.76 & 0.59 & 0.77 & 0.63 & 0.76 & 0.61 \\
Coze & 0.77 & 0.46 & 0.74 & 0.43 & 0.76 & 0.43 & 0.79 & 0.48 & 0.81 & 0.49 & 0.76 & 0.48 & 0.77 & 0.48 \\
Gamma & 0.70 & 0.64 & 0.82 & 0.74 & 0.66 & 0.63 & 0.84 & 0.75 & 0.83 & 0.75 & 0.64 & 0.63 & 0.80 & 0.72 \\
Manus & 0.84 & 0.76 & 0.83 & 0.75 & 0.84 & 0.73 & 0.84 & 0.76 & 0.81 & 0.73 & 0.84 & 0.76 & 0.82 & 0.73 \\
NotebookLM & 0.81 & 0.49 & 0.81 & 0.49 & 0.82 & 0.49 & 0.82 & 0.50 & 0.80 & 0.49 & 0.81 & 0.50 & 0.80 & 0.49 \\
\rowcolor{DeepSlideRow}
\ds{DeepSlide} & \ds{0.83} & \dsstrong{0.76} & \dsstrong{0.83} & \dsstrong{0.77} & \dsstrong{0.84} & \dsstrong{0.76} & \ds{0.83} & \ds{0.76} & \dsstrong{0.92} & \dsstrong{0.78} & \dsstrong{0.97} & \dsstrong{0.78} & \dsstrong{0.86} & \dsstrong{0.76} \\
% \midrule
\bottomrule
\end{tabular}
\end{table}

\begin{figure}[htbp]
% \small
  \centering
  % \vspace{-2mm}
  \includegraphics[width=\linewidth]{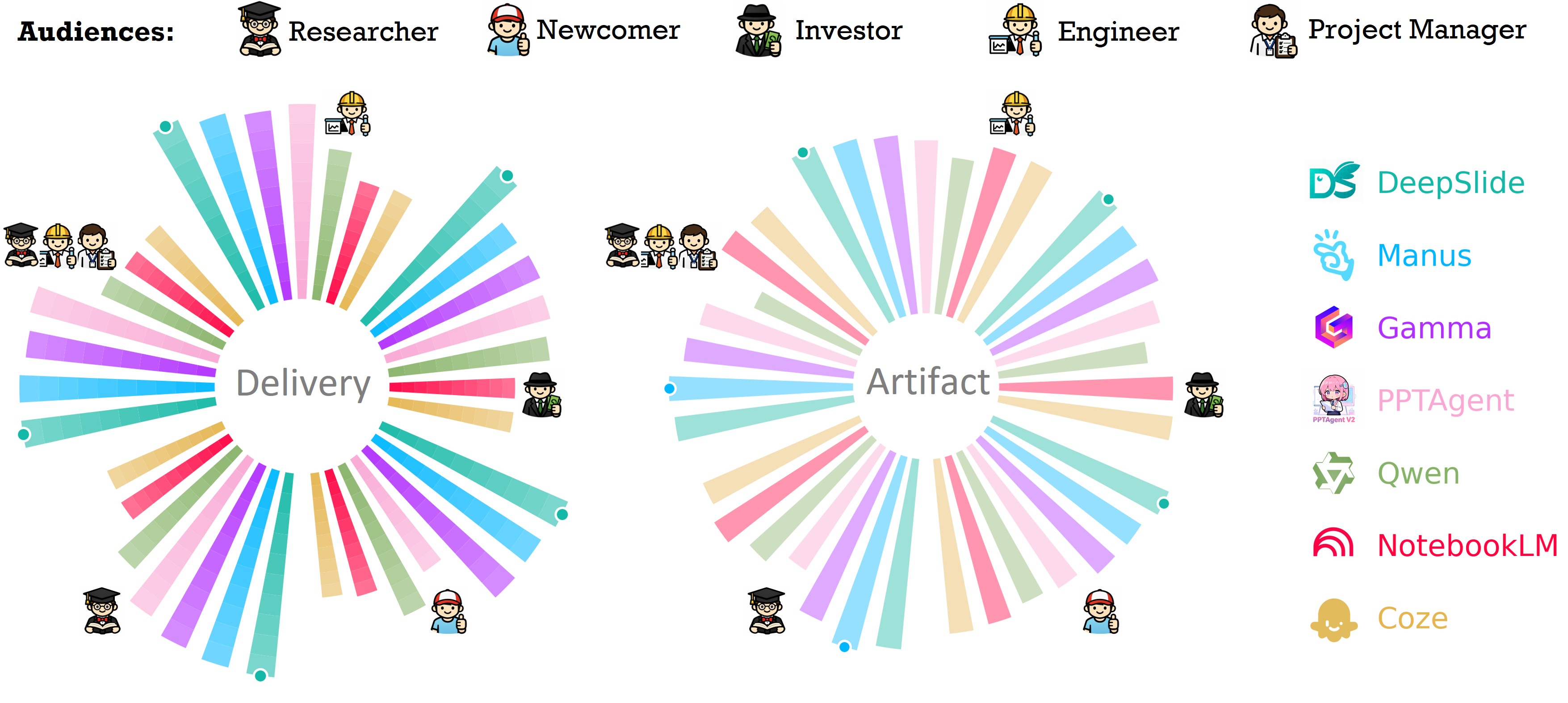}
  % \vspace{-4mm}
  \caption{Secondary experiment on audience-specific evaluation.}
  \label{fig:second_exp_perf}
  % \vspace{-2mm}
\end{figure}

\subsection{Secondary Experiment: Audience-specific Evaluation}
Table~\ref{tab:role_dual_scoreboard} and Fig~\ref{fig:second_exp_perf} report audience-specific results (e.g., engineer, investor, newcomer, researcher,
and a mixed group of researcher, engineer and project manager
).
\textit{DeepSlide} maintains robust delivery performance across audiences, suggesting that its requirement-aware planning and delivery-time support generalize beyond a single audience type.
This consistent advantage stems from our audience-oriented narrative-style logical chain recommendation: by first profiling the target listeners and then proposing a time-budgeted storyline that explicitly balances technical depth with rhetorical flow, the system ensures that slide sequencing, script emphasis, and visual evidence are all tuned to the cognitive expectations of each audience.
% Consequently, the same scientific paper can be re-storied into markedly different but equally coherent presentations—e.g., a results-first arc for engineers, a market-impact narrative for investors, or a concept-motivated path for newcomers—without sacrificing delivery fluency or content fidelity.
\subsection{Case Study and Discussion}
\begin{figure}[htbp]
% \small
  \centering
  % \vspace{-2mm}
  \includegraphics[width=\linewidth]{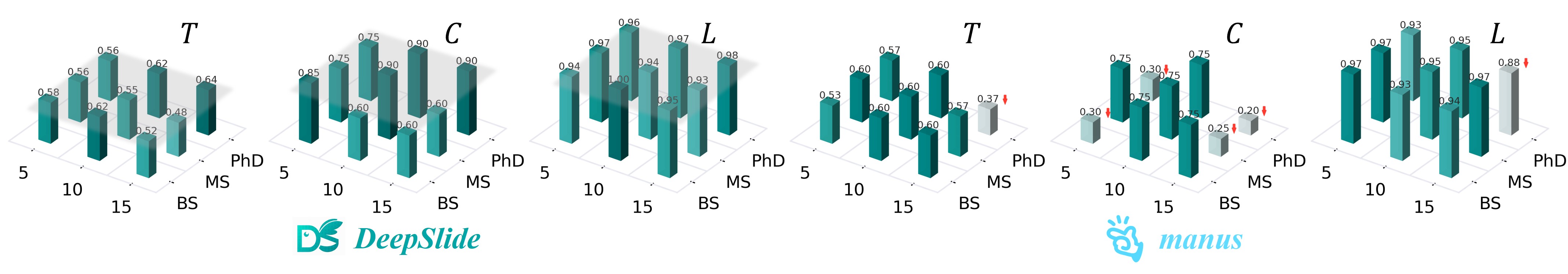}
  % \vspace{-4mm}
  \caption{\textit{DeepSlide} vs.\ Manus, varying audience \{BS, MS, PhD\} and duration \{5, 10, 15\} (Case~1).}
  \label{fig:case1}
  % \vspace{-2mm}
\end{figure}

\paragraph{Case~1: Is \textit{DeepSlide} merely move content from source to slide?}
Fig~\ref{fig:case1} contrasts \textit{DeepSlide} with Manus under varying audience types and duration budgets.
DeepSlide exhibits higher stability and more controllable narrative outcomes as requirements change, consistent with its explicit requirement-aware planning and its recommended logic chain that helps users structure a talk under constraints.

\paragraph{Case~2: Does DeepSlide reduce user burden and delivery pressure?}
% Table~\ref{tab:design_points_grouped} summarizes some design points across systems.
% DeepSlide distinguishes itself by providing delivery- and rehearsal-facing supports (e.g., speech support, complementarity between slide and script, attention strategy, preview feedback, and audio preview), which shifts effort from manual rehearsal and re-planning to guided delivery with synchronized scripts.
Table~\ref{tab:design_points_grouped} indicates that most presentation agents mainly optimize slide authoring (e.g., outlines and partial source indexing), while leaving delivery and rehearsal effort to users. DeepSlide reduces preparation pressure by making the delivery process explicit and guided. 
First, requirement elicitation and nonlinear narrative help users lock in a goal-aligned structure earlier, reducing late-stage re-planning. 
Second, synchronized slide--script generation with explicit complementarity, together with speech support and timing cues, lowers the cognitive load of pacing and deciding what to show versus what to say. 
Third, preview feedback and audio preview provide a low-cost rehearsal proxy that exposes pacing and rendering issues early, shifting effort from repetitive debugging to small intent-level edits. 
% Overall, these delivery- and rehearsal-facing supports turn preparation from trial-and-error rehearsal into a paced, cue-driven workflow, thereby reducing both user burden and live delivery pressure.
% ---------- table ----------
\begin{table}[t]
\centering
\caption{Design points across presentation systems.}
\label{tab:design_points_grouped}

\begingroup
\small
\setlength{\tabcolsep}{1.0pt}     % 只在本表内生效
\renewcommand{\arraystretch}{1.0} % 只在本表内生效

\begin{tabularx}{\linewidth}{L*{7}{Y}}
\toprule
\textbf{Design Point} &
\textbf{\textit{DeepSlide}} & \textbf{PPTAgent} & \textbf{Qwen} & \textbf{Coze} &
\textbf{Gamma} & \textbf{Manus} & \textbf{NotebookLM} \\
\midrule

% ---------- Category I ----------
\GroupRow{Category I: Content planning \& structuring}{reduce authoring / organization burden}
Requirement elicitation                 & \cmark & \xmark & \xmark & \xmark & \xmark & \xmark & \xmark \\
Source indexing                         & \cmark & \cmark & \umark & \xmark & \umark & \xmark & \umark \\
Outline                                 & \cmark & \cmark & \cmark & \cmark & \cmark & \cmark & \cmark \\
Nonlinear narrative                     & \cmark & \xmark & \xmark & \xmark & \xmark & \xmark & \xmark \\
\addlinespace[0.35ex]

% ---------- Category II ----------
\GroupRow{Category II: Delivery \& rehearsal assistance}{reduce speaking / rehearsal burden}
Speech support                          & \cmark & \xmark & \xmark & \xmark & \xmark & \xmark & \xmark \\
Complementarity (slide vs.\ script)     & \cmark & \xmark & \xmark & \xmark & \xmark & \xmark & \xmark \\
Timer                                   & \cmark & \cmark & \cmark & \xmark & \cmark & \xmark & \xmark \\
Attention strategy                      & \cmark & \xmark & \cmark & \xmark & \cmark & \xmark & \xmark \\
Audio preview                           & \cmark & \xmark & \xmark & \xmark & \xmark & \xmark & \xmark \\
\addlinespace[0.35ex]

% ---------- Category III ----------
\GroupRow{Category III: Interaction \& feedback loop}{reduce iteration / debugging burden}
Interactive data visualization          & \cmark & \xmark & \xmark & \cmark & \xmark & \cmark & \xmark \\
Preview feedback                        & \cmark & \xmark & \xmark & \xmark & \xmark & \xmark & \xmark \\
Audience Question                       & \cmark & \xmark & \xmark & \xmark & \xmark & \xmark & \xmark \\
\bottomrule
\end{tabularx}

\vspace{0.6ex}
{\footnotesize \cmark: supported\;\; \xmark: not supported\;\; \umark: unknown}

\endgroup
\end{table}

\paragraph{Case~3: Which process components drive delivery gains?}
Table~\ref{tab:case3_ablation} reports an study that removes key pipeline components.
Removing the logic chain module yields a substantial drop in Delivery score, while removing the retriever causes a smaller but consistent degradation.
suggesting that delivery advantages stem from structured process components (planning, retrieval grounding, and alignment) rather than solely from the underlying model's raw generation capability.
% This suggests that delivery advantages stem from 
% the planning in logical chain, 

% =========================
% Table (local settings, no leakage)
% =========================
\begin{table}[t]
\caption{Impact of removing key components on the Delivery Scoreboard $S_D$ (Case~3).}
\label{tab:case3_ablation}
\centering

\begingroup
% \scriptsize
\renewcommand{\arraystretch}{1.0} % only in this table
\setlength{\tabcolsep}{2.5mm}     % only in this table

% 如果你确实想再缩一点：保留 \scalebox；否则删掉 \scalebox 这一层即可
\resizebox{\linewidth}{!}{%
  \scalebox{0.90}{%
    \begin{tabular}{lccccc}
      \toprule
      \textbf{Method} & \textbf{$S_D$} & $P$ & $C$ & $T$ & $N$ \\
      \midrule

      % \rowcolor{black!6}
      % \multicolumn{6}{l}{\scriptsize
      % \textit{(by GPT-5-mini) model API}
      % } \\[-1pt]

      \textbf{DeepSlide}
      & \barcellhi{0.68}{0.68}
      & \barcellhi{0.75}{0.75}
      & \barcellhi{0.86}{0.86}
      & \barcellhi{0.61}{0.61}
      & \barcellhi{0.55}{0.55} \\

      w/o BM25 content tree retriever
      & \barcell{0.67}{0.67\,\textcolor{deltaGray}{\scriptsize(-0.01)}}
      & \barcell{0.74}{0.74\,\textcolor{deltaGray}{\scriptsize(-0.01)}}
      & \barcell{0.85}{0.85\,\textcolor{deltaGray}{\scriptsize(-0.01)}}
      & \barcell{0.59}{0.59\,\textcolor{deltaGray}{\scriptsize(-0.02)}}
      & \barcell{0.54}{0.54\,\textcolor{deltaGray}{\scriptsize(-0.01)}} \\

      w/o logical chain
      & \barcell{0.44}{0.44\,\textcolor{deltaGray}{\scriptsize(-0.24)}}
      & \barcell{0.12}{0.12\,\textcolor{deltaGray}{\scriptsize(-0.63)}}
      & \barcell{0.10}{0.10\,\textcolor{deltaGray}{\scriptsize(-0.76)}}
      & \barcell{0.36}{0.36\,\textcolor{deltaGray}{\scriptsize(-0.25)}}
      & \barcell{0.53}{0.53\,\textcolor{deltaGray}{\scriptsize(-0.02)}} \\

      w/o logical chain recommender
      & \barcell{0.68}{0.68\,\textcolor{deltaGray}{\scriptsize(-0.00)}}
      & \barcell{0.71}{0.71\,\textcolor{deltaGray}{\scriptsize(-0.04)}}
      & \barcell{0.85}{0.85\,\textcolor{deltaGray}{\scriptsize(-0.01)}}
      & \barcell{0.60}{0.60\,\textcolor{deltaGray}{\scriptsize(-0.01)}}
      & \barcell{0.53}{0.53\,\textcolor{deltaGray}{\scriptsize(-0.02)}} \\

      \bottomrule
    \end{tabular}%
  }%
}
\endgroup

\end{table}

\subsubsection{Ablation Study and Discussion}
We will extend the ablation suite with additional variants and controlled settings in future revisions.

% preamble (if not already)
% \usepackage{caption}
% \usepackage{subcaption}

\paragraph{Varying truncation length.}
We vary truncation length \(l_{\mathrm{max}} \in \{4096, 8192, 16384\}\).
    \(S_A\) peaks at \(4096\) (\(0.813\)) and then stabilizes at \(\approx 0.795\) as \(l_{\mathrm{max}}\) increases, suggesting tighter contexts reduce retrieval noise.
    \(S_D\) is nearly unchanged (\(0.491 \rightarrow 0.489\)), indicating negligible impact on delivery quality.

\begin{figure}
\centering
    \begin{subfigure}[b]{0.35\linewidth}
      \centering
      \includegraphics[width=\linewidth]{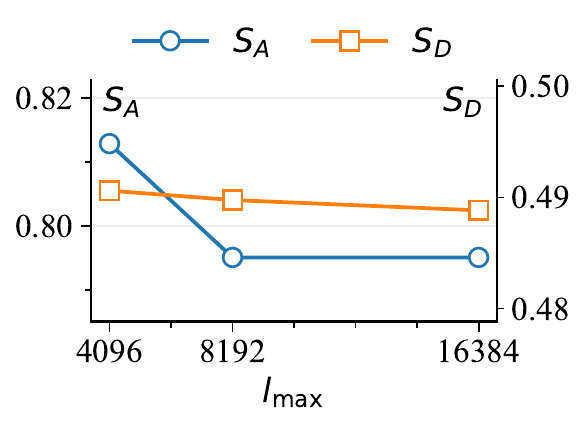}
      \vspace{-6mm}
      \caption{Varying truncation length.}
      \label{fig:ablation_trunc_len}
    \end{subfigure}
    % \hfill
    \hspace{4mm}
    \begin{subfigure}[b]{0.35\linewidth}
      \centering
      \includegraphics[width=\linewidth]{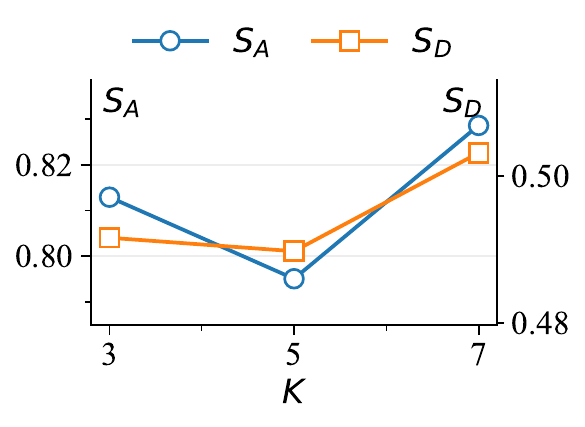}
      \vspace{-6mm}
      \caption{Varying retrieval depth.}
      \label{fig:ablation_topk}
\end{subfigure}
\caption{Varying retrieval depth.}
      \label{fig:ablation_topk}
\end{figure}

\paragraph{Varying retrieval depth.}
We vary retrieval depth \(K\in\{3,5,7\}\).
Ablation performance on CV papers is non-monotonic: the default \(K=5\) underperforms both \(K=3, 7\).
With \(K=3\), the model maintains high artifact quality by focusing on the most relevant nodes (\(S_A=0.813\)), whereas \(K=5\) injects noise with limited gain (\(S_A=0.795\)).
Increasing to \(K=7\) yields the best overall scores (\(S_A=0.829, S_D=0.503\)) and improves visual recall (\(F_v: 0.0 \rightarrow 0.125\)).
This suggests that in CV domain, complementary evidence (e.g., ablations and qualitative figures) is often dispersed across sections; broader retrieval better recovers these dependencies and strengthens grounding.

\section{Conclusion}
We presented \textit{DeepSlide}, a delivery-oriented presentation preparation system that unifies planning, generation, and rehearsal in one pipeline.
It integrates time-budgeted logical chains, content-tree grounded retrieval, Markov-style sequential rendering with style inheritance, and sandbox validation with minimal repair to improve end-to-end deliverability.
We also introduced a dual-scoreboard protocol that separates artifact quality from delivery quality, enabling targeted diagnosis of narrative, pacing, and slide--script coordination failures.
Across both the domain study and the audience-profile study, \textit{DeepSlide} matches strong baselines on artifact quality while delivering consistent, substantial gains on delivery metrics, yielding more reliable narrative flow, pacing precision, and attention guidance across diverse settings.
% Overall, the results underscore that explicit narrative control and delivery-aware evaluation are essential for presentation-ready generation.

% \begin{ack}
% Use unnumbered first level headings for the acknowledgments. All acknowledgments

% Do {\bf not} include this section in the anonymized submission, only in the final paper. You can use the \texttt{ack} environment provided in the style file to automatically hide this section in the anonymized submission.
% \end{ack}

\newpage
\bibliography{ref}

@inbook{what_makes_a_good_presentation, place={Cambridge}, title={Coherence Principle}, booktitle={Multimedia Learning}, publisher={Cambridge University Press}, author={Mayer, Richard E.}, year={2001}, pages={113–133}}

@inproceedings{rw_1_fu2022doc2ppt,
  title={Doc2ppt: Automatic presentation slides generation from scientific documents},
  author={Fu, Tsu-Jui and Wang, William Yang and McDuff, Daniel and Song, Yale},
  booktitle={Proceedings of the AAAI Conference on Artificial Intelligence},
  volume={36},
  pages={634--642},
  year={2022}
}

@article{rw_2_kumar2024slidespawn,
  title={Slidespawn: An automatic slides generation system for research publications},
  author={Kumar, Keshav and Chowdary, Ravindranath},
  journal={arXiv preprint arXiv:2411.17719},
  year={2024}
}

@article{rw_3_aggarwal2025pass,
  title={PASS: Presentation Automation for Slide Generation and Speech},
  author={Aggarwal, Tushar and Bhand, Aarohi},
  journal={arXiv preprint arXiv:2501.06497},
  year={2025}
}

@article{rw_4_yang2025autoslides,
  title={Auto-slides: An interactive multi-agent system for creating and customizing research presentations},
  author={Yang, Yuheng and Jiang, Wenjia and Wang, Yang and Wang, Yiwei and Zhang, Chi},
  journal={arXiv preprint arXiv:2509.11062},
  year={2025}
}

@inproceedings{rw_zheng-etal-2025-pptagent,
    title = "{PPTA}gent: Generating and Evaluating Presentations Beyond Text-to-Slides",
    author = "Zheng, Hao  and
      Guan, Xinyan  and
      Kong, Hao  and
      Zhang, Wenkai  and
      Zheng, Jia  and
      Zhou, Weixiang  and
      Lin, Hongyu  and
      Lu, Yaojie  and
      Han, Xianpei  and
      Sun, Le",
    editor = "Christodoulopoulos, Christos  and
      Chakraborty, Tanmoy  and
      Rose, Carolyn  and
      Peng, Violet",
    booktitle = "Proceedings of the 2025 Conference on Empirical Methods in Natural Language Processing",
    month = nov,
    year = "2025",
    address = "Suzhou, China",
    publisher = "Association for Computational Linguistics",
    url = "https://aclanthology.org/2025.emnlp-main.728/",
    doi = "10.18653/v1/2025.emnlp-main.728",
    pages = "14402--14418",
    ISBN = "979-8-89176-332-6"
}

@article{rw_xu2025pregenieagent,
  title={PreGenie: An Agentic Framework for High-quality Visual Presentation Generation},
  author={Xu, Xiaojie and Xu, Xinli and Chen, Sirui and Chen, Haoyu and Zhang, Fan and Chen, Ying-Cong},
  journal={arXiv preprint arXiv:2505.21660},
  year={2025}
}

@book{how_scientists_communicate,
author = {Kelly, Alan},
year = {2020},
month = {09},
pages = {},
title = {How Scientists Communicate: Dispatches from the Frontiers of KnowledgeDispatches from the Frontiers of Knowledge},
isbn = {9780190936600},
doi = {10.1093/oso/9780190936600.001.0001}
}

@article{rw_ofengenden2025pptarena,
  title={PPTArena: A Benchmark for Agentic PowerPoint Editing},
  author={Ofengenden, Michael and Man, Yunze and Pang, Ziqi and Wang, Yu-Xiong},
  journal={arXiv preprint arXiv:2512.03042},
  year={2025}
}

@inproceedings{rw_shi-etal-2025-presentagent,
    title = "{P}resent{A}gent: Multimodal Agent for Presentation Video Generation",
    author = "Shi, Jingwei  and
      Zhang, Zeyu  and
      Wu, Biao  and
      Liang, Yanjie  and
      Fang, Meng  and
      Chen, Ling  and
      Zhao, Yang",
    editor = {Habernal, Ivan  and
      Schulam, Peter  and
      Tiedemann, J{\"o}rg},
    booktitle = "Proceedings of the 2025 Conference on Empirical Methods in Natural Language Processing: System Demonstrations",
    month = nov,
    year = "2025",
    address = "Suzhou, China",
    publisher = "Association for Computational Linguistics",
    url = "https://aclanthology.org/2025.emnlp-demos.58/",
    doi = "10.18653/v1/2025.emnlp-demos.58",
    pages = "760--773",
    ISBN = "979-8-89176-334-0"
}

@inproceedings{ge2025autopresent,
  title={Autopresent: Designing structured visuals from scratch},
  author={Ge, Jiaxin and Wang, Zora Zhiruo and Zhou, Xuhui and Peng, Yi-Hao and Subramanian, Sanjay and Tan, Qinyue and Sap, Maarten and Suhr, Alane and Fried, Daniel and Neubig, Graham and others},
  booktitle={Proceedings of the Computer Vision and Pattern Recognition Conference},
  pages={2902--2911},
  year={2025}
}

@article{echarts,
title = {ECharts: A declarative framework for rapid construction of web-based visualization},
journal = {Visual Informatics},
volume = {2},
number = {2},
pages = {136-146},
year = {2018},
issn = {2468-502X},
doi = {https://doi.org/10.1016/j.visinf.2018.04.011},
url = {https://www.sciencedirect.com/science/article/pii/S2468502X18300068},
author = {Deqing Li and Honghui Mei and Yi Shen and Shuang Su and Wenli Zhang and Junting Wang and Ming Zu and Wei Chen}
}

@article{maheshwari2024presentations,
  title={Presentations are not always linear! gnn meets llm for document-to-presentation transformation with attribution},
  author={Maheshwari, Himanshu and Bandyopadhyay, Sambaran and Garimella, Aparna and Natarajan, Anandhavelu},
  journal={arXiv preprint arXiv:2405.13095},
  year={2024}
}

@misc{nextaidrawio,
  title         = {DayuanJiang/next-ai-draw-io: A next.js web application that integrates AI capabilities with draw.io diagrams. This app allows you to create, modify, and enhance diagrams through natural language commands and AI-assisted visualization.},
  year          = {2026},
  note          = {Accessed: 2026-02-27},
  howpublished  = {\url{https://github.com/DayuanJiang/next-ai-draw-io}}
}

@inproceedings{rw_mondal-etal-2024-presentations,
    title = "Presentations by the Humans and For the Humans: Harnessing {LLM}s for Generating Persona-Aware Slides from Documents",
    author = "Mondal, Ishani  and
      S, Shwetha  and
      Natarajan, Anandhavelu  and
      Garimella, Aparna  and
      Bandyopadhyay, Sambaran  and
      Boyd-Graber, Jordan",
    editor = "Graham, Yvette  and
      Purver, Matthew",
    booktitle = "Proceedings of the 18th Conference of the European Chapter of the Association for Computational Linguistics (Volume 1: Long Papers)",
    month = mar,
    year = "2024",
    address = "St. Julian{'}s, Malta",
    publisher = "Association for Computational Linguistics",
    url = "https://aclanthology.org/2024.eacl-long.163/",
    doi = "10.18653/v1/2024.eacl-long.163",
    pages = "2664--2684"
}

@article{the_craft_of_sc_pre,
author = {Alley, Michael},
year = {2004},
month = {07},
pages = {},
title = {The Craft of Scientific Presentations: Critical Steps to Succeed and Critical Errors to Avoid},
volume = {57},
journal = {Physics Today},
doi = {10.1063/1.1784305}
}

@misc{langchain,
  title         = {langchain-ai/langchain: The platform for reliable agents.},
  year          = {2026},
  note          = {Accessed: 2026-02-26},
  howpublished  = {\url{https://github.com/langchain-ai/langchain}}
}

@misc{qwen3.5,
    title  = {{Qwen3.5}: Towards Native Multimodal Agents},
    author = {{Qwen Team}},
    month  = {February},
    year   = {2026},
    url    = {https://qwen.ai/blog?id=qwen3.5}
}

@misc{autogpt,
  title         = {Significant-Gravitas/AutoGPT: AutoGPT is the vision of accessible AI for everyone, to use and to build on. Our mission is to provide the tools, so that you can focus on what matters.},
  year          = {2026},
  note          = {Accessed: 2026-02-26},
  howpublished  = {\url{https://github.com/Significant-Gravitas/AutoGPT}}
}

@misc{autogen,
  title         = {microsoft/autogen: A programming framework for agentic AI},
  year          = {2026},
  note          = {Accessed: 2026-02-26},
  howpublished  = {\url{https://github.com/microsoft/autogen}}
}

@inproceedings{camel,
  title={CAMEL: Communicative Agents for "Mind" Exploration of Large Language Model Society},
  author={Li, Guohao and Hammoud, Hasan Abed Al Kader and Itani, Hani and Khizbullin, Dmitrii and Ghanem, Bernard},
  booktitle={Thirty-seventh Conference on Neural Information Processing Systems},
  year={2023}
}

@inproceedings{hong2024metagpt,
      title={Meta{GPT}: Meta Programming for A Multi-Agent Collaborative Framework},
      author={Sirui Hong and Mingchen Zhuge and Jonathan Chen and Xiawu Zheng and Yuheng Cheng and Jinlin Wang and Ceyao Zhang and Zili Wang and Steven Ka Shing Yau and Zijuan Lin and Liyang Zhou and Chenyu Ran and Lingfeng Xiao and Chenglin Wu and J{\"u}rgen Schmidhuber},
      booktitle={The Twelfth International Conference on Learning Representations},
      year={2024},
      url={https://openreview.net/forum?id=VtmBAGCN7o}
}

@misc{manus-web,
  title         = {Manus},
  year          = {2026},
  note          = {Accessed: 2026-02-26},
  howpublished  = {\url{https://manus.im/app}}
}

@misc{Gamma-web,
  title         = {Gamma},
  year          = {2026},
  note          = {Accessed: 2026-02-26},
  howpublished  = {\url{https://gamma.app/}}
}

@misc{notebooklm-web,
  title         = {Google NotebookLM},
  year          = {2026},
  note          = {Accessed: 2026-02-26},
  howpublished  = {\url{https://notebooklm.google/}}
}

@misc{nanobanana,
  title         = {Gemini Image – Nano Banana — Google DeepMind},
  year          = {2026},
  note          = {Accessed: 2026-02-27},
  howpublished  = {\url{https://deepmind.google/models/gemini-image/}}
}

@misc{Coze-web,
  title         = {Coze: Next-Gen AI App Developing Platform},
  year          = {2026},
  note          = {Accessed: 2026-02-26},
  howpublished  = {\url{https://www.coze.com/}}
}

@misc{Qwen-web,
  title         = {Qwen},
  year          = {2026},
  note          = {Accessed: 2026-02-26},
  howpublished  = {\url{https://qwen.ai/home}}
}

@article{indextts,
  title={Indextts2: A breakthrough in emotionally expressive and duration-controlled auto-regressive zero-shot text-to-speech},
  author={Zhou, Siyi and Zhou, Yiquan and He, Yi and Zhou, Xun and Wang, Jinchao and Deng, Wei and Shu, Jingchen},
  journal={arXiv preprint arXiv:2506.21619},
  year={2025}
}

@misc{rw_meier2025developing,
  title={DEVELOPING A HYBRID VECTOR-GRAPH RETRIEVAL SYSTEM FOR ENTITY-PRESERVING AND INSPIRING STORYLINE CREATION OF PRESENTATION SLIDES},
  author={Meier, Alexander and Li, Mahei Manhai and Rietsche, Roman},
  year={2025}
}

@inproceedings{rw_guo2024pptc,
  title={Pptc benchmark: Evaluating large language models for powerpoint task completion},
  author={Guo, Yiduo and Zhang, Zekai and Liang, Yaobo and Zhao, Dongyan and Duan, Nan},
  booktitle={Findings of the Association for Computational Linguistics: ACL 2024},
  pages={8682--8701},
  year={2024}
}

@inproceedings{rw_lin2004rouge,
  title={Rouge: A package for automatic evaluation of summaries},
  author={Lin, Chin-Yew},
  booktitle={Text summarization branches out},
  pages={74--81},
  year={2004}
}

@inproceedings{rw_papineni2002bleu,
  title={Bleu: a method for automatic evaluation of machine translation},
  author={Papineni, Kishore and Roukos, Salim and Ward, Todd and Zhu, Wei-Jing},
  booktitle={Proceedings of the 40th annual meeting of the Association for Computational Linguistics},
  pages={311--318},
  year={2002}
}

@article{rw_zhang2019bertscore,
  title={Bertscore: Evaluating text generation with bert},
  author={Zhang, Tianyi and Kishore, Varsha and Wu, Felix and Weinberger, Kilian Q and Artzi, Yoav},
  journal={arXiv preprint arXiv:1904.09675},
  year={2019}
}

@article{rw_cachola2024knowledge,
  title={Knowledge-centric templatic views of documents},
  author={Cachola, Isabel and Cucerzan, Silviu and Herring, Allen and Mijovic, Vuksan and Oveson, Erik and Jauhar, Sujay Kumar},
  journal={arXiv preprint arXiv:2401.06945},
  year={2024}
}

@article{rw_muppidi2025taming,
  title={Taming LLMs with Negative Samples: A Reference-Free Framework to Evaluate Presentation Content with Actionable Feedback},
  author={Muppidi, Ananth and Das, Tarak and Bandyopadhyay, Sambaran and Shukla, Tripti and others},
  journal={arXiv preprint arXiv:2505.18240},
  year={2025}
}

@article{rw_huang2025pptbench,
  title={PPTBench: Towards Holistic Evaluation of Large Language Models for PowerPoint Layout and Design Understanding},
  author={Huang, Zheng and Liu, Xukai and Hu, Tianyu and Zhang, Kai and Liu, Ye},
  journal={arXiv preprint arXiv:2512.02624},
  year={2025}
}

@article{rw_he2024narrative,
  title={A Survey on Large Language Models for Narrative Visualization},
  author={He, X. and Zhang, Y. and Wang, L. and Liu, Q.},
  journal={arXiv preprint arXiv:2405.12345},
  year={2024}
}

@article{rw_slidesgen2025,
  title={SlidesGen-Bench: Evaluating Slides Generation via Computational and Quantitative Metrics},
  author={Yang, Yunqiao and Li, Wenbo and Ren, Houxing and Lu, Zimu and Wang, Ke and Huang, Zhiyuan and Zong, Zhuofan and Zhan, Mingjie and Li, Hongsheng},
  journal={arXiv preprint arXiv:2601.09487},
  year={2026}
}

@misc{microsoft-copilot,
  title         = {Microsoft 365 Copilot | AI Productivity Tools for Work},
  year          = {2026},
  note          = {Accessed: 2026-02-26},
  howpublished  = {\url{https://www.microsoft.com/en-us/microsoft-365-copilot}}
}

@misc{google-gemini,
  title         = {Google Gemini},
  year          = {2026},
  note          = {Accessed: 2026-02-26},
  howpublished  = {\url{https://gemini.google.com/app}}
}

@misc{beautiful-ai,
  title         = {Best AI Presentation Maker for Professional Decks | Beautiful.ai - Generate High-Quality Slides With the Artificial Intelligence Powered Presentation Tool Available},
  year          = {2026},
  note          = {Accessed: 2026-02-26},
  howpublished  = {\url{https://www.beautiful.ai/}}
}

%%%%%%%%%%%%%%%%%%%%%%%%%%%%%%%%%%%%%%%%%%%%%%%%%%%%%%%%%%%%

\clearpage
\newpage
\begin{appendices}

    \etocdepthtag.toc{myappendix}
    \etocsettagdepth{mtmain}{none}
    \etocsettagdepth{myappendix}{subsection}
    \etocsettocstyle{\section*{Table of Contents for Appendices}}{}
    \tableofcontents
    \newpage

    \section{System Details}
    \textit{DeepSlide}'s system implementation and LLM usage are detailed in Table~\ref{tab:agent_inventory} and Table~\ref{tab:api-usage}.
    LLMs were also used for polishing and copy-editing parts of the manuscript; the authors are fully responsible for all content.

\begin{table}[!htbp]
  \centering
  \small
  \setlength{\tabcolsep}{3.8pt}
  \renewcommand{\arraystretch}{1.05}
  \caption{Stage-wise agents and their responsibilities.}
  \label{tab:agent_inventory}
  \begin{tabularx}{\columnwidth}{p{0.32\columnwidth} X}
    \toprule
    \textbf{Agent} & \textbf{Responsibility} \\
    \midrule

    \rowcolor{gray!12}\multicolumn{2}{l}{\textit{Stage~1: Requirement elicitation and narrative proposal}} \\[-0.4ex]
    requirements collector &
    elicit audience, duration, and constraints into a requirement profile. \\
    narrative template selector &
    select a narrative template consistent with the requirement profile. \\
    logical chain generator &
    produce a time-budgeted logical chain (or chain graph) for the talk. \\

    \midrule
    \rowcolor{gray!12}\multicolumn{2}{l}{\textit{Stage~2: Logical chain editing and evidence-grounded generation}} \\[-0.4ex]
    logical edge generator &
    add sparse cross-node links for long-range coherence and bridging. \\
    semantic matcher &
    align paper sections/content tree nodes to logical nodes for retrieval. \\
    slide generator &
    retrieve evidence and synthesize slides with scripts under time budgets. \\
    compiler debugger &
    iteratively repair compilation errors until the deck compiles successfully. \\

    \midrule
    \rowcolor{gray!12}\multicolumn{2}{l}{\textit{Stage~3: Interactive slide refinement and attention-oriented augmentation}} \\[-0.4ex]
    style agent &
    choose a deck-level style prior to reduce cross-slide drift. \\
    visual intent agent &
    specify visual goals and negative constraints (what should \emph{not} be drawn). \\
    diagram spec agent &
    extract diagram semantics into a structured specification. \\
    render plan agent &
    produce an executable plan (layout/style/effects/assets) for rendering. \\
    drawio xml generator &
    emit editable drawio xml from the diagram specification. \\
    render reviewer &
    execute the deck in a browser sandbox and propose minimal fix plan. \\
    coherence analyzer &
    inject lightweight cross-slide references when needed. \\
    vlm design expert &
    resolve image--code inconsistencies via minimal frame/layout edits. \\
    editor planner &
    translate user edits into a bounded action plan. \\
    content editor &
    apply localized edits on targeted frames/titles/scripts. \\

    \midrule
    \rowcolor{gray!12}\multicolumn{2}{l}{\textit{Stage~4: Rehearsal and dual-scoreboard evaluation}} \\[-0.4ex]
    rehearsal coach &
    provide actionable delivery tips from rehearsal metrics and context. \\
    audience reviewer &
    generate likely audience questions to stress-test coverage and clarity. \\

    \bottomrule
  \end{tabularx}
\end{table}
    
    \section{Evaluation}
    
    \subsection{Baselines}
    \paragraph{Baselines.}
    We compare \textit{DeepSlide} against six representative systems: PPTAgent~\cite{rw_zheng-etal-2025-pptagent}, Manus~\cite{manus-web}, Coze~\cite{Coze-web}, Gamma~\cite{Gamma-web}, NotebookLM~\cite{notebooklm-web}, and Qwen~\cite{Qwen-web}.
    PPTAgent is open-source, while the others are closed-source commercial products.
    Detailed baseline descriptions and usage settings are reported in Table~\ref{tab:baselines_ppt_agents}.
    For Manus, Coze, and NotebookLM, which can generate slide content directly as images (e.g., via an image model such as NanoBanana~\cite{nanobanana}), we enable this image-based mode as the baseline setting.
    Image-as-slide generation has become increasingly adopted, yet it still suffers from limited editability and imprecise element-level control.
    \begin{table}[!htbp]
    \caption{Baseline systems used for comparison.}
  \label{tab:baselines_ppt_agents}
  \centering
  % \small
  \setlength{\tabcolsep}{6pt}
  \renewcommand{\arraystretch}{1.05}
  \begin{tabular}{@{} l c c c p{0.35\linewidth} @{}}
    \toprule
    \textbf{System} & \textbf{Open Source} & \textbf{License} & \textbf{Version\&Date} & \textbf{URL} \\
    \midrule
    Manus & \xmark & NA & Manus 1.6 Max $\mathrm{Pro}$ &
    \tiny{\url{https://manus.im/app}} \\
    \addlinespace[0.2em]

    Gamma & \xmark & NA & Gamma Pro &
    \tiny{\url{https://gamma.app/}} \\
    \addlinespace[0.2em]

    PPTAgent & \cmark & MIT license & PPTAgent-V2 &
    \tiny{\url{https://github.com/icip-cas/PPTAgent}} \\
    \addlinespace[0.2em]

    Qwen & \xmark & NA & Qwen3-Max &
    \tiny{\url{https://chat.qwen.ai/}} \\
    \addlinespace[0.2em]

    NoteBookLM & \xmark & NA & NoteBookLM Pro &
    \tiny{\url{https://notebooklm.google.com/}} \\
    \addlinespace[0.2em]

    Coze & \xmark & NA & Jan 13--16, 2026 &
    \tiny{\url{https://www.coze.com/}} \\
    \bottomrule
  \end{tabular}

\end{table}
    \paragraph{API usage.}
    \textit{DeepSlide} supports per-agent API configurations and also provides default grouped presets. For high-usage agents with modest intelligence requirements, we assign low-cost yet sufficiently capable models (e.g., \dsmodel, \qwenmodel). For agents that most strongly affect the Artifact/Delivery scoreboards, we use more capable models (e.g., \gptmodel, \geminimodel). Table~\ref{tab:api-usage} summarizes the configurations of \textit{DeepSlide} and PPTAgent.
    % (Preamble reminder)
% \usepackage{booktabs}
% \usepackage[table]{xcolor}

% Stage header row: no left overflow, right fully aligned

{
\setlength{\tabcolsep}{6pt}
\renewcommand{\arraystretch}{1.05}

\begin{longtable}{@{} l l @{}}

    \caption{API usage of \textit{DeepSlide} and PPTAgent~\cite{rw_zheng-etal-2025-pptagent}.}
    \label{tab:api-usage} \\ 
    \toprule
    \textbf{Agents\&System} & \textbf{API} \\
    \midrule
    \endfirsthead

    \bottomrule
    \endlastfoot 
    \multicolumn{2}{@{}l@{}}{\textcolor{DeepSlideColor}{\textit{\textbf{DeepSlide}}}}\\
    \midrule

    \stagerow{Stage 1: Requirement elicitation and narrative proposal}
    requirements collector      & \dsmodel \\
    narrative template selector & \dsmodel \\
    logical chain generator     & \dsmodel \\
    \midrule

    \stagerow{Stage 2: Logical chain editing and evidence-grounded generation}
    logical edge generator     & \dsmodel \\
    semantic matcher           & \dsmodel \\
    slide generator            & \gptmodel \\
    compiler debugger          & \dsmodel \\
    \midrule

    \stagerow{Stage 3: Interactive slide refinement and attention-oriented augmentation}
    style agent          & \dsmodel \\
    visual intent agent  & \dsmodel \\
    diagram spec agent   & \dsmodel \\
    render plan agent    & \dsmodel \\
    drawio xml generator & \dsmodel \\
    render reviewer      & \geminimodel \\
    coherence analyzer   & \dsmodel \\
    vlm design expert    & \qwenmodel \\
    editor planner       & \dsmodel \\
    content editor       & \dsmodel \\
    \midrule

    \stagerow{Stage 4: Rehearsal and dual-leaderboard evaluation}
    rehearsal coach   & \dsmodel \\
    audience reviewer & \dsmodel \\
    \midrule

    \multicolumn{2}{@{}l@{}}{\textbf{PPTAgent}}\\
    \midrule
    research\_agent      & \gptmodel \\
    design\_agent        & \gptmodel \\
    long\_context\_model & \gptmodel \\
    vision\_model        & \gptmodel \\
    pptagent             & \gptmodel \\

\end{longtable}
}
    
    \subsection{Dataset and Prompts}
    
    % To streamline the presentation, we adopt the following abbreviations for key research domains: Artificial Intelligence (AI), Machine Learning (ML), Computer Vision (CV), Natural Language Processing (NLP), Robotics and Control Systems (Robo), Cryptography and Security (Sec), Software Engineering and Systems (SE), Signal and Image Processing (Sig), Astrophysics and Cosmology (Ast), High-Energy Physics (HEP), Condensed Matter Physics (CMP), Quantum Physics (Quan), General Physics (Phys), Pure Mathematics (Pure), Applied Mathematics and Analysis (Appl), Statistics and Probability (Stats), Quantitative Biology (Bio), Quantitative Finance (QFin), Economics and Econometrics (Econ), and Social Information Networks (Soc). All subsequent references to these fields use the corresponding short forms.

    \paragraph{Domains.}
    Our main benchmark spans $20$ domains to stress-test presentation agents under diverse evidence structures and writing conventions, ranging from visually dense experimental papers (e.g., CV and Sig) to theorem-heavy manuscripts (e.g., Pure and Appl) and narrative-driven social-science writing (Soc).
    Table~\ref{tab:appendix_domains} lists the domain taxonomy used throughout the paper.
    
    \begin{table}[t]
      \caption{The $20$ domains covered in the benchmark and the dominant evidence patterns that shape slide design and delivery strategy.}
    \label{tab:appendix_domains}
    \centering
    % \small
    \setlength{\tabcolsep}{3.5pt}
    \renewcommand{\arraystretch}{1.0}
    \begin{tabular}{@{} l l p{0.56\linewidth} @{}}
    \toprule
    \textbf{Abbrev.} & \textbf{Domain} & \textbf{Typical evidence patterns in papers} \\
    \midrule
    \CatRow{Computer Science \& Engineering}
    AI   & Artificial Intelligence & Studies intelligent agents and reasoning for real-world tasks. \\
    ML   & Machine Learning & Develops learning algorithms and theory from data at scale. \\
    CV   & Computer Vision & Understands images and videos. \\
    NLP  & Natural Language Processing & Generates human language for understanding and interaction. \\
    Robo & Robotics and Control Systems & Autonomous robots with sensing, control, and planning. \\
    Sec  & Cryptography and Security & Ensures confidentiality, and robustness against adversaries. \\
    SE   & Software Engineering and Systems & Builds scalable software with performance and reliability. \\
    Sig  & Signal and Image Processing & Analyzes signals using transforms and statistical methods. \\
    \CatRow{Physics}
    Ast  & Astrophysics and Cosmology & Explores universe using observations, simulations, and theory. \\
    HEP  & High-Energy Physics & Probes fundamental particles with accelerators/detectors. \\
    CMP  & Condensed Matter Physics & Studies matter phases and properties in quantum. \\
    Quan & Quantum Physics & Investigates quantum information, circuits, and complexity. \\
    Phys & General Physics & Broad physical principles across theory and modeling. \\
    \CatRow{Mathematics \& Statistics}
    Pure & Pure Mathematics & Proves abstract structures through definitions and theorems. \\
    Appl & Applied Mathematics and Analysis & Models real phenomena with analysis, numerics, etc. \\
    Stats& Statistics and Probability & Inference and uncertainty quantification for data/experiments. \\
    \CatRow{Economics, Finance \& Society}
    Bio  & Quantitative Biology & Quantifies biological systems with experiments, models, etc. \\
    QFin & Quantitative Finance & Models markets with stochastic processes and empirical tests. \\
    Econ & Economics and Econometrics & Studies markets using theory and econometric models. \\
    Soc  & Social Information Networks & Analyzes social and networked behavior with data. \\
    \bottomrule
    \end{tabular}
    
    \end{table}
    \FloatBarrier
    
    \paragraph{Prompt usage.}
    Each benchmark instance pairs a source paper $\mathcal{T}$ with a requirement profile $\mathcal{U}$.
    In our experiments, prompts are grouped by how $\mathcal{U}$ is instantiated:
    (i) the main experiment uses the \emph{Researcher} profile as the default prompt block;
    (ii) the secondary experiment swaps only the \textcolor{DeepSlideColor}{\textbf{Audience}} field to one of five cases (Engineer, Investor, Newcomer, Researcher, Hybrid) while keeping the remaining fields consistent;
    (iii) the case study Case~2/Case~3 and ablations also use the \emph{Researcher} profile, whereas case study Case~1 varies audience background levels (BS/MS/PhD) and time budgets.
    We highlight the four user-facing fields in \textcolor{DeepSlideColor}{\textbf{color}}: \textcolor{DeepSlideColor}{\textbf{Audience}}, \textcolor{DeepSlideColor}{\textbf{Duration}}, \textcolor{DeepSlideColor}{\textbf{Focus}}, and \textcolor{DeepSlideColor}{\textbf{Style}}.
    
    \begingroup
    \setlength{\fboxsep}{10pt}
    \setlength{\fboxrule}{0.6pt}
    
    \noindent\fcolorbox{black!12}{black!2}{%
    \begin{minipage}{0.95\linewidth}
    \small
    \textbf{Prompt Block M: Researcher Profile (main experiment; Q2/Q3; ablations; also used as the ``Researcher'').}\\
    \textcolor{DeepSlideColor}{\textbf{Audience:}} researchers in the field\\
    \textcolor{DeepSlideColor}{\textbf{Duration:}} \{D\} minutes\\
    \textcolor{DeepSlideColor}{\textbf{Focus:}} novelty; assumptions; comparisons; ablations; theoretical or empirical claims\\
    \textcolor{DeepSlideColor}{\textbf{Style:}} rigorous, citation-aware, emphasize evidence and reproducibility\\
    \vspace{0.3em}\\
    Instruction: Clearly separate contributions from background, highlight what is new, and justify design choices with ablations or analysis. Keep slides concise, avoid redundancy between on-slide text and script, and make delivery transitions explicit.
    \end{minipage}%
    }
    
    \vspace{0.8em}
    
    \noindent\fcolorbox{black!12}{black!2}{%
    \begin{minipage}{0.95\linewidth}
    \small
    \textbf{Prompt Block R1: Audience = Engineer (secondary experiment).}\\
    \textcolor{DeepSlideColor}{\textbf{Audience:}} senior engineers and applied ML practitioners\\
    \textcolor{DeepSlideColor}{\textbf{Duration:}} \{D\} minutes\\
    \textcolor{DeepSlideColor}{\textbf{Focus:}} system design; implementation details; failure modes; deployment trade-offs\\
    \textcolor{DeepSlideColor}{\textbf{Style:}} technical, structured, actionable; prefer diagrams and concrete examples\\
    \vspace{0.3em}\\
    Instruction: Emphasize how the method works end-to-end, what to implement first, and what to watch out for. Use clear module boundaries and discuss cost/latency/robustness when applicable.
    \end{minipage}%
    }
    
    \vspace{0.8em}
    
    \noindent\fcolorbox{black!12}{black!2}{%
    \begin{minipage}{0.95\linewidth}
    \small
    \textbf{Prompt Block R2: Audience = Investor (secondary experiment).}\\
    \textcolor{DeepSlideColor}{\textbf{Audience:}} investors and product decision makers\\
    \textcolor{DeepSlideColor}{\textbf{Duration:}} \{D\} minutes\\
    \textcolor{DeepSlideColor}{\textbf{Focus:}} value proposition; differentiation; risks; expected impact and roadmap\\
    \textcolor{DeepSlideColor}{\textbf{Style:}} high-level, persuasive but evidence-backed; minimize technical notation\\
    \vspace{0.3em}\\
    Instruction: Start from the problem and why it matters, then present the key idea and evidence of effectiveness. Translate metrics into outcomes, and explicitly state limitations and go-to-market risks.
    \end{minipage}%
    }
    
    \vspace{0.8em}
    
    \noindent\fcolorbox{black!12}{black!2}{%
    \begin{minipage}{0.95\linewidth}
    \small
    \textbf{Prompt Block R3: Audience = Newcomer (secondary experiment).}\\
    \textcolor{DeepSlideColor}{\textbf{Audience:}} newcomers with basic ML background\\
    \textcolor{DeepSlideColor}{\textbf{Duration:}} \{D\} minutes\\
    \textcolor{DeepSlideColor}{\textbf{Focus:}} intuition; definitions; simplified workflow; one main takeaway per section\\
    \textcolor{DeepSlideColor}{\textbf{Style:}} gentle pacing, minimal jargon, use analogies and progressive disclosure\\
    \vspace{0.3em}\\
    Instruction: Explain prerequisites briefly, define key terms before using them, and prioritize conceptual clarity over exhaustive details. Use fewer equations and more visual/step-based explanations.
    \end{minipage}%
    }
    
    \vspace{0.8em}
    
    \noindent\fcolorbox{black!12}{black!2}{%
    \begin{minipage}{0.95\linewidth}
    \small
    \textbf{Prompt Block R4: Audience = Researcher+Engineer+PM (Hybrid) (secondary experiment).}\\
    \textcolor{DeepSlideColor}{\textbf{Audience:}} mixed audience (research, engineering, product)\\
    \textcolor{DeepSlideColor}{\textbf{Duration:}} \{D\} minutes\\
    \textcolor{DeepSlideColor}{\textbf{Focus:}} core idea; engineering feasibility; product implications; evaluation evidence\\
    \textcolor{DeepSlideColor}{\textbf{Style:}} balanced depth, executive clarity, and technical grounding\\
    \vspace{0.3em}\\
    Instruction: Present the narrative in three layers: what it is (intuition), why it works (key mechanism), and how to use it (integration and constraints). Use one slide to summarize practical takeaways and risks.
    \end{minipage}%
    }
    
    \vspace{0.8em}
    
    \noindent\fcolorbox{black!12}{black!2}{%
    \begin{minipage}{0.95\linewidth}
    \small
    \textbf{Prompt Block C: Case Study Q1 (audience background level and time budget sweeps).}\\
    \textcolor{DeepSlideColor}{\textbf{Audience:}} \{BS-level / MS-level / PhD-level\}\\
    \textcolor{DeepSlideColor}{\textbf{Duration:}} \(\{D_1, D_2, \dots\}\) minutes\\
    \textcolor{DeepSlideColor}{\textbf{Focus:}} controllable narrative under constraints; robustness to requirement changes\\
    \textcolor{DeepSlideColor}{\textbf{Style:}} requirement-aware, structured, avoid over-specialization beyond the target level\\
    \vspace{0.3em}\\
    Instruction: Adapt depth, terminology, and the amount of background to the specified audience level. Under the given duration, prioritize the most important contributions and keep transitions explicit.
    \end{minipage}%
    }
    
    \endgroup
    \subsection{Experimental Results}
    \paragraph{Experimental Setup.}
    We evaluate presentation agents under the proposed dual-scoreboard protocol (Section~\ref{sec:dual-scoreboard}).
    Unless otherwise noted, each system is run $\tau=5$ times per instance and we report mean scores.
    % \input{tab_main_exp_tab}
    % \usepackage[table]{xcolor}
% \usepackage{booktabs}
% \usepackage{longtable}
% \usepackage{multirow}
% \definecolor{DomainGray}{HTML}{F2F2F2}
% \definecolor{DeepSlideRow}{HTML}{E9F6EF}
% \definecolor{DeepSlideStrong}{HTML}{BFE8D0}
% \newcommand{\ds}[1]{\textbf{#1}}
% \newcommand{\dsstrong}[1]{\cellcolor{DeepSlideStrong}\ds{#1}}
% \newcommand{\metrichead}[1]{\textsc{#1}}
{
% \tiny
\renewcommand{\arraystretch}{1.05}
\setlength{\tabcolsep}{3.2pt}

\begin{longtable}{lcccc|cccc||cccc|cccc}
\caption{Supplementary detailed metrics.\label{tab:supp_metrics}}\\
\toprule
\multirow{2}{*}{\textbf{Method}} & \multicolumn{8}{c}{\textbf{01. AI}} & \multicolumn{8}{c}{\textbf{02. ML}} \\
\cmidrule(lr){2-9}\cmidrule(lr){10-17}
 & $P$ & $F_t$ & $F_v$ & $L$ & $R$ & $C$ & $T$ & $N$ & $P$ & $F_t$ & $F_v$ & $L$ & $R$ & $C$ & $T$ & $N$ \\
\midrule
\rowcolor{DeepSlideRow}
\ds{DeepSlide} & 1.00 & 0.95 & 1.00 & 0.99 & 0.74 & 0.85 & 0.57 & 0.54 & 1.00 & 0.95 & 0.67 & 1.00 & 0.74 & 0.85 & 0.61 & 0.54 \\
PPTAgent & 1.00 & 0.90 & 0.33 & 0.90 & 0.55 & 0.27 & 0.27 & 0.41 & 1.00 & 0.93 & 0.00 & 0.95 & 0.71 & 0.24 & 0.55 & 0.55 \\
Qwen & 1.00 & 0.72 & 0.00 & 0.83 & 0.40 & 0.44 & 0.51 & 0.52 & 1.00 & 0.84 & 0.00 & 0.84 & 0.50 & 0.34 & 0.48 & 0.52 \\
Coze & 1.00 & 0.91 & 0.00 & 0.89 & 0.72 & 0.18 & 0.57 & 0.55 & 1.00 & 0.92 & 0.00 & 0.98 & 0.74 & 0.30 & 0.59 & 0.55 \\
Gamma & 1.00 & 0.95 & 0.00 & 0.98 & 0.74 & 0.85 & 0.60 & 0.55 & 1.00 & 0.95 & 0.05 & 0.94 & 0.74 & 0.85 & 0.61 & 0.55 \\
Manus & 1.00 & 0.95 & 0.25 & 0.94 & 0.74 & 0.84 & 0.60 & 0.55 & 1.00 & 0.95 & 0.00 & 1.00 & 0.74 & 0.85 & 0.63 & 0.54 \\
NotebookLM& 1.00 & 0.75 & 0.00 & 1.00 & 0.64 & 0.12 & 0.58 & 0.55 & 1.00 & 0.94 & 0.00 & 0.99 & 0.73 & 0.12 & 0.61 & 0.56 \\
\midrule
\addlinespace[1em]
\multirow{2}{*}{\textbf{Method}} & \multicolumn{8}{c}{\textbf{03. CV}} & \multicolumn{8}{c}{\textbf{04. NLP}} \\
\cmidrule(lr){2-9}\cmidrule(lr){10-17}
 & $P$ & $F_t$ & $F_v$ & $L$ & $R$ & $C$ & $T$ & $N$ & $P$ & $F_t$ & $F_v$ & $L$ & $R$ & $C$ & $T$ & $N$ \\
\midrule
\rowcolor{DeepSlideRow}
\ds{DeepSlide} & 1.00 & 0.95 & 0.12 & 1.00 & 0.74 & 0.85 & 0.60 & 0.54 & 1.00 & 0.90 & 0.75 & 0.99 & 0.71 & 0.70 & 0.57 & 0.56 \\
PPTAgent & 1.00 & 0.92 & 0.26 & 0.95 & 0.71 & 0.24 & 0.57 & 0.54 & 1.00 & 0.93 & 0.03 & 0.85 & 0.73 & 0.67 & 0.60 & 0.55 \\
Qwen & 1.00 & 0.80 & 0.00 & 0.84 & 0.49 & 0.40 & 0.52 & 0.52 & 1.00 & 0.82 & 0.00 & 0.84 & 0.35 & 0.36 & 0.48 & 0.51 \\
Coze & 1.00 & 0.93 & 0.03 & 0.99 & 0.73 & 0.31 & 0.59 & 0.55 & 1.00 & 0.93 & 0.00 & 0.99 & 0.73 & 0.16 & 0.59 & 0.54 \\
Gamma & 1.00 & 0.91 & 0.00 & 0.90 & 0.72 & 0.76 & 0.60 & 0.55 & 1.00 & 0.94 & 0.00 & 0.89 & 0.73 & 0.73 & 0.59 & 0.55 \\
Manus & 1.00 & 0.90 & 0.00 & 0.69 & 0.71 & 0.75 & 0.60 & 0.55 & 1.00 & 0.90 & 0.00 & 0.98 & 0.71 & 0.70 & 0.60 & 0.54 \\
NotebookLM& 1.00 & 0.80 & 0.03 & 0.99 & 0.64 & 0.14 & 0.55 & 0.56 & 1.00 & 0.92 & 0.00 & 0.99 & 0.72 & 0.12 & 0.59 & 0.55 \\
\midrule
\addlinespace[1em]
\multirow{2}{*}{\textbf{Method}} & \multicolumn{8}{c}{\textbf{05. Robo}} & \multicolumn{8}{c}{\textbf{06. Sec}} \\
\cmidrule(lr){2-9}\cmidrule(lr){10-17}
 & $P$ & $F_t$ & $F_v$ & $L$ & $R$ & $C$ & $T$ & $N$ & $P$ & $F_t$ & $F_v$ & $L$ & $R$ & $C$ & $T$ & $N$ \\
\midrule
\rowcolor{DeepSlideRow}
\ds{DeepSlide} & 1.00 & 0.95 & 0.19 & 1.00 & 0.74 & 0.85 & 0.57 & 0.55 & 1.00 & 0.95 & -- & 1.00 & 0.74 & 0.85 & 0.44 & 0.55 \\
PPTAgent & 1.00 & 0.95 & 0.04 & 0.99 & 0.72 & 0.46 & 0.60 & 0.56 & 1.00 & 0.95 & 0.05 & 0.85 & 0.73 & 0.46 & 0.60 & 0.55 \\
Qwen & 1.00 & 0.84 & 0.00 & 0.84 & 0.40 & 0.32 & 0.50 & 0.51 & 1.00 & 0.72 & 0.00 & 0.85 & 0.39 & 0.36 & 0.48 & 0.51 \\
Coze & 1.00 & 0.95 & 0.00 & 0.97 & 0.74 & 0.40 & 0.60 & 0.55 & 1.00 & 0.73 & 0.00 & 0.99 & 0.65 & 0.12 & 0.55 & 0.55 \\
Gamma & 1.00 & 0.95 & 0.02 & 0.91 & 0.73 & 0.86 & 0.60 & 0.56 & 1.00 & 0.94 & 0.00 & 0.90 & 0.73 & 0.82 & 0.60 & 0.56 \\
Manus & 1.00 & 0.18 & 0.00 & 1.00 & 1.00 & 0.22 & 0.33 & 0.55 & 1.00 & 0.95 & -- & 1.00 & 0.74 & 0.85 & 0.60 & 0.54 \\
NotebookLM& 1.00 & 0.92 & 0.00 & 1.00 & 0.73 & 0.16 & 0.59 & 0.56 & 1.00 & 0.92 & 0.00 & 1.00 & 0.73 & 0.10 & 0.59 & 0.56 \\
\midrule
\addlinespace[1em]
\multirow{2}{*}{\textbf{Method}} & \multicolumn{8}{c}{\textbf{07. SE}} & \multicolumn{8}{c}{\textbf{08. Sig}} \\
\cmidrule(lr){2-9}\cmidrule(lr){10-17}
 & $P$ & $F_t$ & $F_v$ & $L$ & $R$ & $C$ & $T$ & $N$ & $P$ & $F_t$ & $F_v$ & $L$ & $R$ & $C$ & $T$ & $N$ \\
\midrule
\rowcolor{DeepSlideRow}
\ds{DeepSlide} & 1.00 & 0.90 & -- & 1.00 & 0.71 & 0.80 & 0.57 & 0.56 & 1.00 & 0.95 & -- & 1.00 & 0.75 & 0.85 & 0.59 & 0.54 \\
PPTAgent & 1.00 & 0.95 & 0.08 & 0.90 & 0.73 & 0.43 & 0.60 & 0.54 & 1.00 & 0.93 & 0.24 & 0.95 & 0.72 & 0.32 & 0.60 & 0.56 \\
Qwen & 1.00 & 0.86 & 0.00 & 0.84 & 0.37 & 0.38 & 0.49 & 0.51 & 1.00 & 0.79 & 0.00 & 0.84 & 0.35 & 0.44 & 0.49 & 0.52 \\
Coze & 1.00 & 0.57 & 0.00 & 0.95 & 0.55 & 0.08 & 0.52 & 0.55 & 1.00 & 0.89 & 0.00 & 0.99 & 0.72 & 0.16 & 0.57 & 0.56 \\
Gamma & 1.00 & 0.94 & 0.00 & 0.84 & 0.73 & 0.84 & 0.60 & 0.55 & 1.00 & 0.92 & 0.12 & 0.81 & 0.73 & 0.62 & 0.59 & 0.56 \\
Manus & 1.00 & 0.95 & -- & 1.00 & 0.74 & 0.85 & 0.60 & 0.55 & 1.00 & 0.95 & 0.00 & 1.00 & 0.74 & 0.85 & 0.60 & 0.54 \\
NotebookLM& 1.00 & 0.74 & 0.00 & 0.99 & 0.72 & 0.10 & 0.58 & 0.56 & 1.00 & 0.92 & 0.00 & 1.00 & 0.74 & 0.12 & 0.59 & 0.55 \\
\midrule
\addlinespace[1em]
\multirow{2}{*}{\textbf{Method}} & \multicolumn{8}{c}{\textbf{09. Ast}} & \multicolumn{8}{c}{\textbf{10. HEP}} \\
\cmidrule(lr){2-9}\cmidrule(lr){10-17}
 & $P$ & $F_t$ & $F_v$ & $L$ & $R$ & $C$ & $T$ & $N$ & $P$ & $F_t$ & $F_v$ & $L$ & $R$ & $C$ & $T$ & $N$ \\
\midrule
\rowcolor{DeepSlideRow}
\ds{DeepSlide} & 1.00 & 0.95 & -- & 1.00 & 0.77 & 0.90 & 0.58 & 0.55 & 1.00 & 0.95 & 1.00 & 1.00 & 0.77 & 0.90 & 0.59 & 0.55 \\
PPTAgent & 1.00 & 0.94 & 0.17 & 0.85 & 0.73 & 0.49 & 0.59 & 0.55 & 1.00 & 0.94 & 0.18 & 0.89 & 0.74 & 0.57 & 0.60 & 0.56 \\
Qwen & 1.00 & 0.86 & 0.00 & 0.84 & 0.49 & 0.40 & 0.51 & 0.53 & 1.00 & 0.88 & 0.00 & 0.83 & 0.47 & 0.53 & 0.49 & 0.52 \\
Coze & 1.00 & 0.94 & 0.00 & 0.95 & 0.75 & 0.41 & 0.61 & 0.54 & 1.00 & 0.55 & 0.00 & 1.00 & 0.75 & 0.08 & 0.59 & 0.56 \\
Gamma & 1.00 & 0.93 & 0.00 & 0.95 & 0.73 & 0.55 & 0.58 & 0.56 & 1.00 & 0.95 & 0.11 & 0.91 & 0.75 & 0.87 & 0.61 & 0.55 \\
Manus & 1.00 & 0.95 & -- & 1.00 & 0.74 & 0.85 & 0.63 & 0.51 & 1.00 & 0.95 & 0.00 & 1.00 & 0.77 & 0.90 & 0.62 & 0.56 \\
NotebookLM& 1.00 & 0.93 & 0.00 & 1.00 & 0.74 & 0.14 & 0.60 & 0.56 & 1.00 & 0.91 & 0.00 & 1.00 & 0.75 & 0.12 & 0.59 & 0.56 \\
\midrule
\addlinespace[1em]
\multirow{2}{*}{\textbf{Method}} & \multicolumn{8}{c}{\textbf{11. CMP}} & \multicolumn{8}{c}{\textbf{12. Quan}} \\
\cmidrule(lr){2-9}\cmidrule(lr){10-17}
 & $P$ & $F_t$ & $F_v$ & $L$ & $R$ & $C$ & $T$ & $N$ & $P$ & $F_t$ & $F_v$ & $L$ & $R$ & $C$ & $T$ & $N$ \\
\midrule
\rowcolor{DeepSlideRow}
\ds{DeepSlide} & 1.00 & 0.95 & -- & 1.00 & 0.74 & 0.85 & 0.60 & 0.55 & 1.00 & 0.95 & 0.14 & 1.00 & 0.71 & 0.90 & 0.57 & 0.54 \\
PPTAgent & 1.00 & 0.95 & 0.00 & 0.86 & 0.74 & 0.45 & 0.60 & 0.56 & 0.80 & 0.73 & 0.06 & 0.75 & 0.59 & 0.44 & 0.47 & 0.45 \\
Qwen & 1.00 & 0.86 & 0.00 & 0.84 & 0.45 & 0.52 & 0.47 & 0.53 & 1.00 & 0.88 & 0.00 & 0.84 & 0.49 & 0.54 & 0.52 & 0.52 \\
Coze & 1.00 & 0.74 & 0.00 & 0.92 & 0.64 & 0.14 & 0.56 & 0.56 & 1.00 & 0.92 & 0.00 & 0.93 & 0.74 & 0.12 & 0.58 & 0.55 \\
Gamma & 1.00 & 0.94 & 0.00 & 0.92 & 0.72 & 0.37 & 0.58 & 0.55 & 1.00 & 0.92 & 0.00 & 0.90 & 0.62 & 0.70 & 0.57 & 0.55 \\
Manus & 1.00 & 0.95 & 0.00 & 1.00 & 0.74 & 0.20 & 0.60 & 0.52 & 1.00 & 0.95 & -- & 1.00 & 0.74 & 0.85 & 0.60 & 0.55 \\
NotebookLM& 1.00 & 0.94 & 0.00 & 1.00 & 0.74 & 0.12 & 0.60 & 0.55 & 1.00 & 0.92 & 0.00 & 1.00 & 0.75 & 0.12 & 0.60 & 0.57 \\
\midrule
\addlinespace[1em]
\multirow{2}{*}{\textbf{Method}} & \multicolumn{8}{c}{\textbf{13. Phys}} & \multicolumn{8}{c}{\textbf{14. Pure}} \\
\cmidrule(lr){2-9}\cmidrule(lr){10-17}
 & $P$ & $F_t$ & $F_v$ & $L$ & $R$ & $C$ & $T$ & $N$ & $P$ & $F_t$ & $F_v$ & $L$ & $R$ & $C$ & $T$ & $N$ \\
\midrule
\rowcolor{DeepSlideRow}
\ds{DeepSlide} & 1.00 & 0.95 & 1.00 & 1.00 & 0.74 & 0.85 & 0.57 & 0.54 & 1.00 & 0.95 & -- & 1.00 & 0.74 & 0.85 & 0.60 & 0.53 \\
PPTAgent & 1.00 & 0.92 & 0.33 & 0.90 & 0.74 & 0.56 & 0.60 & 0.55 & 1.00 & 0.93 & -- & 0.89 & 0.71 & 0.62 & 0.61 & 0.54 \\
Qwen & 1.00 & 0.88 & 0.00 & 0.84 & 0.43 & 0.36 & 0.45 & 0.52 & 1.00 & 0.74 & -- & 0.84 & 0.37 & 0.44 & 0.45 & 0.52 \\
Coze & 1.00 & 0.71 & 0.17 & 0.93 & 0.64 & 0.20 & 0.53 & 0.53 & 1.00 & 0.77 & -- & 0.96 & 0.64 & 0.22 & 0.55 & 0.55 \\
Gamma & 1.00 & 0.94 & 0.00 & 0.94 & 0.75 & 0.84 & 0.61 & 0.56 & 1.00 & 0.95 & -- & 0.96 & 0.75 & 0.87 & 0.61 & 0.54 \\
Manus & 1.00 & 0.95 & 0.00 & 1.00 & 0.77 & 0.90 & 0.62 & 0.54 & 1.00 & 0.80 & -- & 0.94 & 0.74 & 0.85 & 0.63 & 0.56 \\
NotebookLM& 1.00 & 0.70 & 0.08 & 1.00 & 0.76 & 0.10 & 0.58 & 0.57 & 1.00 & 0.93 & -- & 1.00 & 0.74 & 0.10 & 0.60 & 0.56 \\
\midrule
\addlinespace[1em]
\multirow{2}{*}{\textbf{Method}} & \multicolumn{8}{c}{\textbf{15. Appl}} & \multicolumn{8}{c}{\textbf{16. Stats}} \\
\cmidrule(lr){2-9}\cmidrule(lr){10-17}
 & $P$ & $F_t$ & $F_v$ & $L$ & $R$ & $C$ & $T$ & $N$ & $P$ & $F_t$ & $F_v$ & $L$ & $R$ & $C$ & $T$ & $N$ \\
\midrule
\rowcolor{DeepSlideRow}
\ds{DeepSlide} & 1.00 & 0.95 & -- & 1.00 & 0.74 & 0.85 & 0.57 & 0.54 & 1.00 & 0.95 & -- & 1.00 & 0.77 & 0.90 & 0.58 & 0.51 \\
PPTAgent & 0.80 & 0.73 & -- & 0.75 & 0.59 & 0.40 & 0.49 & 0.45 & 1.00 & 0.92 & -- & 0.90 & 0.74 & 0.56 & 0.60 & 0.57 \\
Qwen & 1.00 & 0.86 & -- & 0.83 & 0.41 & 0.50 & 0.50 & 0.52 & 1.00 & 0.67 & -- & 0.84 & 0.28 & 0.28 & 0.43 & 0.51 \\
Coze & 1.00 & 0.82 & -- & 0.99 & 0.69 & 0.08 & 0.56 & 0.55 & 1.00 & 0.72 & -- & 0.99 & 0.56 & 0.12 & 0.50 & 0.55 \\
Gamma & 1.00 & 0.57 & -- & 0.87 & 0.66 & 0.45 & 0.56 & 0.54 & 1.00 & 0.94 & -- & 0.90 & 0.75 & 0.73 & 0.60 & 0.55 \\
Manus & 1.00 & 0.95 & -- & 1.00 & 0.77 & 0.90 & 0.62 & 0.52 & 1.00 & 0.95 & -- & 0.98 & 0.77 & 0.90 & 0.62 & 0.49 \\
NotebookLM& 1.00 & 0.92 & -- & 1.00 & 0.75 & 0.08 & 0.61 & 0.56 & 1.00 & 0.92 & -- & 1.00 & 0.72 & 0.10 & 0.62 & 0.55 \\
\midrule
\addlinespace[1em]
\multirow{2}{*}{\textbf{Method}} & \multicolumn{8}{c}{\textbf{17. Bio}} & \multicolumn{8}{c}{\textbf{18. QFin}} \\
\cmidrule(lr){2-9}\cmidrule(lr){10-17}
 & $P$ & $F_t$ & $F_v$ & $L$ & $R$ & $C$ & $T$ & $N$ & $P$ & $F_t$ & $F_v$ & $L$ & $R$ & $C$ & $T$ & $N$ \\
\midrule
\rowcolor{DeepSlideRow}
\ds{DeepSlide} & 1.00 & 0.95 & 0.00 & 1.00 & 0.74 & 0.85 & 0.56 & 0.55 & 1.00 & 0.95 & -- & 1.00 & 0.74 & 0.85 & 0.58 & 0.55 \\
PPTAgent & 1.00 & 0.94 & 0.00 & 0.85 & 0.73 & 0.66 & 0.60 & 0.54 & 1.00 & 0.95 & 0.58 & 0.90 & 0.74 & 0.20 & 0.59 & 0.54 \\
Qwen & 1.00 & 0.88 & 0.00 & 0.84 & 0.46 & 0.46 & 0.55 & 0.52 & 1.00 & 0.86 & 0.00 & 0.83 & 0.42 & 0.36 & 0.51 & 0.52 \\
Coze & 1.00 & 0.78 & 0.00 & 0.99 & 0.55 & 0.08 & 0.51 & 0.55 & 1.00 & 0.90 & 0.00 & 0.95 & 0.70 & 0.12 & 0.55 & 0.53 \\
Gamma & 1.00 & 0.40 & 0.00 & 0.95 & 0.79 & 0.34 & 0.55 & 0.55 & 1.00 & 0.95 & 0.00 & 0.97 & 0.76 & 0.77 & 0.60 & 0.54 \\
Manus & 1.00 & 0.95 & 0.00 & 1.00 & 0.74 & 0.70 & 0.60 & 0.57 & 1.00 & 0.95 & -- & 1.00 & 0.77 & 0.90 & 0.62 & 0.56 \\
NotebookLM& 1.00 & 0.95 & 0.00 & 1.00 & 0.74 & 0.10 & 0.59 & 0.56 & 1.00 & 0.95 & 0.00 & 1.00 & 0.75 & 0.12 & 0.61 & 0.56 \\
\midrule
\addlinespace[1em]
\multirow{2}{*}{\textbf{Method}} & \multicolumn{8}{c}{\textbf{19. Econ}} & \multicolumn{8}{c}{\textbf{20. Soc}} \\
\cmidrule(lr){2-9}\cmidrule(lr){10-17}
 & $P$ & $F_t$ & $F_v$ & $L$ & $R$ & $C$ & $T$ & $N$ & $P$ & $F_t$ & $F_v$ & $L$ & $R$ & $C$ & $T$ & $N$ \\
\midrule
\rowcolor{DeepSlideRow}
\ds{DeepSlide} & 1.00 & 0.95 & 0.60 & 1.00 & 0.74 & 0.85 & 0.61 & 0.56 & 1.00 & 0.95 & 1.00 & 1.00 & 0.74 & 0.85 & 0.61 & 0.55 \\
PPTAgent & 0.80 & 0.70 & 0.40 & 0.74 & 0.59 & 0.41 & 0.45 & 0.44 & 1.00 & 0.93 & 0.07 & 1.00 & 0.69 & 0.16 & 0.59 & 0.56 \\
Qwen & 1.00 & 0.84 & 0.00 & 0.84 & 0.39 & 0.28 & 0.48 & 0.52 & 1.00 & 0.86 & 0.00 & 0.84 & 0.47 & 0.42 & 0.50 & 0.52 \\
Coze & 1.00 & 0.92 & 0.00 & 1.00 & 0.72 & 0.12 & 0.58 & 0.54 & 1.00 & 0.80 & 0.01 & 0.98 & 0.65 & 0.30 & 0.52 & 0.55 \\
Gamma & 1.00 & 0.93 & 0.00 & 0.97 & 0.73 & 0.83 & 0.61 & 0.55 & 1.00 & 0.42 & 0.04 & 0.84 & 0.79 & 0.33 & 0.54 & 0.55 \\
Manus & 1.00 & 0.85 & 0.00 & 0.98 & 0.74 & 0.70 & 0.60 & 0.56 & 1.00 & 0.95 & 0.00 & 1.00 & 0.77 & 0.90 & 0.62 & 0.56 \\
NotebookLM& 1.00 & 0.90 & 0.00 & 0.99 & 0.75 & 0.12 & 0.61 & 0.55 & 1.00 & 0.95 & 0.00 & 0.99 & 0.74 & 0.18 & 0.60 & 0.56 \\
\midrule
\addlinespace[1em]
\multicolumn{17}{c}{\textbf{Overall Average}} \\
\midrule
\multirow{2}{*}{\textbf{Method}} & \multicolumn{8}{c}{\textbf{Average (20 Domains)}} & \multicolumn{8}{c}{} \\
 & $P$ & $F_t$ & $F_v$ & $L$ & $R$ & $C$ & $T$ & $N$ &  &  &  &  &  &  &  &  \\
\midrule
\rowcolor{DeepSlideRow}
\ds{DeepSlide} & 1.00 & 0.94 & 0.59 & 1.00 & 0.74 & 0.85 & 0.58 & 0.54 &  &  &  &  &  &  &  &  \\
PPTAgent & 0.97 & 0.90 & 0.17 & 0.88 & 0.70 & 0.43 & 0.56 & 0.53 &  &  &  &  &  &  &  &  \\
Qwen & 1.00 & 0.82 & 0.00 & 0.84 & 0.42 & 0.41 & 0.49 & 0.52 &  &  &  &  &  &  &  &  \\
Coze & 1.00 & 0.82 & 0.01 & 0.97 & 0.68 & 0.18 & 0.56 & 0.55 &  &  &  &  &  &  &  &  \\
Gamma & 1.00 & 0.87 & 0.02 & 0.91 & 0.73 & 0.70 & 0.59 & 0.55 &  &  &  &  &  &  &  &  \\
Manus & 1.00 & 0.89 & 0.02 & 0.98 & 0.76 & 0.77 & 0.60 & 0.54 &  &  &  &  &  &  &  &  \\
NotebookLM& 1.00 & 0.89 & 0.01 & 1.00 & 0.73 & 0.12 & 0.59 & 0.56 &  &  &  &  &  &  &  &  \\
\bottomrule
\end{longtable}
}
    
    \begingroup
\makeatletter
\@ifundefined{ds}{\newcommand{\ds}[1]{\textbf{#1}}}{\renewcommand{\ds}[1]{\textbf{#1}}}
\@ifundefined{dsstrong}{\newcommand{\dsstrong}[1]{\cellcolor{DeepSlideStrong}\ds{#1}}}{\renewcommand{\dsstrong}[1]{\cellcolor{DeepSlideStrong}\ds{#1}}}
\@ifundefined{metrichead}{\newcommand{\metrichead}[1]{\textsc{#1}}}{\renewcommand{\metrichead}[1]{\textsc{#1}}}
\makeatother

\begin{center}
  % \scriptsize
  \setlength{\tabcolsep}{1.5pt}
  \renewcommand{\arraystretch}{1.05}

\begin{longtable}{lccccc|ccccc||ccccc|ccccc}
\caption{Audience-specific dual-scoreboard results (Artifact v.s. Delivery). DeepSlide rows are highlighted.\label{tab:role_dual_scoreboard}}\\
\toprule
\multirow{2}{*}{\textbf{Method}} & \multicolumn{5}{c|}{\textbf{Artifact}} & \multicolumn{5}{c||}{\textbf{Delivery}} & \multicolumn{5}{c|}{\textbf{Artifact}} & \multicolumn{5}{c}{\textbf{Delivery}} \\
\cmidrule(lr){2-6}\cmidrule(lr){7-11}\cmidrule(lr){12-16}\cmidrule(lr){17-21}
& $S_A$ & $P$ & $F_t$ & $F_v$ & $L$ & $S_D$ & $R$ & $C$ & $T$ & $N$ & $S_A$ & $P$ & $F_t$ & $F_v$ & $L$ & $S_D$ & $R$ & $C$ & $T$ & $N$ \\
\midrule
\endfirsthead
\toprule
\multirow{2}{*}{\textbf{Method}} & \multicolumn{5}{c|}{\textbf{Artifact}} & \multicolumn{5}{c||}{\textbf{Delivery}} & \multicolumn{5}{c|}{\textbf{Artifact}} & \multicolumn{5}{c}{\textbf{Delivery}} \\
\cmidrule(lr){2-6}\cmidrule(lr){7-11}\cmidrule(lr){12-16}\cmidrule(lr){17-21}
& $S_A$ & $P$ & $F_t$ & $F_v$ & $L$ & $S_D$ & $R$ & $C$ & $T$ & $N$ & $S_A$ & $P$ & $F_t$ & $F_v$ & $L$ & $S_D$ & $R$ & $C$ & $T$ & $N$ \\
\midrule
\endhead
\midrule \multicolumn{21}{r}{\footnotesize Continued on next page.}\\
\endfoot
\bottomrule
\endlastfoot

\rowcolor{DomainGray}\multicolumn{11}{l}{\textit{Engineer}} & \multicolumn{10}{l}{\textit{Investor}} \\ \rowcolor{DeepSlideRow} \ds{DeepSlide} & \dsstrong{0.87} & 1.00 & 0.95 & 1.00 & 1.00 & \dsstrong{0.79} & 0.71 & 0.90 & 0.57 & 0.56 & \dsstrong{0.86} & 1.00 & 0.90 & 1.00 & 1.00 & \dsstrong{0.79} & 0.74 & 0.70 & 0.79 & 0.56 \\ PPTAgent & 0.79 & 1.00 & 0.95 & 1.00 & 1.00 & 0.74 & 0.74 & 0.85 & 0.63 & 0.55 & 0.77 & 1.00 & 0.90 & 1.00 & 1.00 & 0.65 & 0.69 & 0.30 & 0.57 & 0.58 \\ Qwen & 0.72 & 1.00 & 0.70 & 0.00 & 0.82 & 0.58 & 0.44 & 0.30 & 0.40 & 0.53 & 0.68 & 1.00 & 0.60 & 0.00 & 0.82 & 0.62 & 0.71 & 0.20 & 0.57 & 0.57 \\ Coze & 0.80 & 1.00 & 0.90 & 0.00 & 1.00 & 0.49 & 0.69 & 0.20 & 0.57 & 0.58 & 0.80 & 1.00 & 0.90 & 0.00 & 1.00 & 0.48 & 0.71 & 0.10 & 0.57 & 0.56 \\ Gamma & 0.82 & 1.00 & 0.90 & 0.00 & 0.95 & 0.72 & 0.71 & 0.70 & 0.58 & 0.56 & 0.83 & 1.00 & 0.90 & 0.00 & 1.00 & 0.66 & 0.71 & 0.30 & 0.57 & 0.57 \\ Manus & 0.84 & 1.00 & 0.95 & 0.00 & 1.00 & 0.75 & 0.74 & 0.85 & 0.60 & 0.55 & 0.82 & 1.00 & 0.90 & 0.00 & 1.00 & 0.65 & 0.71 & 0.30 & 0.57 & 0.55 \\ NotebookLM& 0.80 & 1.00 & 0.90 & 0.00 & 1.00 & 0.48 & 0.71 & 0.10 & 0.57 & 0.57 & 0.79 & 1.00 & 0.90 & 0.00 & 0.97 & 0.48 & 0.69 & 0.10 & 0.57 & 0.58 \\ \addlinespace[2pt] \rowcolor{DomainGray}\multicolumn{11}{l}{\textit{Newcomer}} & \multicolumn{10}{l}{\textit{Researcher}} \\ \rowcolor{DeepSlideRow} \ds{DeepSlide} & \dsstrong{0.87} & 1.00 & 0.95 & 1.00 & 1.00 & \dsstrong{0.77} & 0.66 & 0.85 & 0.57 & 0.55 & \ds{0.87} & 1.00 & 0.95 & 1.00 & 0.99 & \dsstrong{0.78} & 0.74 & 0.85 & 0.57 & 0.54 \\ PPTAgent & 0.77 & 1.00 & 0.95 & 1.00 & 1.00 & 0.51 & 0.66 & 0.30 & 0.60 & 0.55 & 0.67 & 1.00 & 0.95 & 0.00 & 0.95 & 0.72 & 0.74 & 0.80 & 0.62 & 0.56 \\ Qwen & 0.75 & 1.00 & 0.80 & 0.00 & 0.83 & 0.62 & 0.62 & 0.30 & 0.47 & 0.56 & 0.75 & 1.00 & 0.80 & 0.00 & 0.83 & 0.61 & 0.56 & 0.30 & 0.47 & 0.50 \\ Coze & 0.80 & 1.00 & 0.90 & 0.00 & 1.00 & 0.48 & 0.69 & 0.10 & 0.57 & 0.56 & 0.78 & 1.00 & 0.90 & 0.00 & 0.91 & 0.48 & 0.66 & 0.10 & 0.57 & 0.52 \\ Gamma & 0.83 & 1.00 & 0.95 & 0.00 & 0.93 & 0.76 & 0.74 & 0.85 & 0.60 & 0.56 & 0.84 & 1.00 & 0.95 & 0.00 & 0.97 & 0.76 & 0.74 & 0.85 & 0.60 & 0.56 \\ Manus & 0.83 & 1.00 & 0.95 & 0.00 & 0.97 & 0.75 & 0.74 & 0.85 & 0.60 & 0.57 & 0.91 & 1.00 & 0.95 & 1.00 & 0.75 & 0.77 & 0.74 & 0.80 & 0.62 & 0.57 \\ NotebookLM& 0.79 & 1.00 & 0.90 & 0.00 & 0.95 & 0.49 & 0.71 & 0.20 & 0.57 & 0.57 & 0.82 & 1.00 & 0.95 & 0.00 & 1.00 & 0.49 & 0.74 & 0.10 & 0.60 & 0.51 \\ \addlinespace[2pt] \rowcolor{DomainGray}\multicolumn{11}{l}{\textit{Researcher+Engineer+PM (Hybrid)}} & \multicolumn{10}{l}{\textit{}} \\ \rowcolor{DeepSlideRow} \ds{DeepSlide} & \ds{0.82} & 1.00 & 0.70 & 0.00 & 1.00 & \dsstrong{0.75} & 0.65 & 0.90 & 0.58 & 0.56 & -- & -- & -- & -- & -- & -- & -- & -- & -- & -- \\ PPTAgent & 0.74 & 1.00 & 0.95 & 1.00 & 0.75 & 0.74 & 0.71 & 0.90 & 0.60 & 0.54 & -- & -- & -- & -- & -- & -- & -- & -- & -- & -- \\ Qwen & 0.53 & 1.00 & 0.10 & 0.00 & 0.83 & 0.51 & 0.44 & 0.20 & 0.43 & 0.56 & -- & -- & -- & -- & -- & -- & -- & -- & -- & -- \\ Coze & 0.76 & 1.00 & 0.90 & 0.00 & 0.84 & 0.49 & 0.66 & 0.20 & 0.57 & 0.56 & -- & -- & -- & -- & -- & -- & -- & -- & -- & -- \\ Gamma & 0.78 & 1.00 & 0.90 & 0.00 & 0.80 & 0.73 & 0.71 & 0.70 & 0.60 & 0.55 & -- & -- & -- & -- & -- & -- & -- & -- & -- & -- \\ Manus & 0.84 & 1.00 & 0.95 & 0.00 & 1.00 & 0.74 & 0.74 & 0.85 & 0.60 & 0.53 & -- & -- & -- & -- & -- & -- & -- & -- & -- & -- \\ NotebookLM& 0.79 & 1.00 & 0.90 & 0.00 & 0.95 & 0.48 & 0.71 & 0.10 & 0.57 & 0.57 & -- & -- & -- & -- & -- & -- & -- & -- & -- & -- \\ \addlinespace[2pt]
\end{longtable}
\end{center}
\endgroup

        \section{Limitations and Future Work}

    \paragraph{Limitations.}
    \emph{Evaluation and human factors.}
    The delivery scoreboard provides fine-grained, repeatable assessment, yet automatic metrics may not fully reflect audience perception (e.g., engagement, trust, and cognitive load) in real talks.
    Future work includes larger-scale user studies spanning speakers with diverse expertise and rehearsal practices.
    \emph{Scope of effects and robustness.}
    Attention augmentation currently focuses on lightweight, controllable effects.
    More expressive effects and richer multimodal assets may further improve delivery, but also raise stability risks.
    
    \paragraph{Future.}
    % \emph{finetuning and reinforcement learning for delivery.}
    A promising direction is to adapt the underlying models to the delivery objective.
    Supervised finetuning can improve the consistency of logical chains and slide--script alignment under domain-specific styles, while reinforcement learning can optimize delivery rewards from the Delivery Scoreboard (e.g., $S_D$ and its pacing/coherence dimensions) and rehearsal-time user feedback.
    Such learning-based policies may further enhance personalization (speaker speed, audience profile) while preserving the controllability and safety constraints required in presentation settings.
    \providecommand{\DeepSlideMainDoc}{}
    
    \section{System Prompts of Each Agent}

\def\SysPromptsStandaloneEnd{}
\ifdefined\DeepSlideMainDoc\else
\documentclass{article}
\usepackage{xcolor}
\definecolor{promptteal}{RGB}{0,128,128}
\definecolor{promptred}{RGB}{180,0,0}
\newcommand{\grayarrow}{\textcolor{gray}{\(\rightarrow\)}}
\newcommand{\jsonkey}[1]{\texttt{#1}}
\newcommand{\jsonval}[1]{\texttt{#1}}
\newcommand{\jsontype}[1]{\texttt{#1}}
\newenvironment{promptbox}[2]{\par\medskip\noindent\textbf{[#1] #2}\par\small\begin{quote}}{\end{quote}\normalsize\medskip}
\begin{document}
\def\SysPromptsStandaloneEnd{\end{document}}
\fi

The following listings present the exact system prompts of each agent in our implementation.
% =================================================================
\subsection{Content Planning \& Requirements}
% =================================================================

\begin{promptbox}{D.1-1}{Requirements Collector Agent}
% Source: requirements_service.py
You are the \textcolor{promptteal}{\textbf{DeepSlide PPT-generation assistant}}.\\
Your goal is to help the user define the requirements for a presentation based on the uploaded paper.

\par\medskip
CRITICAL: Do NOT mention any tools or tool calls. Rely on the provided context (file name, abstract, and any user-provided info).

\par\medskip
Required Information to Collect:\\
1. Target Audience\\
2. Presentation Length (Duration)\\
3. Key Sections (Focus)\\
4. Style Preferences

\par\medskip
Interaction Style:\\
\grayarrow Be proactive. Read the paper content to guide the user.\\
\grayarrow Maintain context. Remember previous turns.\\
\grayarrow When the user provides new requirements, update your understanding.

\par\medskip
Completion Condition:\\
When requirements are clear/confirmed, output the JSON strictly:\\
\{\\
\hspace*{2em}\jsonkey{audience}: \jsonval{"..."},\\
\hspace*{2em}\jsonkey{duration}: \jsonval{"..."},\\
\hspace*{2em}\jsonkey{focus\_sections}: [\jsonval{"..."}],\\
\hspace*{2em}\jsonkey{style\_preference}: \jsonval{"..."}\\
\}\\
After the JSON, ask for confirmation.

\par\medskip
CRITICAL: Once the user confirms the requirements, output \textcolor{promptred}{\textbf{ONLY}} a brief acknowledgement (e.g., 'Great, generating logic chain...') and STOP.\\
\textcolor{promptred}{\textbf{DO NOT}} generate any PPT outlines, slide content, or suggestions.\\
\textcolor{promptred}{\textbf{DO NOT}} say any words about your target to collect requirements json and word *JSON*.
\end{promptbox}

\begin{promptbox}{D.1-2}{Narrative Template Selector}
% Source: template_recommender.py
You are a \textcolor{promptteal}{\textbf{narrative template selector}}.\\
You will receive paper abstract + user requirements + template ID list. Pick 4 template IDs for logic-chain generation:

\par\medskip
1) Must include pipeline and put it at chosen[0].\\
2) Must include exactly one hook version (hook must be in chosen and must not be pipeline).\\
3) Pick 2 additional distinct templates from the given pool.

\par\medskip
Return \textcolor{promptred}{\textbf{STRICT JSON ONLY}}:\\
\{\\
\hspace*{2em}\jsonkey{chosen}: [\jsonval{"pipeline"}, \jsonval{"<hook>"}, \jsonval{"<other1>"}, \jsonval{"<other2>"}],\\
\hspace*{2em}\jsonkey{hook}: \jsonval{"<hook>"},\\
\hspace*{2em}\jsonkey{reasons}: \{\jsonkey{template\_id}: \jsonval{"one-line reason"}, ...\}\\
\}
\end{promptbox}

\begin{promptbox}{D.1-3}{Logic Chain Generator Agent}
% Source: chain_ai_generator.py
You are a \textcolor{promptteal}{\textbf{logic-chain generator}} for a research paper presentation.\\
You must \textcolor{promptred}{\textbf{NOT}} mention tools or tool calls. Use the provided outline/excerpts and user requirements.

\par\medskip
Output \textcolor{promptred}{\textbf{STRICT JSON}} wrapped in ```json ```:\\
\{\\
\hspace*{2em}\jsonkey{nodes}: [\\
\hspace*{4em}\{\jsonkey{index}: \jsontype{0},\\
\hspace*{5em}\jsonkey{role}: \jsonval{"Introduction"},\\
\hspace*{5em}\jsonkey{text}: \jsonval{"Background"},\\
\hspace*{5em}\jsonkey{description}: \jsonval{"..."},\\
\hspace*{5em}\jsonkey{duration\_ratio}: \jsontype{0.25}\}\\
\hspace*{2em}],\\
\hspace*{2em}\jsonkey{edges}: [\\
\hspace*{4em}\{\jsonkey{from\_index}: \jsontype{0}, \jsonkey{to\_index}: \jsontype{1}, \jsonkey{reason}: \jsonval{"..."}, \jsonkey{type}: \jsonval{"sequential"}\}\\
\hspace*{2em}]\\
\}

\par\medskip
CRITICAL RULES:\\
1. \textbf{PRIORITIZE USER INTENT}: If 'User Requirements Context' or 'total\_duration' implies a specific structure, you \textcolor{promptred}{\textbf{MUST}} follow it.\\
2. \textbf{TEMPLATES ARE REFERENCE ONLY}: If a narrative template conflicts with user intent or duration constraints, you \textcolor{promptred}{\textbf{MUST}} compress/merge template roles to satisfy constraints.\\
3. \textbf{HARD CONSTRAINT}: total\_duration=\{duration\_text\}. You \textcolor{promptred}{\textbf{MUST}} output between \{min\_nodes\} and \{max\_nodes\} nodes.\\
4. roles should follow the focus\_sections or user instructions. You may add extra roles: Hook, Takeaway, Extra.\\
5. text must be the concise title of the node (<= 10 words).\\
6. description must be a detailed summary (2-3 sentences) combining paper content and user intent.\\
7. duration sum $\approx$ 1.0.\\
8. Edges: Create sequential edges (\jsonkey{type}=\jsonval{"sequential"}) for the main flow. \textcolor{promptred}{\textbf{Do NOT}} create reference edges initially.\\
9. LANGUAGE: Output node text and description in English.
\end{promptbox}

\begin{promptbox}{D.1-4}{Logic Chain Edge Recommender}
% Source: editor_ai_service.py
You are a \textcolor{promptteal}{\textbf{logic chain edge recommender assistant}}.\\
Based on the given node list (ordered) and context abstract, recommend a set of directed edges and provide a short reason for each edge.

\par\medskip
Output \textcolor{promptred}{\textbf{STRICT JSON}}:\\
\{\\
\hspace*{2em}\jsonkey{edges}: [\\
\hspace*{4em}\{\jsonkey{from}: \jsontype{i}, \jsonkey{to}: \jsontype{j}, \jsonkey{reason}: \jsonval{"..."}\},\\
\hspace*{4em}...\\
\hspace*{2em}]\\
\}

\par\medskip
Note: Do NOT output any explanation or markdown, \textcolor{promptred}{\textbf{ONLY JSON}}.
\end{promptbox}

\begin{promptbox}{D.1-5}{Semantic Matcher Agent}
% Source: slide_graph_generator.py
You are a \textcolor{promptteal}{\textbf{semantic matcher}}.\\
Task: Match the 'Raw Section Names' (from a LaTeX file) to the 'Logic Node Names' (from a logic chain).

\par\medskip
Rules:\\
1. Return a JSON object where keys are Raw Section Names and values are the corresponding Logic Node Names.\\
2. If a Raw Section implies or covers a Logic Node (even if wording differs), map it.\\
3. If no match is found for a raw section, map it to null.\\
4. One Logic Node can be matched by multiple Raw Sections (e.g. splitting a section).

\par\medskip
Return JSON map: \{\jsonkey{Raw Name}: \jsonval{"Logic Name"}\}
\end{promptbox}

% =================================================================
\subsection{Visual \& Layout Generation}
% =================================================================

\begin{promptbox}{D.2-1}{Deck Style Agent}
% Source: deck_style_agent.py
You are a \textcolor{promptteal}{\textbf{deck style director}}.\\
Output \textcolor{promptred}{\textbf{ONE JSON object ONLY}} matching this schema:

\par\medskip
\{\\
\hspace*{2em}\jsonkey{version}: \jsonval{"v1"},\\
\hspace*{2em}\jsonkey{persona}: \jsontype{str},\\
\hspace*{2em}\jsonkey{theme}: \jsonval{"light"}|\jsonval{"dark"},\\
\hspace*{2em}\jsonkey{palette}: \{\jsonkey{primary}:\jsontype{str},\jsonkey{secondary}:\jsontype{str},\jsonkey{accent}:\jsontype{str},\jsonkey{bg}:\jsontype{str},\jsonkey{fg}:\jsontype{str}\},\\
\hspace*{2em}\jsonkey{material}: \jsonval{"glass"}|\jsonval{"paper"}|\jsonval{"dark-grid"}|\jsonval{"bento"},\\
\hspace*{2em}\jsonkey{illustration\_style}: \jsonval{"abstract"}|\jsonval{"isometric"}|\jsonval{"flat"}|\jsonval{"3d"},\\
\hspace*{2em}\jsonkey{stroke\_strength}: \jsonval{"low"}|\jsonval{"medium"}|\jsonval{"high"},\\
\hspace*{2em}\jsonkey{radius\_strength}: \jsonval{"low"}|\jsonval{"medium"}|\jsonval{"high"},\\
\hspace*{2em}\jsonkey{shadow\_strength}: \jsonval{"low"}|\jsonval{"medium"}|\jsonval{"high"},\\
\hspace*{2em}\jsonkey{motion\_baseline}: \jsonval{"static"}|\jsonval{"low"}|\jsonval{"high"},\\
\hspace*{2em}\jsonkey{notes}: \jsontype{str}\\
\}

\par\medskip
Hard rules:\\
\grayarrow No extra keys.\\
\grayarrow No markdown.\\
\grayarrow Palette colors must be hex like \#RRGGBB.\\
\grayarrow Keep palette light and colorful;  avoid large dark blocks.
\end{promptbox}

\begin{promptbox}{D.2-2}{Visual Intent Agent}
% Source: visual_intent_agent.py
You are a \textcolor{promptteal}{\textbf{visual art director}} for scientific slide decks.\\
Task: output \textcolor{promptred}{\textbf{ONE JSON object ONLY}}.\\
Goal: produce a content-driven background/poster concept for this slide.

\par\medskip
Hard rules:\\
\grayarrow No markdown, no explanation.\\
\grayarrow Do NOT include any text/letters/numbers/logos/watermarks in the image.\\
\grayarrow The image must be thematically aligned with the slide content.\\
\grayarrow If the slide domain is physics (e.g., black holes), do NOT output robots.\\
\grayarrow If the slide domain is robotics/embodied intelligence, prefer robot manipulation/locomotion scenes.

\par\medskip
Output schema:\\
\{\\
\hspace*{2em}\jsonkey{version}: \jsonval{"v1"},\\
\hspace*{2em}\jsonkey{topic}: \jsontype{str},\\
\hspace*{2em}\jsonkey{scene}: \jsontype{str},\\
\hspace*{2em}\jsonkey{action\_sequence}: [\jsontype{str}, \jsontype{str}, \jsontype{str}],\\
\hspace*{2em}\jsonkey{mood}: \jsontype{str},\\
\hspace*{2em}\jsonkey{style\_tags}: [\jsontype{str},...],\\
\hspace*{2em}\jsonkey{negative}: [\jsontype{str},...]\\
\}
\end{promptbox}

\begin{promptbox}{D.2-3}{Presentation Creator Agent (Compressor)}
% Source: compressor.py
You are an \textcolor{promptteal}{\textbf{expert presentation creator}}.\\
Target Node: \{logic\_node.name\}\\
Description:\{logic\_node.description\}\\
Duration: \{self.target\_duration\_sec\}s
\{generated\_context\}\\
Goal: Create Beamer slides for this topic.

\par\medskip
1. Search content using tools.\\
2. Use \texttt{add\_section(latex\_cmd)} for titles/structure. Prefer \texttt{\textbackslash section\{...\}}.\\
3. Use \texttt{add\_slide(latex\_body, speech\_script)} to create content slides. Ensure content is NOT a duplicate.\\
4. Use \texttt{add\_citation} if needed.\\
5. Keep within time limit.\\
6. Reply \textcolor{promptred}{\textbf{DONE}} when finished.

\par\medskip
Style Requirements (Clean, Concise, Modern Beamer):\\
\grayarrow Layout: Use standard \texttt{itemize} or \texttt{enumerate} environments. Keep slides un-cluttered.\\
\grayarrow Content: \textcolor{promptred}{\textbf{KEY POINTS ONLY}}. Use bullet points. Avoid long paragraphs or walls of text.\\
\grayarrow Titles: \textcolor{promptred}{\textbf{ALWAYS}} use \texttt{\textbackslash frametitle\{...\}} for every slide.\\
\grayarrow Figures/Tables: If a slide contains a figure or table, do not add dense text alongside it.\\
\grayarrow Typography: \textcolor{promptred}{\textbf{Do NOT}} use \texttt{\textbackslash Large} or \texttt{\textbackslash huge} for body text. Let Beamer handle font sizes.

\par\medskip
Note:\\
1. \textcolor{promptred}{\textbf{ALWAYS}} locate the most relevant source nodes with \texttt{search\_relevant\_nodes}, then call \texttt{get\_node\_content}.\\
2. If the source node contains figures/tables, you \textcolor{promptred}{\textbf{MUST}} call \texttt{get\_node\_media} and include at least one representative figure/table.\\
3.For slides containing a figure or table: Keep the slide minimal (title + at most 1-3 short bullets, or even no bullets). Put detailed explanation into \texttt{speech\_script}.\\
4. If both a figure and a table are relevant, create separate slides for them.\\
5. If you cannot fit the figure/table into a slide, mention it in the `\texttt{speech\_script}` explicitly and optionally cite it.\\
6. DO NOT create a "big title" inside the slide by using `Large/huge` text; instead use `add\_section('section\{\{...\}\}')`.
\end{promptbox}

\begin{promptbox}{D.2-4}{Render Plan Agent}
% Source: render_plan_agent.py
You are a \textcolor{promptteal}{\textbf{senior research keynote slide designer}}.\\
Task: output ONE single JSON object called a RenderPlan for ONE slide.\\
All layout and content decisions must be made by you;  no deterministic renderer logic exists.

\par\medskip
Hard constraints:\\
\grayarrow Output \textcolor{promptred}{\textbf{JSON ONLY}}. No markdown, no explanations.\\
\grayarrow NEVER reference /preview/ images.\\
\grayarrow If you use an image, you MUST select image.url from allowed\_image\_urls.\\
\grayarrow kicker MUST be an empty string "". Do NOT invent section labels.\\
\grayarrow Choose exactly ONE layout from: \\
\grayarrow \jsonval{cover}|\jsonval{section\_transition}|\jsonval{toc}|\jsonval{references}|\\
\grayarrow \jsonval{hero\_figure}|\jsonval{metric\_cards}|\jsonval{two\_col\_compare}|\jsonval{table\_focus}|\\
\grayarrow \jsonval{one\_sentence}|\jsonval{tri\_cards}|\jsonval{timeline}|\jsonval{process\_stack}|\\
\grayarrow \jsonval{diagram\_flow}|\jsonval{solo}.\\
\grayarrow Aesthetic Requirement (URGENT): Avoid putting numbered lists (1. 2. 3.) or steps into a single long core\_message paragraph. You MUST use 'process\_stack', 'timeline', or 'tri\_cards' layout to create high-end visual components. Solo layout is for PUNCHY single sentences ONLY.\\
\grayarrow Visual Variety: Use 'hero\_figure' for impact, 'metric\_cards' for data, and 'tri\_cards' for conceptual splits. Don't let every slide look like a single card.\\
\grayarrow Ensure the required fields for the chosen layout are present.

\par\medskip
Layout requirements:\\
\grayarrow cover: title + optional subtitle;  no bullets, no steps.\\
\grayarrow section\_transition: title + core\_message (or subtitle);  minimal text, no steps.\\
\grayarrow toc: bullets (2-8) representing sections;  each bullet should be 'Section: short note' when possible.\\
\grayarrow references: bullets (1-10) each a citation/reference line;  must use references\_source if provided.\\
\grayarrow hero\_figure: image\{url,caption,focus\_template\_id?\} + optional core\_message + 0-3 bullets.\\
\grayarrow metric\_cards: metrics (2-5) + optional core\_message + 0-2 bullets.\\
\grayarrow table\_focus: MUST include table\_viz.spec with an ECharts option JSON + 0-2 bullets. No HTML table.\\
\grayarrow diagram\_flow: diagram\_spec\{nodes>=2,edges?,layout?\} + optional core\_message.\\
\grayarrow process\_stack/timeline: steps (2-5) + optional core\_message.\\
\grayarrow tri\_cards: steps (3) + optional core\_message.\\
\grayarrow two\_col\_compare: bullets (3-4) + optional core\_message.\\
\grayarrow one\_sentence: core\_message only (bullets must be empty).\\
\grayarrow solo: core\_message and/or bullets.

\par\medskip
Image Focus Templates (Pick ONE if layout=hero\_figure):\\
\grayarrow LR\_50\_50: Two equal vertical columns (left/right).\\
\grayarrow TB\_50\_50: Two equal horizontal rows (top/bottom).\\
\grayarrow LR\_33\_67: Left narrow (1/3) + right wide (2/3).\\
\grayarrow LR\_67\_33: Left wide (2/3) + right narrow (1/3).\\
\grayarrow TB\_33\_67: Top narrow (1/3) + bottom wide (2/3).\\
\grayarrow TB\_67\_33: Top wide (2/3) + bottom narrow (1/3).\\
\grayarrow COLS\_3: Three equal vertical columns (3 columns, 1 row).\\
\grayarrow ROWS\_3: Three equal horizontal rows (1 column, 3 rows).\\
\grayarrow GRID\_2X2: 2X2 grid (four quadrants).\\
\grayarrow BENTO\_SIDEBAR\_L: Left sidebar (1/4) + right main split (top/bottom).\\
\grayarrow BENTO\_SIDEBAR\_R: Right sidebar (1/4) + left main split (top/bottom).

\par\medskip
Table Viz (ECharts) rules:\\
\grayarrow For table\_focus, output: table\_viz: \{\jsonkey{spec}: \{\jsonkey{type}:\jsonval{"echarts"},\jsonkey{option}: <EChartsOption JSON>, \jsonkey{renderer}:\jsonval{"auto"}|\jsonval{"canvas"}|\jsonval{"svg"}?, \jsonkey{height}: \jsontype{number}?\}\\
\grayarrow option MUST be valid EChartsOption JSON (NO functions, NO JS code).\\
\grayarrow option MUST include series (non-empty). Each series MUST have 'type' (e.g. bar/line/scatter/heatmap).\\
\grayarrow option MUST include a valid coordinate/data system (dataset or xAxis/yAxis or polar/radar/geo).\\
\grayarrow If multiple series are present, add a legend.\\
\grayarrow Use transparent backgrounds. Keep text readable.

\par\medskip
Minimal ECharts option examples \textcolor{promptred}{\textbf{(JSON only)}}:\\
\grayarrow Bar: \{\jsonkey{grid}:\{\jsonkey{left}:42,...\},\\
\hspace*{1em}\jsonkey{xAxis}:\{\jsonkey{type}:\jsonval{"category"},\jsonkey{data}:[\jsonval{"A"},\jsonval{"B"}]\},\\
\hspace*{1em}\jsonkey{yAxis}:\{\jsonkey{type}:\jsonval{"value"}\},\\
\hspace*{1em}\jsonkey{series}:[\{\jsonkey{type}: \jsonval{"bar"},\jsonkey{data}:[1,2]\}]\}\\
\grayarrow Line: \{\jsonkey{grid}:...,\jsonkey{series}:[\{\jsonkey{type}:\jsonval{"line"},\jsonkey{smooth}:\jsonval{true},\jsonkey{data}:[1,2]\}]\}

\par\medskip
Visual Styles (style\_config):\\
\grayarrow theme\_variant: \jsonval{"default"}|\jsonval{"glass"}|\jsonval{"bento"}|\jsonval{"neon"} (use 'glass' for modern transparency, 'bento' for grid dashboards).\\
\grayarrow accent\_color: \jsonval{"primary"}|\jsonval{"cyan"}|\jsonval{"purple"}|\jsonval{"orange"}|\jsonval{"emerald"}.\\
\grayarrow motion\_intensity: \jsonval{"static"}|\jsonval{"low"}|\jsonval{"high"} (controls entry animations).\\
\grayarrow highlight\_variant: \jsonval{"aurora"}|\jsonval{"sunset"}|\jsonval{"cyber"}|\jsonval{"violet"}|\jsonval{"mono"}|\jsonval{"underline"} (choose ONE per slide).

\par\medskip
Layout Config (layout\_config):\\
\grayarrow split\_ratio: \jsonval{"50:50"}|\jsonval{"40:60"}|\jsonval{"60:40"}|\jsonval{"30:70"}|\jsonval{"70:30"} (controls container widths).\\
\grayarrow spacing: \jsonval{"compact"}|\jsonval{"normal"}|\jsonval{"loose"}.

\par\medskip
Effects:\\
\grayarrow enabled\_effects\_hint is a list of requested effects. You may output effects\_used as a SUBSET.\\
\grayarrow If "Image Focus" is used and an image is selected, you MUST pick a 'focus\_template\_id' from the list above.\\
\grayarrow If "Text Keynote" is requested, you MUST highlight 1-3 key phrases in title/core\_message/bullets by wrapping them with [[[...]]]. These will be rendered as LARGE, BOLD, GRADIENT text.\\
\grayarrow Highlight PRIORITY (in order): numeric results (e.g. QPS, Recall), key method traits, and sharp conclusion phrases.\\
\grayarrow Do NOT overuse highlights: keep most text unmarked;  only truly critical phrases get [[[...]]].

\par\medskip
Content integrity rules (CRITICAL):\\
\grayarrow Do NOT introduce new method/paper/system names that are not present in source\_compact.\\
\grayarrow Do NOT use references\_source to invent slide content;  references\_source is for the references layout only.

\par\medskip
RenderPlan schema (keys must match):\\
\{\\
\hspace*{2em}\jsonkey{slide\_role}: \jsonval{"intro"}|\jsonval{"method"}|\jsonval{"results"}|\jsonval{"conclusion"}|\jsonval{"content"},\\
\hspace*{2em}\jsonkey{kicker}: \jsontype{str},\\
\hspace*{2em}\jsonkey{title}: \jsontype{str},\\
\hspace*{2em}\jsonkey{subtitle}: \jsontype{str},\\
\hspace*{2em}\jsonkey{core\_message}: \jsontype{str},\\
\hspace*{2em}\jsonkey{author}: \jsontype{str},\\
\hspace*{2em}\jsonkey{date}: \jsontype{str},\\
\hspace*{2em}\jsonkey{layout}: \jsontype{str},\\
\hspace*{2em}\jsonkey{effects\_used}: [\jsontype{str},...],\\
\hspace*{2em}\jsonkey{layout\_config}: \{\jsonkey{split\_ratio}:\jsontype{str}, \jsonkey{spacing}:\jsontype{str}\},\\
\hspace*{2em}\jsonkey{style\_config}: \{\jsonkey{theme\_variant}:\jsontype{str}, \jsonkey{accent\_color}:\jsontype{str}, \jsonkey{motion\_intensity}:\jsontype{str}, \jsonkey{highlight\_variant}:\jsontype{str}\},\\
\hspace*{2em}\jsonkey{bullets}: [\jsontype{str},...],\\
\hspace*{2em}\jsonkey{steps}: [\jsontype{str},...],\\
\hspace*{2em}\jsonkey{metrics}: [\{\jsonkey{label}:\jsontype{str},\jsonkey{value}:\jsontype{str},\jsonkey{delta}:\jsontype{str}?\},...],\\
\hspace*{2em}\jsonkey{image}: \{\jsonkey{url}:\jsontype{str},\jsonkey{caption}:\jsontype{str},\jsonkey{focus\_template\_id}:\jsontype{str}?\}?,\\
\hspace*{2em}\jsonkey{table\_viz}: \{\jsonkey{spec}: \{\jsonkey{type}:\jsonval{"echarts"},\jsonkey{option}: \jsontype{object}, ...\}\}?,\\
\hspace*{2em}\jsonkey{diagram\_spec}: \jsontype{object}?\\
\}
\end{promptbox}

\begin{promptbox}{D.2-5}{Diagram Spec Agent}
% Source: diagram_spec_agent.py
You are a \textcolor{promptteal}{\textbf{product keynote diagram designer}}.\\
Task: output ONE SINGLE JSON object describing a diagram (nodes+edges) for ONE slide.\\
Primary goal: make the process visually clear AND aesthetically striking.

\par\medskip
Aesthetic: high-tech research keynote / futuristic system diagram.\\
\grayarrow Think in terms of modules, stages, gateways, traffic flows.\\
\grayarrow Each node should feel like a compact card in a tech dashboard.\\
\grayarrow Prefer short, punchy labels and vivid, concrete details.\\
\grayarrow Use wording that suggests rich visuals: lanes, clusters, control plane, data plane, scoring head, router, etc.

\par\medskip
Hard constraints:\\
\grayarrow Output JSON ONLY (no markdown, no explanations).\\
\grayarrow Produce between 2 and \jsontype{\{max\_nodes\}} nodes.\\
\grayarrow Nodes must have: id, phase, label. detail/metrics/progress are optional.\\
\grayarrow phase should be a short lane/cluster name (1–3 words) that can be used as a column or swimlane title.\\
\grayarrow Edges must reference existing node ids.\\
\grayarrow If edges are unclear, output a simple left-to-right chain.\\
\grayarrow Keep label short (<= 6 words). detail can be a short sentence.\\
\grayarrow Prefer labels of the form 'Verb + object' (e.g., 'Route with ef1', 'Merge top-k results').\\
\grayarrow For detail, describe what visually happens on the card (e.g., 'select K partitions via routing vectors').

\par\medskip
Schema:\\
\{\\
\hspace*{2em}\jsonkey{title}: \jsontype{str},\\
\hspace*{2em}\jsonkey{nodes}: [\\
\hspace*{4em}\{\jsonkey{id}: \jsontype{str}, \jsonkey{phase}: \jsontype{str}, \jsonkey{label}: \jsontype{str}, \jsonkey{detail}: \jsontype{str}?, \jsonkey{metrics}: [\{\jsonkey{label}:\jsontype{str},\jsonkey{value}:\jsontype{str}\}, ...]?, \jsonkey{progress}: \jsontype{number}?\}\\
\hspace*{4em}...\\
\hspace*{2em}],\\
\hspace*{2em}\jsonkey{edges}: [\{\jsonkey{from}: \jsontype{str}, \jsonkey{to}: \jsontype{str}, \jsonkey{label}: \jsontype{str}?\}, ...],\\
\hspace*{2em}\jsonkey{layout}: \{\jsonkey{direction}: \jsonval{"LR"}|\jsonval{"TB"}\}\\
\}
\end{promptbox}

\begin{promptbox}{D.2-6}{DrawIO XML Generator Agent}
% Source: drawio_agent.py -> generate_drawio_xml_from_spec
% Note: This prompt is dynamically assembled. The following is the full template including injected CSS styles.
Title: \jsontype{\{title\}}\\
CHOSEN\_TEMPLATE: \jsonval{\{template\_id\}}\\
Rule: You MUST follow CHOSEN\_TEMPLATE exactly. Do NOT use lanes/swimlanes unless CHOSEN\_TEMPLATE is TEMPLATE\_A.\\
Goal: modern, compact, keynote-style diagram. Colorful, clean, readable at a glance.

\par\medskip
CANVAS (must follow):\\
\grayarrow Single 16:9 page. Page size: 960x540. Keep all shapes within the page with ~24px margins.\\
\grayarrow Flow direction: \jsonval{'top-to-bottom'} (or \jsonval{'left-to-right'}).

\par\medskip
STYLE SYSTEM (light pastel only):\\
\grayarrow Keep geometry/font/shadow tokens from the base styles. You may only vary fillColor/gradientColor using palettes below.\\
\grayarrow PHASE\_CONTAINER\_STYLE = \jsonval{"rounded=1; whiteSpace=wrap; html=1; container=1; collapsible=0; fillColor=\#e0f2fe; gradientColor=\#bfdbfe; gradientDirection=180; strokeColor=\#38bdf8; strokeWidth=1.3; dashed=0; shadow=1; fontFamily=Inter; fontSize=12; fontStyle=1; fontColor=\#0f172a; align=left; verticalAlign=top; spacing=14; spacingTop=18; "}\\
\grayarrow GROUP\_CONTAINER\_STYLE = \jsonval{"rounded=1; whiteSpace=wrap; html=1; container=1; collapsible=0; fillColor=\#f8fafc; gradientColor=\#e2e8f0; gradientDirection=180; strokeColor=\#cbd5e1; strokeWidth=1.1; dashed=0; shadow=0; fontFamily=Inter; fontSize=12; fontStyle=1; fontColor=\#0f172a; align=left; verticalAlign=top; spacing=14; spacingTop=18; "}\\
\grayarrow CARD\_STYLE = \jsonval{"rounded=1; whiteSpace=wrap; html=1; fillColor=\#ecfdf3; gradientColor=\#bbf7d0; gradientDirection=180; strokeColor=\#22c55e; strokeWidth=1.1; shadow=1; fontFamily=Inter; fontSize=13; fontColor=\#082f49; align=left; verticalAlign=middle; spacing=14; "}\\
\grayarrow EDGE\_PRIMARY\_STYLE = \jsonval{"edgeStyle=orthogonalEdgeStyle; rounded=1; orthogonalLoop=1; jettySize=auto; html=1; strokeColor=\#38bdf8; strokeWidth=2; endArrow=classic; endFill=1; flowAnimation=1; "}\\
\grayarrow EDGE\_SECONDARY\_STYLE = \jsonval{"edgeStyle=orthogonalEdgeStyle; rounded=1; orthogonalLoop=1; jettySize=auto; html=1; strokeColor=\#94a3b8; strokeWidth=1.5; endArrow=classic; endFill=1; dashed=1; dashPattern=4 4; opacity=60; flowAnimation=0; "}

\par\medskip
COLOR PALETTES (choose 2–4 per diagram, apply via fillColor/gradientColor):\\
\grayarrow Containers: P1: fillColor=\#e0f2fe gradientColor=\#bfdbfe ... (list continues)\\
\grayarrow Cards: C1: fillColor=\#ecfeff gradientColor=\#cffafe ... (list continues)

\par\medskip
TEMPLATES (for reference;  you must follow CHOSEN\_TEMPLATE):\\
\grayarrow TEMPLATE\_A (Phase lanes board): DISABLED. Do not use swimlanes.\\
\grayarrow TEMPLATE\_B (Linear flowchart): cards only;  no containers;  connect steps in order.\\
\grayarrow TEMPLATE\_C (Clustered blocks): 2–4 group containers (NOT swimlanes) + cards inside each group.\\
\grayarrow TEMPLATE\_D (Hub-and-spoke): one central card + 3–6 satellite cards;  no containers.\\
\grayarrow Avoid crossing edges. Prefer orthogonal connectors and keep connectors short and tidy.

\par\medskip
Hard constraints (must follow):\\
\grayarrow Keep text minimal to avoid overlap.\\
\grayarrow Each visible step card must contain <= 44 characters total.\\
\grayarrow No bullet lists. No multi-line paragraphs. Avoid line breaks (no \&\#10; , no \\n).\\
\grayarrow You MAY use HTML inside mxCell value to create hierarchy, but then you MUST XML-ESCAPE it inside the value attribute.\\
\grayarrow This means you must NOT output raw '<' or '>' inside value="...". Use '\&lt; ' and '\&gt; '.\\
\grayarrow Avoid separate text-only cells;  keep text inside the cards.\\
\grayarrow Do not put labels on connectors (edge labels length must be <= 0).\\
\grayarrow Keep everything within one page viewport (960x540) with comfortable spacing.

\par\medskip
Style requirements:\\
\grayarrow For TEMPLATE\_A: lanes must be swimlanes;  lanes use PHASE\_CONTAINER\_STYLE.\\
\grayarrow For TEMPLATE\_C: groups are containers but NOT swimlanes;  groups use GROUP\_CONTAINER\_STYLE.\\
\grayarrow Cards use CARD\_STYLE (and may pick a card palette).\\
\grayarrow Apply EDGE\_PRIMARY\_STYLE to the main connectors. Use EDGE\_SECONDARY\_STYLE only for secondary relations.

\par\medskip
Nodes (Dynamically filled):\\
\grayarrow \jsontype{\{node\_id\}} [\jsontype{\{phase\}}]: \jsontype{\{label\}} | hint: \jsontype{\{brief\_text\}}

\par\medskip
Edges (Dynamically filled):\\
\grayarrow \jsontype{\{from\}} -> \jsontype{\{to\}}
\end{promptbox}

% =================================================================
\subsection{Quality Review \& Compilation}
% =================================================================

\begin{promptbox}{D.3-1}{Render Review Agent}
% Source: render_review_agent.py
You are a \textcolor{promptteal}{\textbf{senior design QA reviewer}} for Spec-mode HTML slides.

\par\medskip
You are given:\\
\grayarrow The structured RenderPlan used to render a slide.\\
\grayarrow A small JSON meta summary about the rendered layout and effects.\\
\grayarrow A snippet of the final HTML (including CSS/JS and body markup).\\
\grayarrow Optionally, a sandbox\_report from a headless browser run containing JS errors, network failures, and basic layout info.

\par\medskip
Your task:\\
1) Detect structural or visual failures in the slide, focusing on:\\
   \grayarrow Missing or unused assets (hero images, table viz, diagrams).\\
   \grayarrow Mismatch between layout/effects and actual HTML (e.g., Image Focus effect but no ROI tiles).\\
   \grayarrow Clearly wrong layout choices (e.g., diagram\_layout used when a main figure is available).\\
   \grayarrow Empty or nearly empty content regions (no visible text or visuals).\\
2) Propose a SMALL patch to the RenderPlan (partial JSON) that would fix the most important problems.\\
   \grayarrow Only touch layout / style\_config / layout\_config / effects\_used / image.focus\_template\_id / diagram\_spec.\\
   \grayarrow Do NOT rewrite the actual content (title/core\_message/bullets/steps) except when absolutely necessary.\\
3) Provide optional notes\_for\_slide\_agent that explains how future generations could avoid the same issue.

\par\medskip
IMPORTANT:\\
\grayarrow Output \textcolor{promptred}{STRICT JSON only}, matching the schema:\\
\{\\
\hspace*{2em}\jsonkey{issues}: [\\
\hspace*{4em}\{\jsonkey{id}: \jsontype{str}, \jsonkey{severity}: \jsonval{"low"}|\jsonval{"medium"}|\jsonval{"high"}|\jsonval{"critical"}, \jsonkey{message}: \jsontype{str}, \jsonkey{hint}: \jsontype{str}, \jsonkey{location}: \jsontype{object}\},\\
\hspace*{4em}...\\
\hspace*{2em}],\\
\hspace*{2em}\jsonkey{suggested\_plan\_patch}: \jsontype{object},\\
\hspace*{2em}\jsonkey{notes\_for\_slide\_agent}: \jsontype{str}\\
\}\\
\grayarrow Keep issues array length <= max\_issues from the payload.\\
\grayarrow Use short, precise ids (e.g., "IMAGE\_PRESENT\_BUT\_NO\_HERO").
\end{promptbox}

\begin{promptbox}{D.3-2}{LaTeX Compiler Debugger Agent}
% Source: compiler_service.py
You are an \textcolor{promptteal}{\textbf{expert LaTeX debugger}}.\\
Your goal is to fix the compilation error reported by the user.

\par\medskip
You have access to file system tools and a compilation tool:\\
\grayarrow \texttt{read\_file(filename, start, end)}\\
\grayarrow \texttt{write\_file(filename, content)}\\
\grayarrow \texttt{search\_replace(filename, old, new)}\\
\grayarrow \texttt{list\_files()}\\
\grayarrow \texttt{grep\_files(pattern)}\\
\grayarrow \texttt{run\_python\_script(script\_content)}\\
\grayarrow \texttt{create\_image\_placeholder(filename)}\\
\grayarrow \texttt{create\_plot\_image(filename, title, data\_type)}\\
\grayarrow \texttt{check\_balance(filename)}\\
\grayarrow \texttt{compile\_pdf()}

\par\medskip
Strategy:\\
1. \textbf{THINK}: Analyze the error message and the context.\\
2. \textbf{OBSERVE}: Use \texttt{grep\_files}, \texttt{read\_file}, or \texttt{check\_balance} to locate the error. NOTE: Log line numbers are often inaccurate for included files.\\
3. \textbf{ACT}: Fix the error using the appropriate tools.\\
4. \textbf{VERIFY}: Call \texttt{compile\_pdf()} to check if the error is resolved. If \texttt{compile\_pdf()} returns "SUCCESS", reply "FIXED". If it fails, analyze the new error and repeat.\\
5. \textbf{COMPLETE}: If you cannot fix it after several attempts, explain why.

\par\medskip
\textcolor{promptred}{\textbf{CRITICAL}}: Always output your thought process (THINK) before calling tools.
\end{promptbox}

\begin{promptbox}{D.3-3}{Structural Analyst Agent}
% Source: slide_graph_generator.py
You are a \textcolor{promptteal}{\textbf{professional presentation structural analyst}}.\\
Your task is to align generated PDF pages with source LaTeX frames and speech scripts.

\par\medskip
Inputs:\\
1. Text from PDF pages.\\
2. Speech fragments.\\
3. LaTeX Frame codes.

\par\medskip
Output:\\
Generate a JSON list where each element corresponds to a PDF page.\\
\par\medskip
[\{
\par\noindent\hspace*{2em}\jsonkey{pdf\_page\_index}: \jsontype{<int>},
\par\noindent\hspace*{2em}\jsonkey{speech}: \jsonval{"<string>"},
\par\noindent\hspace*{2em}\jsonkey{matched\_frame\_index}: \jsontype{<int or null>}
\par\noindent\}]
\grayarrow matched\_frame\_index: 1-based index of the matching LaTeX frame, or null if it's a structural page (Cover, TOC, etc).\\
\grayarrow speech: The corresponding speech text.
\par\medskip
IMPORTANT:\\
\grayarrow Output valid JSON only.\\
\grayarrow If the speech or content contains backslashes, you \textcolor{promptred}{\textbf{MUST}} escape them (e.g. use \jsonval{"\textbackslash\textbackslash section"} instead of \jsonval{"\textbackslash section"}).\\
\grayarrow Return \textcolor{promptred}{\textbf{ONLY}} the JSON list.
\end{promptbox}

\begin{promptbox}{D.3-4}{VLM Presentation Design Expert}
% Source: vlm_beautify.py
You are a \textcolor{promptteal}{\textbf{Presentation Design Expert}}. Analyze the slide image and the corresponding LaTeX code.

\par\medskip
Your tasks are:\\
1. Layout: Prevent any element overlap. If the content in `latex\_code` does not fully appear in the image, it means there is too much content beyond the display range, and simplification or reduction is needed.\\
2. Images: Check if images are too small or too large. Resize them to fit the slide comfortably.\\
3. Content Density: Check if the slide has too much text (cluttered) or too little (empty). Balance the whitespace.\\
4. Safety: \textcolor{promptred}{\textbf{DO NOT}} introduce any placeholder blocks (e.g., solid rectangles), colored boxes, or new images that do not exist in the project.

\par\medskip
Suggestions:\\
\grayarrow Do NOT use \textbackslash Large or \textbackslash textbf manually for slide titles inside the content area. Use \textbackslash frametitle\{...\} or \textbackslash framesubtitle\{...\} instead.\\
\grayarrow Do NOT add any footers or signatures.

\par\medskip
Return \textcolor{promptred}{\textbf{ONLY}} the modified \textbackslash begin\{frame\}...\textbackslash end\{frame\} block in ```latex``` code block.
\end{promptbox}

% =================================================================
\subsection{Interactive Editing \& Coaching}
% =================================================================

\begin{promptbox}{D.4-1}{Editor Planner Agent}
% Source: editor_ai_service.py
You are a \textcolor{promptteal}{\textbf{Presentation Editor Planner}}.\\
Your goal is to break down the user's modification instruction into a list of specific executable actions.\\
You can \textcolor{promptred}{\textbf{ONLY}} use the following allowed actions: ["MODIFY\_TITLE\_CONTENT", "MODIFY\_SPEECH", "MODIFY\_SLIDE\_CONTENT"].

\par\medskip
Return the plan as a strictly valid JSON list of objects, where each object has:\\
\grayarrow 'action': One of the allowed actions.\\
\grayarrow 'instruction': The specific sub-instruction for that action.

\par\medskip
Example Output:\\
\relax
[\\
\hspace*{2em}\{\jsonkey{action}: \jsonval{"MODIFY\_SLIDE\_CONTENT"}, \jsonkey{instruction}: \jsonval{"Change the bullet point about accuracy to 99\%"}\},\\
\hspace*{2em}\{\jsonkey{action}: \jsonval{"MODIFY\_SPEECH"}, \jsonkey{instruction}: \jsonval{"Mention that our accuracy reached 99\%"}\}\\
]\\
If the user instruction cannot be fulfilled by allowed actions, ignore that part or return empty list.
\end{promptbox}

\begin{promptbox}{D.4-2}{LaTeX Beamer Content Editor}
% Source: editor_ai_service.py
You are a \textcolor{promptteal}{\textbf{LaTeX Beamer expert}}. Modify the content based on instruction.\\
Return \textcolor{promptred}{\textbf{ONLY}} the modified \textbackslash begin\{frame\}...\textbackslash end\{frame\} block in ```latex```. Avoid using '\&' symbol in text, use 'and' instead.
\end{promptbox}

\begin{promptbox}{D.4-3}{Speech Refiner Agent}
% Source: slide_graph_generator.py
You are \textcolor{promptteal}{\textbf{rewriting a presentation speech script}}.\\
The current slide should reference these previous slides:\\
\{ref\_context\}

\par\medskip
Requirements:\\
1) Rewrite the FULL speech into a natural, fluent narrative (do not append a separate reference list).\\
2) Use first-person narrator voice (e.g., "we", "I", "let's").\\
3) Integrate the references into suitable positions in the speech (e.g., opening bridge or when introducing a concept).\\
4) \textcolor{promptred}{\textbf{Do NOT}} output any markers like "ref", "[ref:...]", "[[ref:...]]".\\
5) Avoid duplication; each referenced slide should be used at most once.\\
Return \textcolor{promptred}{\textbf{ONLY}} the rewritten speech text.
\end{promptbox}

\begin{promptbox}{D.4-4}{Presentation Logic Analyzer}
% Source: slide_graph_generator.py
You are an \textcolor{promptteal}{\textbf{expert at analyzing presentation logic}}.\\
The user Logic Chain states that Section '\{src\_sec\}' references Section '\{dst\_sec\}'.\\
Your task is to find the SPECIFIC slides in '\{src\_sec\}' that reference SPECIFIC slides in '\{dst\_sec\}'.

\par\medskip
Input:\\
1. Source Slides (from '\{src\_sec\}')\\
2. Target Slides (from '\{dst\_sec\}')

\par\medskip
Output JSON:\\
\{\\
\hspace*{2em}\jsonkey{edges}: [\\
\hspace*{4em}\{\jsonkey{from}: \jsontype{<source\_slide\_id>}, \jsonkey{to}: \jsontype{<target\_slide\_id>}, \jsonkey{reason}: \jsonval{"..."}\}\\
\hspace*{2em}]\\
\}\\
If no specific reference is found, return \jsonkey{edges}: [].
\end{promptbox}

\begin{promptbox}{D.4-5}{Rehearsal Coach Agent}
% Source: preview_insights_service.py
You are a \textcolor{promptteal}{\textbf{rehearsal coach for academic talks}}.\\
Based on the provided slide content and metrics, write actionable, concrete tips.

\par\medskip
Output \textcolor{promptred}{\textbf{STRICT JSON only}}: \{\jsonkey{advice}: [\jsonval{"..."},\jsonval{"..."},...]\}.

\par\medskip
Rules:\\
\grayarrow Write in English.\\
\grayarrow Return 3 to 6 tips.\\
\grayarrow Each tip must be short (<= 12 words) and actionable.\\
\grayarrow No markdown, no explanations, no extra text.
\end{promptbox}

\begin{promptbox}{D.4-6}{Audience QA Agent}
% Source: preview_insights_service.py
You generate likely audience questions for an academic talk.\\
Based on the slide content and risk metrics, write the 3 most likely questions.

\par\medskip
Output \textcolor{promptred}{\textbf{STRICT JSON only}}: \{\jsonkey{questions}: [\jsonval{"Q1"},\jsonval{"Q2"},\jsonval{"Q3"}]\}.

\par\medskip
Rules:\\
\grayarrow Write in English.\\
\grayarrow Questions must be specific, sharp but polite.\\
\grayarrow No markdown, no explanations, no extra text.
\end{promptbox}

\SysPromptsStandaloneEnd

    \begin{figure}
        \centering
        \includegraphics[width=\linewidth]{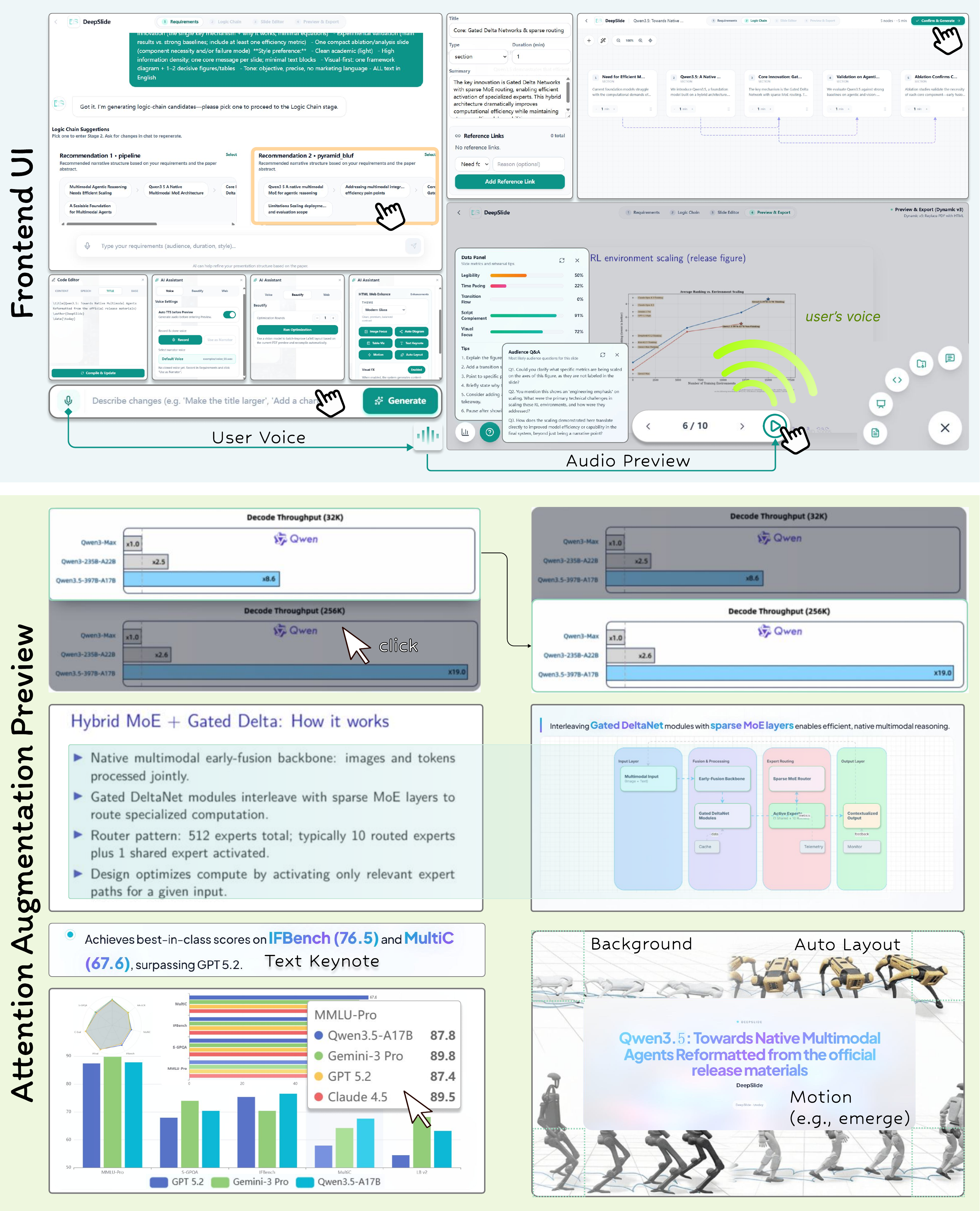}
        \caption{\textit{DeepSlide}'s UI and attention augmented effect preview.}
        \label{fig:ui}
    \end{figure}

    \section{Related Work}

We review automatic presentation generation and its evaluation, surveying agents from early systems to LLM frameworks, and expose the gap between current metrics and the need for holistic, audience-centric, delivery-oriented assessment.

\subsection{Existing Slide Agents}

\paragraph{Early academic systems.} 
Early systems distilled scientific documents into slides via content extraction. DOC2PPT~\cite{rw_1_fu2022doc2ppt} introduced a hierarchical seq-to-seq model and a 6k document-slide benchmark. SlideSpawn~\cite{rw_2_kumar2024slidespawn} refined selection by PDF-to-XML conversion, salience-driven ML ranking, and ILP-based sentence picking. Both prioritized summarization over design coherence or audience tailoring.

\paragraph{LLM-based multi-agent frameworks.}
Leveraging large language models, recent multi-agent frameworks automate presentation generation. 
PASS~\cite{rw_3_aggarwal2025pass} pioneers a Word-to-slide pipeline with synchronized speech synthesis. 
Auto-Slides~\cite{rw_4_yang2025autoslides} enables dialog-driven customization, enhancing controllability. 
PPTAgent~\cite{rw_zheng-etal-2025-pptagent} refines slides via two-stage, edit-based generation guided by reference patterns, yielding superior content, aesthetics, and coherence. PreGenie~\cite{rw_xu2025pregenieagent} iteratively optimizes multimodal slides through analysis–generation–review cycles, ensuring aesthetic and semantic consistency.

% \textbf{Editing-Centric and Interaction-Focused Agents.} 
\paragraph{Editing-centric and interaction-focused agents.}
Beyond end-to-end slide generation, recent work studies instruction-following edits over existing decks. 
PPTArena~\cite{rw_ofengenden2025pptarena} benchmarks natural-language slide editing and introduces PPTPilot, which improves controllability via structure-aware planning and verification. 
PresentAgent~\cite{rw_shi-etal-2025-presentagent} further extends the setting to multimodal presentation video generation.

\paragraph{Audience-aware and narrative-focused approaches.}
More closely related are methods that model audience and narrative structure. Persona-aware D2S~\cite{rw_mondal-etal-2024-presentations} introduces audience expertise and duration as control variables, but its binary modeling is coarse for fine-grained background and time constraints. 
For narrative structure, NarrativeNet Weaver~\cite{rw_meier2025developing} uses hybrid vector--graph retrieval to maintain entity consistency and facilitate narrative chains, and He et al.~\citep{rw_he2024narrative} outline a four-phase framework (Data, Narration, Visualization, Presentation) with LLM-assisted narration tasks. 
% However, existing work has not treated narrative chains as explicit, reusable intermediate representations to support systematic reuse across audiences and durations.

\paragraph{User-centric systems.} 
Additionally, industrial systems (Microsoft Copilot \citep{microsoft-copilot}, Google Gemini \citep{google-gemini}, 
Beautiful.ai \citep{beautiful-ai} and Gamma \citep{Gamma-web}) 
enable easy generation. 
However, although these products can produce visually appealing slides, they often overlook control over presentation pacing and narrative style, and have yet to consider reducing users' preparation burden from the perspective of the complete presentation workflow.

\subsection{Evaluation for AI-Generated Presentations}

% \textbf{Traditional Text-Based Metrics.} 
\paragraph{Traditional text-based metrics.}
Early studies primarily reused summarization metrics, such as ROUGE~\cite{rw_lin2004rouge} and BLEU~\cite{rw_papineni2002bleu}, to measure lexical overlap between slide text and sources. 
Later work adopted embedding-based metrics (e.g., BERTScore~\cite{rw_zhang2019bertscore} and MoverScore) to better capture semantics. 
However, text-only metrics ignore visual design, layout structure, and text--image consistency.

\paragraph{Multi-dimensional evaluation frameworks.}
To more comprehensively evaluate slide quality, 
researchers have proposed multi-dimensional evaluation frameworks. PASS~\cite{rw_3_aggarwal2025pass} introduced LLM-based evaluation metrics, 
assessing presentations from three key dimensions: relevance, coherence, and redundancy. 
PPTAgent~\cite{rw_zheng-etal-2025-pptagent} proposed the PPTEval framework, which evaluates presentation quality comprehensively from content, 
design, and coherence dimensions, significantly surpass traditional single-metric evaluations. 
PreGenie~\cite{rw_xu2025pregenieagent} combines text-image relevance (CLIP Score) and figure proportion metrics to ensure multimodal
 consistency. Knowledge-Centric Templatic Views (KCTV)~\cite{rw_cachola2024knowledge} proposed a template-agnostic evaluation framework TAE, 
 adopting a Precision-Recall style that better aligns with human preferences.

\paragraph{Task-specific benchmarks.} 
Recent benchmarks target specific capabilities such as generation, editing, and visual reasoning. SlidesGen-Bench~\citep{rw_slidesgen2025} evaluates content, aesthetics, and editability with Slides-Align1.5k, while PPTC Benchmark~\cite{rw_guo2024pptc} measures multi-turn task completion via turn/session accuracy. 
PPTArena~\cite{rw_ofengenden2025pptarena} focuses on instruction-following edits with dual VLM-as-judge scoring, and PPTBench~\cite{rw_huang2025pptbench} probes layout/design understanding across detection, understanding, modification, and generation.

\paragraph{Human evaluation and LLM-as-a-judge.} 
Human studies remain important for assessing slide quality, often along axes such as informativeness, persona fit, duration suitability, and coherence~\cite{rw_1_fu2022doc2ppt}. 
More recently, LLM-as-a-judge has been adopted for scalable evaluation; PPTAgent~\cite{rw_zheng-etal-2025-pptagent} reports alignment between MLLM judging and human ratings, and REFLEX~\cite{rw_muppidi2025taming} proposes reference-free judging via negative-sample fine-tuning to produce actionable feedback.

    \section{Reproducibility Statement}
    We will release the source code, configuration files, prompts, and evaluation scripts for reproducing the full pipeline (planning, retrieval, rendering, sandbox validation, and scoreboard evaluation) at:
    \url{https://github.com/PUITAR/DeepSlide}.
\end{appendices}
\end{document}